\documentclass[11pt]{article}

\usepackage[preprint]{acl}

\usepackage{times}
\usepackage{latexsym}
\usepackage[T1]{fontenc}
\usepackage[utf8]{inputenc}
\usepackage{microtype}
\usepackage{inconsolata}
\usepackage{graphicx}
\usepackage{booktabs}
\usepackage{tabularx}
\usepackage{longtable}
\usepackage{array}
\usepackage{enumitem}
\usepackage{xcolor}
\usepackage{colortbl}
\usepackage{colortbl}   %
\usepackage{tikz}
\usepackage{amssymb}                 %
\usetikzlibrary{arrows.meta,calc,positioning,patterns.meta}

\definecolor{abGray}{HTML}{C9C9C9}
\definecolor{abMint}{HTML}{BFDFD2}
\definecolor{abMintEdge}{HTML}{8FB7A6}
\definecolor{abCoral}{HTML}{ED8D5A}
\definecolor{abNearBlack}{HTML}{1A1A1A}
\definecolor{abSubtitle}{HTML}{666666}

\newcommand{\grouprow}[2]{\rowcolor{abGray!35}\multicolumn{#1}{@{}l}{\textbf{#2}}\\}
\newcommand{\comprow}{\rowcolor{abMint!35}}

\usepackage{placeins}

\usepackage{url}

\newcommand{\coderepo}{\url{https://github.com/yzhao062/AI4Fire}}

\setlist[itemize]{leftmargin=*,itemsep=2pt,topsep=3pt}

\title{AI4Fire: Evaluating Large Language Models on Wildfire Tasks}

\author{
  \textbf{Yue Zhao$^1$ \quad Zuobin Xiong$^2$ \quad Xiyang Hu$^3$} \\
  \textbf{Zhangyu Wang$^4$ \quad Xiongye Xiao$^5$ \quad Ruolin Li$^1$} \\
  $^1$University of Southern California \\
  $^2$University of Nevada, Las Vegas \quad $^3$Arizona State University \\
  $^4$University of Maine \quad $^5$University of Tennessee, Knoxville \\
  \texttt{\{yue.z,ruolinl\}@usc.edu} \\
  \texttt{zuobin.xiong@unlv.edu} \quad \texttt{xiyanghu@asu.edu} \\
  \texttt{zhangyu.wang@maine.edu} \quad \texttt{xxiao9@utk.edu}
}

\begin{document}
\maketitle

\begin{abstract}
Large language models (LLMs) are entering wildfire management, where overstated evaluations can cost property and lives. %
How do they perform on wildfire tasks, with and without grounding? %
Bare means a model receives the task input alone. %
Grounded means it also receives one task-specific addition: for smoke detection, a smoke-free reference frame from the same camera. %
AI4Fire runs six core models bare and grounded on five wildfire tasks, zero-shot; a sweep adds 29 more. %
Our literature search on fire tasks found 138 works; none combines this roster, task coverage, and paired bare and grounded runs. %
We report three findings. %
(1)~Grounding helped most where the addition carried the answer: a read-only SQL tool lifted every core model's database accuracy from at most 16 to at least 88 percent. %
(2)~Simple rules were hard to beat: no core model outperformed repeating today's staffing count, and two open-weight models mostly copied the median of similar earlier fire-days, a worse forecast. %
(3)~Public releases carry hazards: a fire-danger column separates the holdout perfectly, and 67 aerial fire frames carry smoldering or fire-free labels read from a clipped thermal maximum. %
We release prompts, responses, scores, code, and the survey record. %
\end{abstract}

\section{Introduction}\label{sec:intro}

\begin{figure*}[t]
\centering
\includegraphics[width=\textwidth]{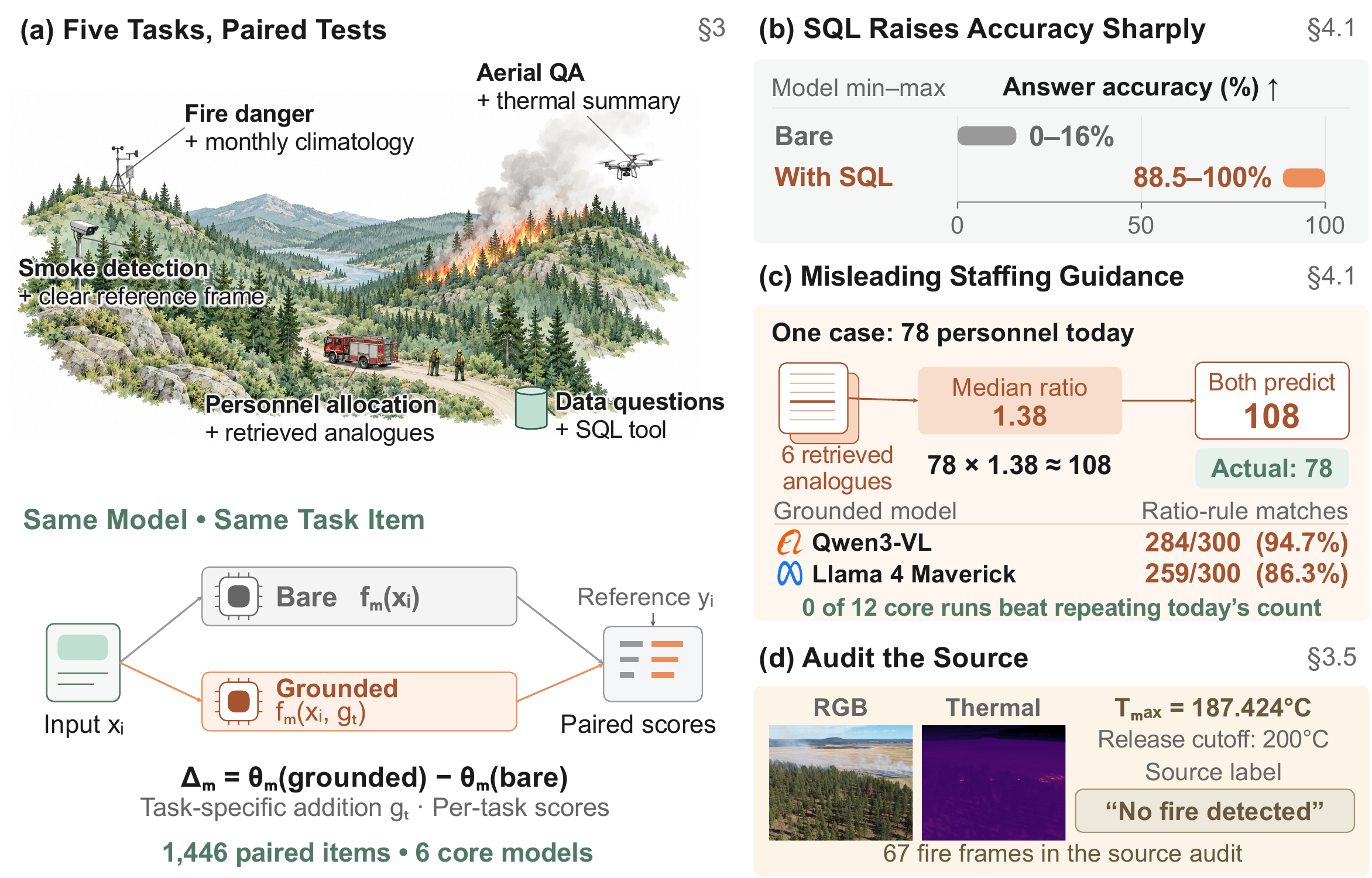}
\caption{\textbf{Task-specific grounding can supply answers or worsen forecasts; source audits reveal label defects.} (a)~Each model answers every item bare and grounded ($g_t$: the task's addition; $\theta_m$: the task metric); the fire scene is illustrative. (b)~Bands span the core six's lowest to highest accuracy in each arm, not confidence intervals. (c)~On Parker Ridge, the allocation example of Appendix~\ref{app:example-prompts}, both grounded open-weight models predict 108, today's count times the displayed median ratio, rounded. Ratio-rule matches are counted over the 300 rule-v1 items (gpt-6-astra: 199) and do not establish internal reasoning. (d)~Actual FLAME~3 inputs and the released CMR4 label for WildFireVQA frame Sycan 2D 00022 \citep{habibpour2026wildfirevqa}. Its maximum, clipped by an unidentified mechanism, falls below the release's 200\,$^\circ$C threshold.}
\label{fig:overview}
\vspace{-8pt}
\end{figure*}

Wildfires kill people and destroy communities: in January 2025, two Los Angeles fires killed 31 people and destroyed more than 16,000 structures \citep{calfire2025eaton,calfire2025palisades}. %
Managing them runs on imagery, records, maps, databases, and forecasts, the same inputs that large language models (LLMs) now take. %
Fire agencies have begun to answer the public with AI chatbots and to test generative tools for operations \citep{calfire2025askcalfire,usdafs2026leveragingai}. %
Evaluations decide which systems earn trust, yet three of the six public fire releases we rebuilt carry hazards that change what a score measures (Section~\ref{subsec:source-audit}). %
Unchecked, such hazards can overstate a score, and a model trusted on that score can miss an ignition or misjudge staffing, errors paid in property and lives. %

This paper asks how frontier LLMs perform across wildfire tasks, and whether grounding them changes what they can do. %
Grounding provides task data, retrieved documents, tools, or rules at inference time. %
Each model runs twice on the same items: bare, with the task input alone, and grounded, with that input plus one task-specific addition (Figure~\ref{fig:overview}a; Section~\ref{subsec:conditions-to-compare}). %
Because each addition appears on one task only, its effect is task-specific, and we estimate no average effect. %

\textbf{Existing Work and Gaps.} Prior work lacks a systematic multi-task assessment across model families (Section~\ref{sec:gaps}): %
(\textit{i})~Fire surveys and organizational reports run no model evaluations \citep{tricomi2024ogcd123,ogc2026genaiwildfire}. %
(\textit{ii})~Domain benchmarks evaluate vision or sequence models without evaluating LLMs or agents \citep{xu2026wildfirefm,tran2025land8fire,zhao2025tssatfire}. %
(\textit{iii})~Existing LLM and agent evaluations cover one task family, one base model, or fire as one hazard among several \citep{hyun2025crewwildfire,zhang2026disasterbench,chen2025geospatial}. %
(\textit{iv})~No surveyed study runs one model set bare and grounded across task families under one protocol. %
The survey may be incomplete (see the Limitations section). %

Breadth is what these gaps call for: a result on one task or model need not carry over. In our runs, grounding's effect changed sign across tasks, and single runs of two reworded fire-danger prompts reordered the models (Section~\ref{sec:results}). %
AI4Fire therefore holds one protocol fixed. Six core models, four proprietary and two open-weight, run all five tasks zero-shot, bare and grounded; a sweep adds 29 more, 10 of them on all five tasks. %
No surveyed evaluation combines this roster, this task coverage, and paired runs. %
AI4Fire's 35 models from twelve vendors exceed any surveyed roster (at most 22), and its tasks span four of the survey's ten task categories against at most three (Table~\ref{tab:positioning}). %

\textbf{Key Findings.} (\textit{i})~\textbf{Grounding helped most where the addition carried the answer, with distinct failure patterns.} A read-only SQL tool lifted core-six database accuracy from at most 0.160 to 0.885 or above (Figure~\ref{fig:overview}b), and the open-weight models' remaining failures were mostly guessed date formats (Table~\ref{tab:failure-modes}). %
Given retrieved analogues (similar earlier fire-days with their outcomes), the two open-weight models copied the displayed analogue median on 284 and 259 of 300 items (Figure~\ref{fig:overview}c). That rule alone forecasts worse than persistence, which repeats today's staffing count (normalized error 0.249 against 0.146). %
(\textit{ii})~\textbf{Simple rules and small trained models were hard to beat.} No core-model run beat persistence, and a trained classifier outscored every prompted fire-danger run. %
(\textit{iii})~\textbf{A leaderboard score can hide miscalibration.} On the negative-subsampled fire-danger sample, all six core models stated probabilities less calibrated than a calendar-month prior. Once the prompt states the sample's positive rate, no proprietary run's Brier score is resolvably worse than the prior's, while the open-weight two stay worse (Section~\ref{subsec:cross-task-insights}). %
(\textit{iv})~\textbf{Question wording changes how far the tool-use task separates models.} The two open-weight core models already trail the proprietary four under the original wording (0.885 and 0.891 against 0.994 to 1.000). Writing schema literals in plain words widens that gap (Section~\ref{subsec:cross-task-insights}). %
(\textit{v})~\textbf{Public releases carry evaluation hazards.} Rebuilding the tasks exposed a Mesogeos column that separates the fire-danger holdout at average precision 1.000. %
It also exposed a clipped thermal maximum on 67 WildFireVQA fire frames, which the release labels smoldering or fire-free. Of the core six's 34 grounded misses on the 48 closed-form aerial items, 15 fall on such frames (Figure~\ref{fig:overview}d). %

Our primary contributions are as follows. %
\setlist{nolistsep}
\begin{enumerate}[leftmargin=*]
  \item \textbf{The Survey and the Benchmark.} We survey 138 recent works and re-check 31 claim by claim (Section~\ref{sec:scope}). %
  AI4Fire covers detection, forecasting, operations, and data questions: 1,446 paired items on five wildfire tasks (Section~\ref{sec:benchmark}). %
  \item \textbf{Gap-Driven Evaluation Design.} The protocol pairs bare and grounded runs, tests non-LLM comparators on every task, clusters uncertainty by dependence unit, and audits each source for defects (Section~\ref{sec:benchmark}, Appendices~\ref{app:dataaudit} and~\ref{app:defects}). %
  \item \textbf{Failure Patterns and Insights.} A table separates model failure patterns, such as copying a displayed median, from benchmark and source properties, such as clipped thermal values (Table~\ref{tab:failure-modes}). Five cross-cutting insights each carry a direction (Section~\ref{sec:results}). %
  \item \textbf{Open and Reproducible Release.} At \coderepo{}, we release every prompt, manifest, item score, raw model response, scoring script, and the literature survey record. %
\end{enumerate}

\section{Fire Tasks and What Existing Work Shows}\label{sec:scope}\label{sec:gaps}\label{sec:crosscut}\label{subsec:main-gap}\label{subsec:closest-works}

\subsection{Scope and Task Categories}

The survey covers LLMs, VLMs, MLLMs, and agents, including multi-agent systems, applied to fire tasks. %
Fire tasks fall in two domains, wildfire and structure fire, spanning fire science, engineering, operations, and emergency management. %
AI4Fire itself is narrower: it runs one model at a time on five wildfire tasks, with a read-only SQL tool on one. It evaluates no multi-agent system and no structure-fire task (Section~\ref{sec:benchmark}). %
LLM agents ran the searches, which kept 138 works first released between 2025-01-01 and 2026-09-13, and tagged each by task category (Appendix~\ref{sec:extended-background}). %
A scoping summary cites 31 of them, which agents re-checked claim by claim; 25 of the 31 carry a fire task. %

\subsection{Closest Works and the Gap}\label{subsec:specific-gaps}

Three lines of prior work bound this paper, and Table~\ref{tab:positioning} scores every listed work against AI4Fire. %
No work found by the searches surveys LLMs and agents across fire tasks and evaluates one set of current models under one protocol. %
Narrower gaps remain on the split between wildfire and structure fire, and on quantitative forecasting and tool use. %
Among the 25 fire records, the two domains cover disjoint task families: structure fire addresses text codes and investigation, while wildfire addresses imagery and forecasting \citep{chen2025regcheck,sun2026triplephase,kim2025firespec,tao2026magrfi}. %
The one direct forecasting comparison, on fire radiative power, favored domain-specific tabular deep learning over a GPT-4-based LLM agent and tested none of the current models evaluated here \citep{ramesh2025wildfiregpt}. %
These searches were not saturated; Appendix~\ref{app:trust-negative} records the per-work search-unit counts. %
Checks made after the window closed found ORCH \citep{ji2026orch}, DORA \citep{wang2026dora}, and FlameVQA \citep{habibpour2026flamevqa}, all released inside the window and not in the survey counts. %
Appendix~\ref{sec:related-work-detail} adds each work's scale and the rest of the closest-works record. The database task relates to text-to-SQL benchmarks such as Spider \citep{yu2018spider}, BIRD \citep{li2023bird}, and Spider~2.0 \citep{lei2025spider2}. Its lookup, aggregate, and multi-step questions query one table, separating memory from database access. Scoring compares final answers on one snapshot, without the query-equivalence checks of distilled test suites \citep{zhong2020semantic}. %

\begin{table}[t]
\centering
\scriptsize
\setlength{\tabcolsep}{2pt}
\renewcommand{\arraystretch}{1.05}
\caption{\textbf{None of the listed prior works combines a re-checked literature survey with a bare-against-grounded evaluation across fire tasks.} $\bullet$ present, $\circ$ survey without claim re-checks, blank absent or unstated. \emph{Cat.}: fire task categories. \emph{Models}: configurations evaluated, with families in parentheses as each work reports them (vendors for AI4Fire). \emph{B/G}: bare and grounded arms on the same items. \emph{Ref.}: non-LLM comparator. Table~\ref{tab:positioning-full} adds scale.}\label{tab:positioning}
\vspace{1pt}
\begin{tabular}{@{}>{\raggedright\arraybackslash}p{3.2cm}ccccc@{}}
\toprule
& \multicolumn{3}{c}{Coverage} & \multicolumn{2}{c}{Compares} \\
\cmidrule(lr){2-4}\cmidrule(l){5-6}
Work & Survey & Cat. & Models & B/G & Ref. \\
\midrule
\grouprow{6}{Fire AI reports with no model evaluation}
OGC D-123 \citeyearpar{tricomi2024ogcd123} & & & & & \\
OGC 25-012 \citeyearpar{ogc2026genaiwildfire} & & & & & \\
\grouprow{6}{Fire benchmarks that evaluate no LLM or agent}
WILDFIRE-FM \citeyearpar{xu2026wildfirefm} & & 1 & & & $\bullet$ \\ %
Land8Fire \citeyearpar{tran2025land8fire} & $\circ$ & 1 & & & $\bullet$ \\
TS-SatFire \citeyearpar{zhao2025tssatfire} & $\circ$ & 2 & & & $\bullet$ \\
\grouprow{6}{LLM and agent evaluations on fire tasks}
CREW-Wildfire \citeyearpar{hyun2025crewwildfire} & & 1 & 4 (1) & & $\bullet$ \\
DisasterBench \citeyearpar{zhang2026disasterbench} & & 3 & 22 (8) & & \\ %
Geospatial Agents \citeyearpar{chen2025geospatial} & & 1 & 14 (3) & $\bullet$ & $\bullet$ \\
SmokeBench \citeyearpar{qi2026smokebench} & & 1 & 7 (6) & & $\bullet$ \\ %
WildfireGPT \citeyearpar{ramesh2025wildfiregpt} & & 1 & 1 (1) & & $\bullet$ \\
WildFireVQA \citeyearpar{habibpour2026wildfirevqa} & & 1 & 4 (4) & $\bullet$ & \\
DisasterVQA \citeyearpar{almohannadi2026disastervqa} & & 1 & 7 (5) & & \\ %
\midrule
\rowcolor{abCoral!25}AI4Fire (this paper) & $\bullet$ & 4 & 35 (12) & $\bullet$ & $\bullet$ \\ %
\bottomrule
\end{tabular}
\vspace{-8pt}
\end{table}

\section{The AI4Fire Benchmark}\label{sec:benchmark}
AI4Fire asks whether grounding a model in the data a fire task needs changes what it can do, under one protocol. %
The benchmark evaluates a core set of six models from five vendors across five wildfire tasks and 1,446 paired items. %
A sweep repeats the protocol on 29 further models, for 35 models from twelve vendors. %
The release carries the harness under the BSD 2-Clause License, with every prompt, manifest, item score, and raw response. %
Four survey patterns (Section~\ref{sec:crosscut}, Appendix~\ref{sec:extended-background}) drove four design decisions (Table~\ref{tab:crosswalk}). %

\subsection{Tasks}\label{subsec:version1-tasks}\label{subsec:candidate-task-families}
Table~\ref{tab:version1tasks} groups the five tasks by the survey's task categories (Table~\ref{tab:taxonomy}) and gives each task's grounding addition, data source, and item count. %
Appendix~\ref{app:conditions-detail} documents how each task's grounding source was built. %
We withdrew Fire360 \citep{tiwari2025fire360}, the one structure-fire source the check passed, because its released prompts provide no mapping to its reported visual evaluation (Appendix~\ref{app:defect-fire360}). %
Finding no other released structure-fire evaluation set, the benchmark reproduces the cross-domain gap of Section~\ref{subsec:specific-gaps} (Appendix~\ref{subsec:left-out} outlines two routes to close it). %

\begin{table}[t]
\centering
\scriptsize
\setlength{\tabcolsep}{2pt}
\renewcommand{\arraystretch}{1.05}
\caption{\textbf{Two of the five additions can carry an item's answer; the other three give context to reason from.} \emph{Addition}: what the grounded arm adds. \emph{Ans.}: $\bullet$ can carry the reference answer (the SQL tool on every item, the thermal block on 48 closed-form and 13 other items). $n$: scored items (smoke: 196 paired of 224 frames). Appendices~\ref{app:conditions-detail} and~\ref{subsec:data-sources-licensing} give details and licenses.}\label{tab:version1tasks}
\vspace{1pt}
\begin{tabular}{@{}llcl r@{}}
\toprule
& \multicolumn{2}{c}{Grounded arm} & & \\
\cmidrule(lr){2-3}
Task & Addition & Ans. & Data source & $n$ \\
\midrule
\grouprow{5}{Detection and Perception}
Smoke & Reference frame & & FIgLib \citeyearpar{dewangan2022figlib} & 196 \\
Aerial QA & Thermal block & $\bullet$ & WildFireVQA \citeyearpar{habibpour2026wildfirevqa} & 408 \\
\grouprow{5}{Forecasting and Prediction}
Fire danger & Climatology & & Mesogeos \citeyearpar{kondylatos2023mesogeos} & 386 \\
\grouprow{5}{Decision Support and Operations}
Allocation & Analogues & & ICS-209-PLUS \citeyearpar{stdenis2023ics209plus} & 300 \\
\grouprow{5}{Data Retrieval and Tool Use}
Tool use & SQL tool & $\bullet$ & FPA-FOD \citeyearpar{short2022fpafod} & 156 \\
\bottomrule
\end{tabular}
\vspace{-8pt}
\end{table}

\subsection{Conditions to Compare}\label{subsec:conditions-to-compare}
Every task runs bare and grounded on the same model set, because of the grounding gap of Section~\ref{subsec:specific-gaps}. %
The bare arm is the task input alone, with the grounded arm's addition withheld, and Appendix~\ref{app:terms} defines grounding. %
Every prompt is zero-shot, with no worked examples (Appendix~\ref{app:benchdetail}). %
Each of the five appears on exactly one task, so the design estimates no average grounding effect. %

Two of the five can carry an item's reference answer. %
A read-only SQL query returns the records a question asks about. %
For 48 aerial items of four question types, a fixed rule reproduces the release's label from the thermal summary block, using class thresholds the prompt does not state (Appendix~\ref{app:defects}). %
The reference frame, monthly climatology, and retrieved analogues supply context the model has to reason from. %
We report those 48 prespecified closed-form items apart from the remaining 360, a split by question type rather than by answer availability. %
Of the remaining 360, 13 items of four further types have labels that a maximum below 200\,$^\circ$C fixes as no hotspot or no fire. %

\subsection{Model Set}\label{subsec:model-set}
The core set of six models spans three axes: vendor, generation, and open versus closed weights. %
Four are proprietary: claude-opus-5, the frontier generation; claude-opus-4.8, the generation before it; and gemini-3.1-pro and gpt-6-astra from two further vendors. %
Two are open-weight, Qwen3-VL-235B-A22B \citep{bai2025qwen3vl} and Llama 4 Maverick \citep{meta2025llama4}. %
Every diagnostic of Section~\ref{sec:results} runs on these six, requesting temperature 0 (gpt-6-astra's endpoint fixes 1.0); effective gateway settings are unverified (Appendix~\ref{app:reasoning}).
A sweep adds 29 further models served through Amazon Bedrock under the same prompts, items, and scorers, selected by a probe of image input and tool use. %
Ten accept both and ran all five tasks; nineteen are text-only (Section~\ref{subsec:results-sweep}). %
Appendices~\ref{app:modelset-detail}, \ref{app:model-sweep}, \ref{app:manifest}, and~\ref{app:repro} give serving paths, output caps, checksums, example prompts, item sampling, and reasoning settings. %

\subsection{Metrics and Scoring}\label{subsec:metrics-and-scoring}
No task needs a human or an LLM judge: every reference answer is exact or numeric, so all scoring is automatic. %
Appendix~\ref{app:scoring-detail} gives each task's scoring rule and comparator. %

Fires differ in size, so allocation error divides each absolute error by the staffing level of its own fire rather than by a single global mean:
\begin{equation}\label{eq:alloc}
E = \frac{1}{n}\sum_{i=1}^{n}\frac{|y_i - \hat{y}_i|}{s_{f(i)}}.
\end{equation}
Here $y_i$ is the filed personnel count on fire-day $i$, $\hat{y}_i$ the prediction, and $n$ the number of fire-days. %
The normalizer $s_{f(i)}$ is the mean filed staffing of the fire $f(i)$ the day belongs to. %

Grounding is measured as a paired difference inside one model and one task: the same items run with and without the addition isolate the addition's effect.
For model $m$,
\begin{equation}\label{eq:contrast}
\Delta_m = \theta_m^{\mathrm{grounded}} - \theta_m^{\mathrm{bare}},
\end{equation}
with $\theta_m$ the metric on items both arms completed. %

Items that share a fire, frame, incident, spatial block, or question family are not independent, so intervals resample whole clusters rather than items (Appendix~\ref{app:scoring-detail}). %
With few clusters these intervals may run too narrow, so we also report a widened interval that claims no nominal coverage. %
A contrast is resolved when its 95 percent interval excludes zero or an exact paired test gives $p<0.05$. Each interval or test conditions on one stored run per arm, unadjusted for the number of models and tasks, so counts of resolved contrasts are descriptive. %
A fixed seed and released code reproduce every interval. %

Every task is also scored beside a simple or trained non-LLM comparator, because a score means little on its own (Section~\ref{sec:results}, Appendix~\ref{app:scoring-detail}). %

\subsection{What Building These Tasks Revealed About the Sources}\label{subsec:source-audit}

Building the six candidate tasks turned up properties that change what a score measures, in three public releases and in our own harness. %
Mesogeos Track A has a column separating its holdout at average precision 1.000. %
Released Fire360 questions name no video or frame and put the correct choice first on 25 of 50 items. %
Closed-form rules on the WildFireVQA thermal block reproduce four question types on all 6,097 frames. %
On 67 frames that FLAME~3 files as fire, the released thermal maximum is clipped: it is exactly 187.424\,$^\circ$C on all 67, and no frame of their Sycan Marsh burn peaks between that value and 200\,$^\circ$C. The release labels them smoldering or fire-free on every temperature-derived question (Appendix~\ref{app:defect-ceiling}). %
Our initial 200-token output cap truncated a reasoning model on 40 of 196 items. %
Each hazard except the clipping was re-derived twice from raw files, first by an agent that did not see the task builders. Table~\ref{tab:source-audit} (Appendix~\ref{app:defects}) records each as a caution for future users. %
None claims a published evaluation is wrong. %

\section{Results, Insights, and Future Directions}\label{sec:results}\label{sec:directions}

Tasks ran paired on six core models under a 1,536-token output cap, beside a 29-model sweep described in Section~\ref{subsec:results-sweep}. %

\begin{table*}[t]
\centering
\footnotesize
\setlength{\tabcolsep}{3pt}
\renewcommand{\arraystretch}{1.15}
\caption{\textbf{Only the read-only SQL tool clears zero on all six core models.} Parentheses give the bootstrap clusters. $^{\dagger}$A held-out frame-difference detector, shown by recall (clear-frame false-positive rate 0.429, accuracy 0.770; on the 168 rerun targets, 0.726 here and 0.536 once gaps are matched, Appendix~\ref{app:figlib-rerun}). \emph{Bare}, \emph{Grounded}, and \emph{$\Delta$ span} give the core six's score ranges and smallest and largest paired differences, as point estimates. \emph{Excl.\ 0} counts models whose 95 percent clustered interval excludes zero ($\blacktriangle$ grounded, $\blacktriangledown$ bare, $\circ$ neither). $^{*}$Widened 11.5 percent as a finite-cluster check over 17 fires, with no nominal coverage. Other comparators: trained regressor 0.144 (allocation), hybrid rule 0.664 (aerial), boosted classifier 0.851 (fire danger), reference SQL 1.000 (tool use).}\label{tab:cross-task-summary}
\begin{tabular}{@{} l l l c c l c @{}}
\toprule
 & \multicolumn{2}{c}{Evaluation} & \multicolumn{2}{c}{Score range, core six} & \multicolumn{2}{c}{Paired contrast} \\
\cmidrule(lr){2-3}\cmidrule(lr){4-5}\cmidrule(lr){6-7}
Metric & Items (clusters) & Comparator & Bare & Grounded & $\Delta$ span & Excl.\ 0 \\
\midrule
\grouprow{7}{Personnel allocation, ICS-209-PLUS.\mdseries{} Grounding: up to six retrieved analogues (rule~v1).} %
Norm. error $\downarrow$ & 300 (245 incidents) & Persistence 0.146 & 0.157--0.193 & 0.170--0.248 & $-0.001$ to $+0.087$ & 3 $\blacktriangledown$ \\ %
\midrule
\grouprow{7}{Smoke detection, FIgLib, first design.\mdseries{} Grounding: one earlier clear frame from the same camera.} %
Smoke recall $\uparrow$ & 196 (17 fires) & Frame diff. 0.920$^{\dagger}$ & 0.509--0.643 & 0.554--0.741 & $+0.036$ to $+0.188$ & 4$^{*}$ $\blacktriangle$ \\ %
\midrule
\grouprow{7}{Aerial question answering, WildFireVQA.\mdseries{} Grounding: a seven-number thermal summary block.} %
Accuracy $\uparrow$ & 408 (390 frames) & Majority 0.627 & 0.534--0.650 & 0.544--0.657 & $+0.000$ to $+0.015$ & 0 $\circ$ \\ %
\midrule
\grouprow{7}{Fire danger forecasting, Mesogeos Track A.\mdseries{} Grounding: monthly climatology.} %
AUPRC $\uparrow$ & 386 (352 blocks) & Temp. rule 0.654 & 0.485--0.697 & 0.515--0.682 & $-0.032$ to $+0.051$ & 1 $\blacktriangle$ \\ %
\midrule
\grouprow{7}{Fire data tool use, FPA-FOD.\mdseries{} Grounding: a read-only SQL tool, up to eight queries.} %
Accuracy $\uparrow$ & 156 (12 families) & Constant 0.141 & 0.000--0.160 & 0.885--1.000 & $+0.808$ to $+0.994$ & 6 $\blacktriangle$ \\ %
\bottomrule
\end{tabular}
\vspace{-8pt}
\end{table*}

\begin{figure*}[t]
\centering
\includegraphics[width=\textwidth]{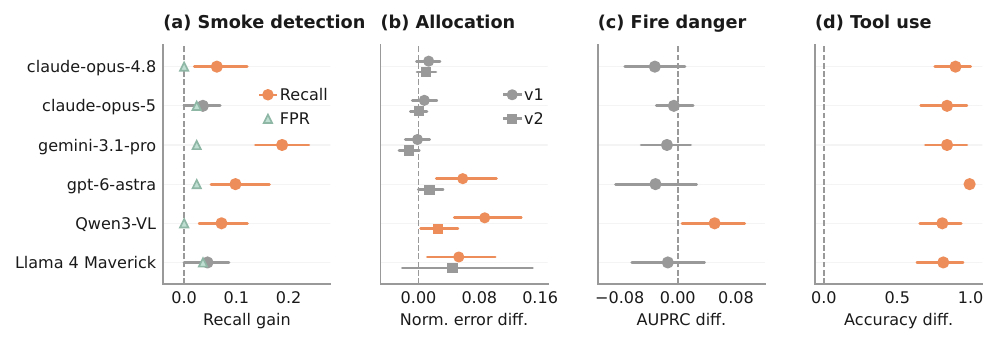}
\caption{\textbf{The reference frame, retrieved analogues, monthly climatology, and read-only SQL tool move four tasks in different directions.} Paired grounded-minus-bare differences with 95 percent intervals clustered as in Table~\ref{tab:cross-task-summary}; coral intervals exclude zero; gray ones include it. (a)~Smoke (FIgLib, first design; rerun in Appendix~\ref{app:figlib-rerun}): recall gain (circles) and false-positive rate change (mint triangles). (b)~Allocation (ICS-209-PLUS): normalized-error difference under retrieval rules v1 (circles) and v2 (squares), which also matches recent staffing change and fire growth (Appendix~\ref{app:retrieval-v2}). (c)~Fire danger (Mesogeos Track A): average precision difference. (d)~Tool use (FPA-FOD): accuracy difference. Figure~\ref{fig:wildfirevqa} shows aerial QA.}\label{fig:grounding-effects}
\vspace{-8pt}
\end{figure*}

\subsection{Results by Task}\label{subsec:results-tasks}

\paragraph{Personnel Allocation on ICS-209-PLUS.}\label{subsec:results-allocation}
The task predicts next-day filed personnel, and the grounded arm adds up to six retrieved analogues with their outcomes (Table~\ref{tab:cross-task-summary}; Appendix~\ref{app:conditions-detail}). %
No core model beats persistence at 0.146 normalized error, and a trained regressor matches it at 0.144 (Appendix~\ref{app:trained-baselines}). %
Analogues raise error on five of six core models, three with intervals excluding zero (Figure~\ref{fig:grounding-effects}b), and the two open-weight models mostly copy their displayed median (Figure~\ref{fig:allocation-analogues}a). Under the movement-matched rule~v2, grounded error falls on all six models relative to rule~v1, and the proprietary four land within 0.015 of bare with no v2-minus-bare interval excluding zero; the displayed-median rule alone still exceeds persistence by 0.044 [0.019, 0.072] (Appendix~\ref{app:retrieval-v2}). %

\paragraph{Smoke Detection on FIgLib.}\label{subsec:results-figlib}
The grounded arm pairs each of the 112 smoke and 84 clear frames with an earlier clear frame from the same camera (Table~\ref{tab:cross-task-summary}; Appendix~\ref{app:figlib-unpaired}). %
Across the core six, the reference frame turns 4 to 21 of 112 smoke frames from missed to found, losing none (Figure~\ref{fig:grounding-effects}a), while four raise clear-frame false positives. %
In this design the reference-to-target gap alone separates the labels, and every frame prints its capture time (Table~\ref{tab:failure-modes}). %
A rerun on 168 of these frames gives both labels gaps of 10, 20, and 30 minutes, masks the time, and calls the reference only an earlier frame (Appendix~\ref{app:figlib-rerun}). %
Pooled recall rises by 0.063 (widened [0.035, 0.102]), resolved for the proprietary four; balanced accuracy rises by 0.024 [0.000, 0.048].
On these pairs the frame-difference detector nears chance (0.536 accuracy), below every model arm. %

\begin{figure*}[t]
\centering
\includegraphics[width=\textwidth]{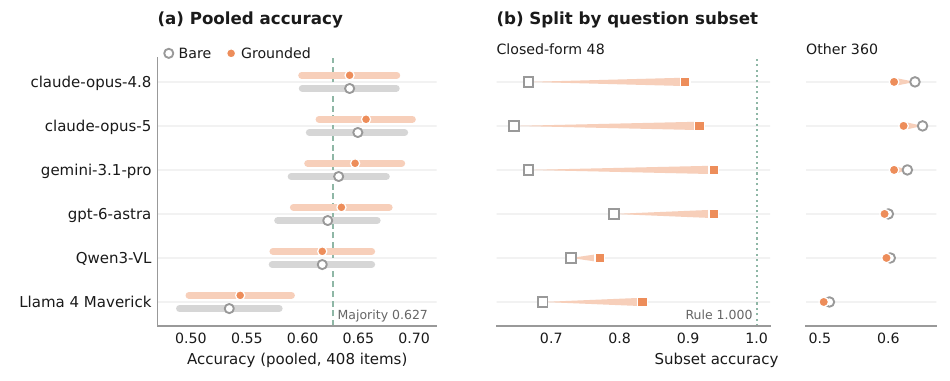}
\caption{\textbf{The thermal summary block leaves pooled accuracy near the majority comparator while splitting closed-form questions from the rest.} (a)~Accuracy on 408 items with frame-clustered 95 percent intervals; dashed: 0.627 majority. Every paired grounded-minus-bare interval includes zero. (b)~48 closed-form (squares) and 360 others (circles), same scale; bare hollow, grounded filled; shading joins pairs; dotted: 1.000 closed-form rule.}\label{fig:wildfirevqa}
\vspace{-8pt}
\end{figure*}

\paragraph{Aerial Question Answering on WildFireVQA.}\label{subsec:results-wildfirevqa}
We score agreement with released labels on 408 multiple-choice questions from 390 FLAME~3 frames across 34 types, adding a seven-number thermal block when grounded (Table~\ref{tab:cross-task-summary}; Appendix~\ref{app:wildfirevqa-detail}).
No core model's paired interval against the held-out 0.627 majority lies wholly above zero, and grounding moves whole-suite accuracy by at most 0.015 (Figure~\ref{fig:wildfirevqa}a). %
The thermal block lifts accuracy on the 48 closed-form items from 0.646--0.792 bare to 0.771--0.938 grounded (Table~\ref{tab:results-wildfirevqa}). Grounded accuracy falls on the other 360 items by 0.006 to 0.031 on all six core models. %
Of the 408 labels, 276 come from multimodal model verification, and the release rates 124 as not fully answerable from their frame. %
Every core-model arm trails the majority on the 132 items labeled by formula, telemetry, or detector, which include 14 hotspot references that conflict with the prompt (Appendix~\ref{subsec:app-wildfirevqa-detail}). %
On the 276 model-verified items, two bare Claude arms have unadjusted frame-clustered intervals above the majority, and neither survives a Holm correction or two-way resampling. %
Each question type has one label source, confounding label quality with question type. %
No core model resolves above the majority in either applicability stratum (Table~\ref{tab:aerial-applicability}). %

\paragraph{Fire Danger Forecasting on Mesogeos Track A.}\label{subsec:results-mesogeos}
The task evaluates 386 items across 352 blocks by month. Negative subsampling yields a 0.339 positive rate, unstated in the prompt, that reflects dataset selection rather than regional prevalence (Table~\ref{tab:cross-task-summary}; Appendix~\ref{app:defect-mesogeos}). %
A boosted classifier reading the prompt numbers scores 0.851, exceeding the best original-prompt run by 0.154 [0.084, 0.227] (Appendix~\ref{app:trained-baselines}). Only two of the six core models beat the 0.654 temperature rule. %
Monthly climatology lowers AUPRC on all four proprietary core models by 0.005 to 0.032, with no block-clustered interval excluding zero (Figure~\ref{fig:grounding-effects}c). Expected calibration errors across the twelve core-model runs span 0.119 to 0.326 against 0.037 for the calendar prior (Figure~\ref{fig:calibration}). %
Stating the rate in the prompt cuts proprietary ECEs to 0.025~to~0.148~(Appendix~\ref{app:baserate-rerun}). %

\paragraph{Fire Data Tool Use on FPA-FOD.}\label{subsec:results-tooluse}
The task asks 156 questions over 2,303,566 records in twelve families. Both arms get the schema, and the tool arm may query a frozen SQLite snapshot up to eight times (Table~\ref{tab:cross-task-summary}; Appendix~\ref{app:tooluse-detail}). %
Bare models score 0.000 to 0.160, but the SQL tool lifts them to 0.885--1.000 with paired gains of 0.808 to 0.994 (Figure~\ref{fig:grounding-effects}d). %
The output cap cut off 61 of gemini-3.1-pro's bare answers (Appendix~\ref{app:defect-ours}). %

\paragraph{Model Sweep Across 29 Additional Systems.}\label{subsec:results-sweep}
Ten of the 29 further models accept images and tool calls, forming sixteen full-capability models with the core six; the other nineteen are text-only (Table~\ref{tab:model-sweep}, Figures~\ref{fig:text-sweep} and~\ref{fig:grounding-sweep}; Appendix~\ref{app:model-sweep}). %
The SQL tool lifts 14 of sixteen full-capability models to 0.827 or above (a descriptive tally); Gemma 3 12B and 4B score 0.000 because interface failures prevented database execution. %

\subsection{Cross-Task Insights and Future Directions}\label{subsec:cross-task-insights}\label{subsec:survey-directions}\label{subsec:results-directions}

\begin{table}[t]
\centering
\scriptsize
\setlength{\tabcolsep}{3pt}
\renewcommand{\arraystretch}{1.0}
\caption{\textbf{Selected model failure patterns and properties of the benchmark or a source.} Counts cover the core six, and aerial counts are grounded misses on the 48 closed-form items; Appendix~\ref{app:failure-modes} defines each pattern, gives examples, and adds two more.}\label{tab:failure-modes}
\vspace{1pt}
\begin{tabular}{@{}>{\raggedright\arraybackslash}p{2.3cm} >{\raggedright\arraybackslash}p{5.0cm}@{}}
\toprule
Entry (task) & Evidence \\
\midrule
\grouprow{2}{Model failure patterns}
Median copying (allocation) & Open-weight forecasts copy the displayed analogue median on 284 and 259 of 300 items \\ %
Guessed date format (tool use) & 30 of the open-weight models' 35 failures filter dates in a guessed format \\ %
\grouprow{2}{Properties of the benchmark or a source}
Unit gloss (aerial, ours) & Our prompt calls the percent fields shares of pixels; 7 of the 9 misses the block added fit a fraction reading \\ %
Clipped maximum (aerial) & 15 of 34 misses, on frames clipped at 187.4\,$^\circ$C and labeled smoldering or fire-free \\ %
Ambiguous wording (aerial) & 9 of 34 misses fit an area reading of ``dominant fire behavior'' \\ %
Time-gap cue (smoke) & The reference precedes clear frames by 9--30 minutes and smoke frames by 44--80; a rerun matches the gaps \\ %
Unstated base rate (fire danger) & Subsampling sets a 0.339 positive rate that the benchmark prompts omit \\ %
\bottomrule
\end{tabular}
\vspace{-8pt}
\end{table}

\paragraph{I1: Task-Specific Grounding Gains Depend on Answer Access and Source Quality.}
These contrasts describe five specified additions; task and answer availability are not independently varied.
The read-only SQL tool raised all six core models to 0.885 or above, and the open-weight models' remaining failures were mostly guessed date formats (Appendix~\ref{app:tooluse-detail}). %
DORA reports a similar limit for disaster-response agents: gold tool-order hints raise final accuracy by only 1.08 to 4.40 points, since argument errors remain \citep{wang2026dora}. %
The thermal block raised accuracy on the 48 closed-form aerial items for all six core models and 14 of the sixteen full-capability models, and the proprietary four lost no closed-form item (Table~\ref{tab:aerial-exact}). %
Of the core six's 34 grounded misses there, 15 fall on the clipped frames of Section~\ref{subsec:source-audit}, and 9 fit an area reading of ``dominant fire behavior'' (Appendix~\ref{app:closed-form-misses}). %
Where an addition held no answer, scores moved in both directions, and the open-weight models copied the displayed analogue median, a rule that trails persistence (0.249 against 0.146 normalized error). %
Matched-gap smoke and movement-matched allocation reruns show how input design changes these contrasts (Section~\ref{subsec:results-tasks}).
On the 347 aerial items outside both Section~\ref{subsec:conditions-to-compare} groups, accuracy fell for all six core models. Neither Claude decline (unadjusted $p$ of 0.019 and 0.035) survives a Holm correction (Table~\ref{tab:aerial-exact}). %

\paragraph{I2: Zero-Shot Prompted Frontier Models Rarely Beat a Trivial or Trained Comparator.}
Prompted models rarely beat comparators (Section~\ref{subsec:results-tasks}): no core model beats persistence, only two core models beat the temperature rule, and no core model resolves above the aerial majority. %
On smoke, all twelve core-model arms beat the always-smoke constant on accuracy, and with gaps matched they also beat the frame-difference detector (Appendix~\ref{app:figlib-rerun}). %
All prompts were zero-shot, and no request set a reasoning parameter (Appendices~\ref{app:conditions-detail} and~\ref{app:reasoning}). %
Chain-of-thought reasoning improved wildfire risk prediction in FireScope \citep{markov2026firescope}, so this deficit bounds the tested zero-shot configurations rather than prompted models generally. %

\paragraph{I3: A Leaderboard Score Hides What a Fire Decision Needs: Calibration, Decision Rate, and Prompt Stability.}
On the selected fire-danger population, every core model stated probabilities less calibrated than a calendar-month prior fit on identically sampled data (Figure~\ref{fig:calibration}). %
Whether run and prior are shifted alike to the unstated 0.339 rate or the prompt states it, no proprietary run's Brier score is resolvably above the prior's (Tables~\ref{tab:recalibration} and~\ref{tab:baserate-rerun}). All four open-weight runs stay above. %
Under the reported scoring rule, two core models called fire on 4 and 6 percent of items against a 34 percent sample positive rate. Told the rate, they called 25 and 17 percent. %
claude-opus-5 omitted the call on 136 of its 772 answers (453 once told the rate), affecting only call-based scores (Appendix~\ref{app:recalibration}). %
Without a stated decision loss, call rates alone do not diagnose a poor decision. %
Two single-run paraphrases reordered standings, cutting Llama 4 Maverick by 0.136 AUPRC (Table~\ref{tab:prompt-sensitivity}). %

\paragraph{I4: A Read-Only SQL Tool Lifted Every Core Model, and Item Wording Changes How Far the Task Separates Them.}
The tool's gains measure access to the database the questions need (Section~\ref{subsec:results-tooluse}). %
Under original, reworded, and de-schematized items, the four proprietary models reach 0.994 to 1.000, too close for the task to rank them. The two open-weight models trail them under all three. %
Removing schema literals costs the two open-weight core models 0.175 and 0.126 on 143 clean items, which leave out one ambiguous question family. Only Qwen3-VL's widened interval excludes zero (Table~\ref{tab:tooluse-paraphrase}). %

\paragraph{I5: Public Fire Releases Carry Hazards That Change What a Score Measures.}
Three release hazards from Section~\ref{subsec:source-audit} change what scores measure (Appendix~\ref{app:defects}). A Mesogeos Track A feature separates holdout labels at average precision 1.000, and Fire360 questions map to no evaluation image. On 67 WildFireVQA fire frames, the labels follow a clipped thermal maximum. %

\paragraph{Future Directions.}
Benchmarks should pair bare and grounded variants across model families and report answer-carrying items and named failure modes separately. %
Every task needs trivial and trained comparators on the same items, and rankings need calibration, call rates, prompt sensitivity, and clustered uncertainty. %
Fire agents should use tools over public fire databases, with execution failures scored beside accuracy. %
Public releases should be audited for leakage, broken links, and labels computed from clipped thermal values, and ship item manifests, raw responses, and scoring code. %

\section{Conclusion}\label{sec:conclusion}

We present AI4Fire, which tests the same models bare and grounded across five wildfire tasks, since a result on one task did not predict the next. Six core models and a 29-model sweep give 77,644 scored responses on 1,446 paired items, beside a survey of 138 works. %
A read-only SQL tool lifted every core model to 0.88 or above on database questions, and Table~\ref{tab:failure-modes} records selected failure patterns and benchmark properties. Where an addition held no answer, scores moved in both directions, and two open-weight models copied an analogue median that forecasts worse than persistence. Zero-shot prompted models rarely beat a trivial rule or a trained comparator, and every task now carries one. %
A ranking hides calibration, decision rates, and paraphrase sensitivity. %
Re-deriving the public releases found hazards, among them a leaking column and labels computed from clipped thermal values, that change what a score on them measures. %
We release all prompts, item sets, harnesses, run manifests, and raw responses at \coderepo{} to support reproducible fire AI research. %
The central next step is to extend the paired design to multi-step agent tasks over the same public fire databases, scored beside trivial and trained comparators. %

\clearpage
\section*{Limitations}\label{sec:limitations}\label{app:limitations}\label{app:directions}\label{app:survey-directions}\label{subsec:data-check-directions}\label{app:results-directions-full}\label{subsec:benchmark-insights}%

\paragraph{Limits of the Survey.}\label{app:survey-limits}
The literature searches were not saturated, targeted works first released between 2025-01-01 and 2026-09-13, and did not systematically measure coverage outside English (Appendix~\ref{sec:method}). %
The claim-by-claim re-check covered a nonrandom set of 31 scoping-summary records, nine of its 25 fire records are preprints or organizational reports, and no search-unit tag was re-checked (Appendix~\ref{app:recheck}). %

\paragraph{Limits of the Benchmark.}\label{app:benchmark-limits}
Six core models cannot establish transfer to other families or generations, and weights are confounded with serving path (Appendix~\ref{app:modelset-detail}). Most configurations ran once, and the repeats show variation without estimating its full distribution (Appendix~\ref{app:prompt-sensitivity}). %
All prompts were zero-shot, and no request set a reasoning parameter (Appendix~\ref{app:reasoning}). We did not isolate explicit reasoning instructions or test trained pipelines such as FireScope \citep{markov2026firescope}. %
Each task's grounding addition differs and was untuned apart from the second allocation rule, so a null or negative effect bounds only that addition (Appendix~\ref{app:retrieval-v2}). %
The evaluated sources are publicly indexed, so training-data contamination remains unmeasured (Appendix~\ref{subsec:contamination-and-dating}). %

Fire-danger items carry Mesogeos Track A's negative-subsampled 0.339 positive rate, which the prompts omit, so results describe that population rather than regional prevalence (Appendix~\ref{app:defect-mesogeos}). %
The prior shift and the rerun that states the rate both use the sample's own rate, which a deployed forecaster would not know (Appendix~\ref{app:recalibration}). %
Aerial references are largely model-verified (276 of 408, with 124 below full applicability; Appendix~\ref{subsec:app-wildfirevqa-detail}). Fourteen hotspot references conflict with the prompt, and nine items use clipped-maximum frames, all scored as released (Appendix~\ref{app:defect-ceiling}). %
Our prompt glossed two percentage fields as shares of pixels, which may have invited the open-weight fraction readings (Appendix~\ref{app:defect-ours}). %
The smoke rerun with matched gaps and masked times covers the core six on 168 of the 196 paired frames, once per arm (Appendix~\ref{app:figlib-rerun}). Its references still precede the plume by about 40 minutes for clear targets and five for smoke targets. The sweep models and the 79-minute frames keep the first design, whose gap separates the labels. %
The tool-use questions are templated single-table queries on which the proprietary models reach the ceiling (Appendix~\ref{app:tooluse-detail}). No task has a human expert baseline, since Fire360 was withdrawn (Appendix~\ref{app:defect-fire360}). %

The five tasks omit structure fire, whose candidate datasets remain unreleased, restricted, or withdrawn, and evaluate no simulation-coupled multi-agent coordination (Appendix~\ref{subsec:left-out}). Unpinned proprietary snapshots may not reproduce these outputs (Appendix~\ref{app:sampling-serving}). Exact-match and numeric scoring leave judge error unmeasured and never check operational advice that could endanger firefighters (Appendix~\ref{app:scoring-detail}). %

\section*{Ethical Considerations}\label{sec:ethics}
All five task sources are publicly released, and Appendix~\ref{subsec:data-sources-licensing} records each license and access route. %
The FIgLib page names no license and requires credit to HPWREN, whose data-use page lists CC BY-NC-ND 4.0 for its data. %
Our release ships FIgLib sequence identifiers and a fetch script rather than raw frames. %
The study recruited no participants or annotators. %
Allocation records carry free-text weather notes from incident reports; a pattern screen of every released allocation file found no e-mail address, phone number, or titled personal name. %
The scores describe zero-shot behavior on past incidents and qualify no model for operational use. %
Staffing, fire-danger, or detection outputs acted on without expert review could endanger firefighters and the public. %

\bibliography{refs}

\clearpage
\appendix
\setcounter{table}{0}
\setcounter{figure}{0}
\renewcommand{\thetable}{A\arabic{table}}
\renewcommand{\thefigure}{A\arabic{figure}}
\renewcommand{\theHtable}{A\arabic{table}}
\renewcommand{\theHfigure}{A\arabic{figure}}
\renewcommand{\dbltopfraction}{0.9}
\renewcommand{\dblfloatpagefraction}{0.85}
\section{Search Protocol, Re-Check, and Category Boundaries}\label{app:protocol}

\subsection{How the Literature Was Collected}\label{sec:method}

Thirteen search units swept eight topic areas and five publication sources on 2026-09-13. %
Topics covered fire knowledge, vision, decision support, simulation, geospatial models, crisis communication, structure fire, and disaster surveys. %
Swept sources included machine learning venues, workshops, domain journals, ISCRAM, regional literature, and grey literature. %
A work was kept only if a fetched page confirmed it and its first public version appeared between 2025-01-01 and 2026-09-13. %
One work, \citet{tricomi2024ogcd123}, was kept although its Zenodo record is dated 2024, because an FGDC page dates it 2025-01-02. %
Under this rule, 138 works were kept (Figure~\ref{fig:survey-landscape}, Table~\ref{tab:collection}). %
An unchecked venue sweep found three NeurIPS 2025 Datasets and Benchmarks entries on fire perception that no topic search surfaced; they enter the survey as citations only \citep{liu2025detectiumfire,tiwari2025fire360,chen2025rscc}. %

\subsection{Profile of the 138 Kept Works}\label{app:collection}

Search units tagged the 138 kept works by kind and task category. %
A keyword screen finds an LLM, VLM, or agentic LLM framework in 109 of the 138 works, and 90 carry an evaluation. %
These provisional tags were never re-checked (Table~\ref{tab:collection}). %
The collection is released at \coderepo{} under \texttt{survey/}, where \texttt{survey/collection\_counts.py} reproduces the counts. %

\begin{table}[t]
\centering
\footnotesize
\setlength{\tabcolsep}{4pt}
\renewcommand{\arraystretch}{1.10}
\caption{\textbf{Search-unit tags over the 138 kept works: kind, task category, and, for the 90 works a keyword screen marks as LLM work with an evaluation, models tested.} Provisional counts that were never re-checked; a work can carry several task tags.}\label{tab:collection}
\begin{tabular}{@{}llr@{}}
\toprule
Grouping & Value & Works \\
\midrule
\grouprow{3}{Paper kind}
Kind & system-with-eval & 75 \\ %
Kind & evaluation & 21 \\
Kind & benchmark & 14 \\
Kind & survey & 14 \\
Kind & position & 8 \\
Kind & dataset & 5 \\
Kind & system-no-eval & 1 \\
\midrule
\grouprow{3}{Task category}
Task tag & decision-support-operations & 71 \\ %
Task tag & detection-perception & 52 \\
Task tag & geospatial-analysis & 37 \\
Task tag & document-understanding & 35 \\
Task tag & simulation-coupled-agents & 30 \\
Task tag & communication-alerts & 30 \\
Task tag & knowledge-qa & 26 \\
Task tag & forecasting-prediction & 20 \\
Task tag & data-retrieval-tools & 10 \\
Task tag & other & 5 \\
\midrule
\grouprow{3}{Models tested}
Models tested & unknown & 34 \\ %
Models tested & one & 12 \\
Models tested & 2--3 & 26 \\
Models tested & 4--9 & 16 \\
Models tested & 10 or more & 2 \\
\bottomrule
\end{tabular}
\end{table}

\subsection{Re-Check Outcomes and Corrections}\label{app:recheck}

A claim-by-claim re-check evaluated the 31 records cited by the initial project scoping summary. %
All 31 works exist, and ten carried at least one problem. The release lists each record's outcome and correction in \texttt{survey/PRIOR-ART.md}. %
Four held a claim the fetched page contradicted or lacked, five were partly supported, and six contained no fire task; these issue groups overlap. %
The 25 records that include a fire task carry the fire evidence of this survey. %
Of the six records without a fire task, five can be cited as adjacent disaster or emergency work, and one record has no citation key. %
The re-check made one error of its own, an undercounted model list for \citet{chen2025geospatial}, which a second fetch caught. %

\subsection{Category Boundaries and Open Questions}\label{app:taxonomy}

Ten working task categories organize the survey by task rather than domain or model family. %
Fire investigation \citep{tao2026magrfi} stays under document-understanding, code-generating Earth-observation agents \citep{lahouel2026iceo} stay under geospatial-analysis, and multi-hazard disaster suites \citep{zhang2026disasterbench,almohannadi2026disastervqa} count for their fire items. %

\subsection{Terms}\label{app:terms}\label{subsec:terms}

Grounding provides task data, retrieved documents, tools, or deterministic rules at inference time \citep{chen2025regcheck}. %
The bare arm supplies the task input alone, with the grounded arm's addition withheld. %
Wildfire and structure fire name the two fire domains. %

\subsection{Search-Unit Counts per Closest Work and Confidence in the Negative Answer}\label{app:trust-negative}\label{subsec:trust-negative}

A search unit is a targeted search run assigned to a distinct topic or publication source. %
Among the 13 units, five found \citet{qi2026smokebench}, four found \citet{ramesh2025wildfiregpt}, and three found \citet{chen2025geospatial}. %
Two units each found \citet{hyun2025crewwildfire} and \citet{zhang2026disasterbench}. %
One unit each found \citet{tricomi2024ogcd123}, \citet{ogc2026genaiwildfire}, \citet{lahouel2026iceo}, \citet{cheng2026retrieval}, and \citet{habibpour2026wildfirevqa}. %
Works found by several independent units support the gap, while single-unit finds show that the searches were not saturated. %

\begin{table*}[t]
\centering
\footnotesize
\setlength{\tabcolsep}{4pt}
\renewcommand{\arraystretch}{1.10}
\caption{\textbf{Our searches found no work that both surveys LLMs and agents across fire tasks and evaluates one model set under one protocol.} Closest works found and what each leaves undone relative to this paper.}\label{tab:closest}
\vspace{1pt}
\begin{tabular}{@{} >{\raggedright\arraybackslash}p{1.20in} >{\raggedright\arraybackslash}p{2.42in} >{\raggedright\arraybackslash}p{2.42in} @{}}
\toprule
Work & What It Does & What It Leaves Undone \\
\midrule
\citet{hyun2025crewwildfire} & Benchmarks four multi-agent frameworks using GPT-4o in a procedurally generated wildfire environment. & Evaluates one task family (operations in simulation), tests one base model, and provides no literature survey. \\
\addlinespace
\citet{zhang2026disasterbench} & Evaluates 21 baseline MLLMs plus DisasterVL on nine UAV disaster-response tasks, including wildfire propagation. & Focuses on general disasters using UAV imagery only; contains no survey of fire literature. \\
\addlinespace
\citet{chen2025geospatial} & Evaluates about 14 LLM configurations on wildfire personnel and cost allocation, with and without geospatial grounding. & Covers only two related decision tasks. \\
\addlinespace
\citet{tricomi2024ogcd123} & Formulates domain adaptation, National Incident Management System and Incident Command System (NIMS/ICS) operational lifecycle mapping, and governance frameworks for generative AI and agent workflows in wildland fire management. & No model evaluation across tasks. \\
\addlinespace
\citet{ogc2026genaiwildfire} & Assesses over 200 Canadian wildfire datasets for generative AI readiness and outlines prototype architectures for LLMs and agents. & No model evaluation across tasks. \\
\addlinespace
\citet{xu2026wildfirefm} & Compares a wildfire foundation model with ten Earth foundation models across six task forms (occupancy, spread, burned area, analog retrieval, smoke PM$_{2.5}$, and extreme heat) under a fixed evaluation contract. & No LLMs or agents. \\
\addlinespace
\citet{tran2025land8fire} & Combines a review with a human-annotated multispectral dataset and a benchmark. & Benchmarks one task, segmentation. \\
\addlinespace
\citet{zhao2025tssatfire} & Offers a satellite image time-series dataset, evaluates 11 models, and includes a background review of prior methods. & Covers three tasks: detection, burned area, and next-day progression. \\
\bottomrule
\end{tabular}
\end{table*}

\section{Data Availability Check}\label{app:dataaudit}

The data availability check evaluated 34 candidate tasks and sources across the working task categories of Section~\ref{sec:scope} (Table~\ref{tab:dataaudit}). %
Of the 34 candidates, 3 are ready, 20 need work, and 11 are not feasible now, with unreleased data explaining 7 infeasible candidates. %
Ready means reachable and scorable as released. Needs work means the data exists but items or a harness must be built first. Not feasible means no route exists without a separate data agreement, a new license, or expert annotation. %

\begin{table*}[t]
\centering
\footnotesize
\setlength{\tabcolsep}{3pt}
\renewcommand{\arraystretch}{1.10}
\caption{\textbf{Of 34 checked candidate tasks and sources, three are ready, 20 need work, and 11 are infeasible primarily from lack of released data.} The 34 task and source candidates checked on 2026-09-15, with category, fire domain, verdict, access route, and license. \emph{Task or Source} names the candidate; \emph{Category} gives the working task category; \emph{Domain} follows the wildfire and structure split; \emph{Verdict} marks readiness; \emph{Access} notes availability; \emph{License} records redistribution terms.}\label{tab:dataaudit}
\vspace{1pt}
\begin{tabular}{@{} >{\raggedright\arraybackslash}p{1.40in} >{\raggedright\arraybackslash}p{1.05in} >{\raggedright\arraybackslash}p{0.55in} >{\raggedright\arraybackslash}p{0.62in} >{\raggedright\arraybackslash}p{1.10in} >{\raggedright\arraybackslash}p{1.13in} @{}}
\toprule
Task or Source & Category & Domain & Verdict & Access & License \\
\midrule
Evacuation guidance dialog, BEACON & Communication and Alerts & wildfire & not feasible now & none found; Watch Duty feed closed to automated access & paper CC BY 4.0; no data license \\ %
Crisis translation and urgency classification & Communication and Alerts & both & needs work & public download for three of four corpora & CC BY 4.0 on the one fire-bearing multilingual corpus \\ %
Fire database and tool use, FPA-FOD and public APIs & Data Retrieval and Tool Use & both & needs work & public download or free API for all five sources & usable without additional permissions or fees \\ %
Fire radiative power forecasting, FIRMS & Forecasting and Prediction & wildfire & needs work & paper table unreleased; FIRMS behind a free key & NASA open data; no license on the paper table \\ %
Wildfire risk rasters, FireScope-Bench & Forecasting and Prediction & wildfire & needs work & public download, 127 GB & CC BY 4.0 plus upstream provider terms \\ %
Risk sub-criteria scoring against expert ranks & Forecasting and Prediction & wildfire & needs work & public repository download, no registration & article CC BY 4.0; repository states none \\ %
Burn-scar and damage assessment, HLS Burn Scars & Geospatial Analysis & wildfire & needs work & IC-EO items none found; HLS set public & HLS Burn Scars CC BY 4.0 \\ %
Next-day spread, WildfireSpreadTS \citep{gerard2023wildfirespreadts} & Forecasting and Prediction & wildfire & needs work & public Zenodo download, 48.4 GB & CC BY 4.0 \\ %
Multi-task time series, TS-SatFire & Forecasting and Prediction & wildfire & needs work & Kaggle download with a free account, 76 GB & CC BY 4.0 \\ %
Multispectral segmentation, Land8Fire & Detection and Perception & wildfire & needs work & public download from an unversioned host, 55 GB & article CC BY 4.0; repository states none \\ %
Fire danger forecasting, Mesogeos & Forecasting and Prediction & wildfire & needs work & public download, no registration & CC BY 4.0 \\ %
Sub-seasonal modeling, SeasFire cube \citep{karasante2025seasfire} & Forecasting and Prediction & wildfire & needs work & public Zenodo download, 43.9 GB & CC BY 4.0 \\ %
\bottomrule
\end{tabular}
\end{table*}

\begin{table*}[t]
\centering
\ContinuedFloat
\footnotesize
\setlength{\tabcolsep}{3pt}
\renewcommand{\arraystretch}{1.10}
\caption{(continued)}
\vspace{1pt}
\begin{tabular}{@{} >{\raggedright\arraybackslash}p{1.40in} >{\raggedright\arraybackslash}p{1.05in} >{\raggedright\arraybackslash}p{0.55in} >{\raggedright\arraybackslash}p{0.62in} >{\raggedright\arraybackslash}p{1.10in} >{\raggedright\arraybackslash}p{1.13in} @{}}
\toprule
Task or Source & Category & Domain & Verdict & Access & License \\
\midrule
Foundation-model transfer tasks, WILDFIRE-FM & Forecasting and Prediction & wildfire & not feasible now & anonymous repository serves code, metadata, and paper-result files; the Hugging Face link returned HTTP 401 to anonymous requests on 2026-09-18; raw inputs and large feature arrays are excluded & paper CC0; code MIT; source data under provider terms \\ %
Fire code compliance, RegCheck audit set & Knowledge Question Answering & structure fire & needs work & audit set none found; code text public & audit set none stated; Approved Document B under OGL v3.0 \\ %
Fire science and safety question answering & Knowledge Question Answering & both & needs work & public download for all four source corpora & OGL v3.0 and CC BY 4.0 \\ %
Barriers to fire spread, WFDSS incident text & Document Understanding & wildfire & not feasible now & restricted for sensitivity & article CC BY-NC-ND 4.0; no data license \\ %
Incident-command transcript task state & Document Understanding & structure fire & needs work & public repository download, no registration & none stated on the repository \\ %
Bushfire damage assessment reports & Document Understanding & wildfire & not feasible now & no dataset released & code CC BY-NC-ND 4.0, which blocks derivatives \\ %
Fire protection specification drafting, FireSpec & Document Understanding & structure fire & not feasible now & paper paywalled; nothing released & Korean standards terms unverified \\ %
Fire investigation reasoning, MAGR-FI cases & Document Understanding & structure fire & needs work & 1,051 cases none found; NIOSH reports public & cases none stated; NIOSH reports public domain \\ %
Daily personnel and cost, ICS-209-PLUS & Decision Support and Operations & wildfire & needs work & public figshare download; derived set not released & CC BY 4.0 on figshare; GAL code MIT \\ %
\bottomrule
\end{tabular}
\end{table*}

\begin{table*}[t]
\centering
\ContinuedFloat
\footnotesize
\setlength{\tabcolsep}{3pt}
\renewcommand{\arraystretch}{1.10}
\caption{(continued)}
\vspace{1pt}
\begin{tabular}{@{} >{\raggedright\arraybackslash}p{1.40in} >{\raggedright\arraybackslash}p{1.05in} >{\raggedright\arraybackslash}p{0.55in} >{\raggedright\arraybackslash}p{0.62in} >{\raggedright\arraybackslash}p{1.10in} >{\raggedright\arraybackslash}p{1.13in} @{}}
\toprule
Task or Source & Category & Domain & Verdict & Access & License \\
\midrule
Operations by incident-command phase, OGC 24-071 & Decision Support and Operations & wildfire & not feasible now & report public; no evaluation items in it & OGC document license; NWCG source text public domain \\ %
Risk and insurance retrieval, OGC 25-012 & Decision Support and Operations & wildfire & not feasible now & report public; no evaluation items in it & OGC document license; no license given for listed sources \\ %
Suppression and rescue, CREW-Wildfire & Simulation-Coupled Agents & wildfire & ready & public download; environment and scoring released & code Apache-2.0; paper CC BY-NC-ND 4.0 \\ %
Policy adaptation, MOASEI and free-range-zoo & Simulation-Coupled Agents & wildfire & needs work & public download; frozen evaluation tag & MIT on the environment; AGPL-3.0 on the starter \\ %
Zero-shot active fire classification, Roboflow set & Detection and Perception & wildfire & not feasible now & Roboflow download with a free account & CC BY 4.0 \\ %
Smoke classification and localization, SmokeBench & Detection and Perception & wildfire & needs work & none found for the benchmark; FIgLib images public & preprint CC BY 4.0; benchmark none \\ %
Temperature-grounded aerial questions, WildFireVQA & Detection and Perception & wildfire & ready & questions public; FLAME 3 images need a free account & Apache-2.0 on the dataset card \\ %
\bottomrule
\end{tabular}
\end{table*}

\begin{table*}[t]
\centering
\ContinuedFloat
\footnotesize
\setlength{\tabcolsep}{3pt}
\renewcommand{\arraystretch}{1.10}
\caption{(continued)}
\vspace{1pt}
\begin{tabular}{@{} >{\raggedright\arraybackslash}p{1.40in} >{\raggedright\arraybackslash}p{1.05in} >{\raggedright\arraybackslash}p{0.55in} >{\raggedright\arraybackslash}p{0.62in} >{\raggedright\arraybackslash}p{1.10in} >{\raggedright\arraybackslash}p{1.13in} @{}}
\toprule
Task or Source & Category & Domain & Verdict & Access & License \\
\midrule
Satellite risk reasoning, Wildfire-Smoke-RS & Detection and Perception & wildfire & not feasible now & none found; the repository holds only a README & none stated \\ %
UAV disaster reasoning, DisasterBench & Detection and Perception & both & needs work & public download of the test split & Apache-2.0 \\ %
Smoke sequences, FIgLib & Detection and Perception & wildfire & ready & public download, no registration & FIgLib page names no license and requires credit; HPWREN data-use page lists CC BY-NC-ND 4.0 \\ %
Fire-scene severity items, DetectiumFire & Detection and Perception & both & not feasible now & Kaggle download with a free account & two authoritative sources disagree, both non-commercial \\ %
Structure-fire video questions, Fire360 & Detection and Perception & structure fire & needs work & public Box download, no login & paper: MIT with research-only restrictions; released license file: standard MIT text \\ %
Post-fire change captions, RSCC & Detection and Perception & wildfire & not feasible now & public download, not gated & paper: CC BY 4.0; repository binds the xBD-derived subset to the xBD terms and research use \\ %
\bottomrule
\end{tabular}
\end{table*}

\section{Defects Found While Building the Tasks}\label{app:defects}

Building benchmark tasks revealed hazards in three released datasets, changing what scores measure. %
None claims published evaluations are wrong. %
For the Mesogeos, Fire360, and closed-form findings, an agent that did not inspect the task builders re-derived each from raw source files. A third check repeated each from raw sources, and Appendix~\ref{app:defect-ceiling} states how the clipping was checked. %

\begin{table*}[t]
\centering
\footnotesize
\setlength{\tabcolsep}{4pt}
\renewcommand{\arraystretch}{1.10}
\caption{\textbf{Source hazards found while building the benchmark, in three other releases and in ours.}}\label{tab:source-audit}
\vspace{1pt}
\begin{tabular}{@{} >{\raggedright\arraybackslash}p{1.10in} >{\raggedright\arraybackslash}p{2.45in} >{\raggedright\arraybackslash}p{2.45in} @{}}
\toprule
Source & What reproduces & What it does not establish \\
\midrule
Mesogeos Track A & A released static column separates the holdout at average precision 1.000; a rule over two window columns reaches 0.814 on the last day & That the published baselines read those columns. All three released configurations name their inputs and exclude them \\
Fire360 & The released 100 questions carry no reference to a video, a frame, or a timestamp, and the answer key puts the correct choice first on 25 of 50 & That the released items are the scored ones. The release calls them sample prompts, and no mapping ties them to the reported evaluation \\
WildFireVQA & Four question types are reproduced exactly by closed-form rules on the supplied thermal block, on 6,097 of 6,097 frames each. On 67 frames filed as fire, the maximum is clipped at 187.4\,$^\circ$C, and every temperature-derived label reads smoldering or no fire & That the block is illicit, that a model missing these labels ignored the block, or where the clipping arises. The source sets out to study thermal retrieval; what is missing is a text-only arm beside the pooled score \\
This benchmark & A few-hundred-token output cap truncated a reasoning model's answer on 40 of 196 items, against 8 for another family & That the affected runs measure model skill. The cap selected which items each family was scored on \\
\bottomrule
\end{tabular}
\end{table*}

Appendix~\ref{app:defect-ours} documents the hazards in our own harness. %

\subsection{The Mesogeos Track A Release Ships Columns That Separate Its Own Holdout}\label{app:defect-mesogeos}

The released Track A files carry three columns describing the label \citep{kondylatos2023mesogeos}. %
On day 30 of the 30-day window, \texttt{burned\_areas} is nonzero for 67.06 percent of test positives and 0 percent of test negatives. %
The column \texttt{ignition\_points} is nonzero for 26.15 percent of test positives and 0 percent of negatives. %
Scoring an item by whether either column is nonzero on the last day gives average precision 0.814 at zero false positives. %
Static \texttt{burned\_area\_has} is nonzero for 100 percent of positives and 0 percent of negatives, separating the holdout at average precision 1.000. %
Evaluations on Track A must withhold these three columns and supply only the 24 runnable meteorological and static drivers. %
Unseeded negative subsampling yields 17,342 negatives against 17,148 reported, and the 2021--2022 holdout holds 4,120 rows (1,369 positives, 2,751 negatives) against 4,107 reported (2,738 negatives). %
Across those holdout negatives, zero ignitions occurred in the cell or adjacent cells on the target date. %
An independent check that did not inspect the task builders re-derived these separations and row counts from raw source files. %

\subsection{The Released Fire360 Questions Cannot Be Matched to the Reported Evaluation}\label{app:defect-fire360}

The public Box repository ships 100 text prompts: 50 multiple-choice questions with an answer key and 50 free-text questions \citep{tiwari2025fire360}. %
None of the 100 prompts references a video file, frame number, or timestamp. %
The answer key has option 1 correct on 25 of 50 items and option 4 on zero, scoring 50.0 percent. %
None of the five reported accuracies (53.8, 50.3, 47.2, 42.7, and 91.4 percent for humans) is a multiple of $100/50$ or $100/100$. %
Because the release provides no verifiable mapping between text prompts and video frames, the released items cannot support a grounded perception evaluation. %
Direct inspection of released Box files confirmed these omissions and distributions. %

\subsection{Rules on the Thermal Block Reproduce Four WildFireVQA Question Types}\label{app:defect-wildfirevqa}

The WildFireVQA prompt includes a thermal summary block with radiometric statistics \citep{habibpour2026wildfirevqa}. %
Fixed rules on this block reproduce released labels for four question types (CL1, CMR4, DS7, DS8) without imagery on all 6,097 frames (24,388 items). %
The rules cut the maximum at 50, 80, and 200\,$^\circ$C for CL1 and read any maximum below 200\,$^\circ$C as no fire detected for CMR4. On DS7 and DS8 they answer None only when no pixel passes the threshold. %
Neither the source prompt template nor our prompt states these cuts, so a model that reads the block correctly can still disagree with the label (Appendix~\ref{app:closed-form-misses}). These four types make up 48 of the 408 benchmark items. The closed-form rule reproduces the label of all 48, and Section~\ref{subsec:results-wildfirevqa} reports them apart from the other question types. %
Class imbalance adds a majority baseline of 0.627 across the 408 items; the vehicle-count question matches on 99.66 percent of frames. %
An independent check verified that these rules reproduce all 24,388 labels across the released frames. %

\subsection{A Clipped Thermal Maximum on 67 WildFireVQA Fire Frames}\label{app:defect-ceiling}

Nine question types derive from temperatures: CL1, CMR4, DS1, DS3, DS7, DS8, LD1, PD1, and PD7. %
Across 2,676 frames with maximum below 200\,$^\circ$C, none reports fire or a hotspot. %
On 67 of 738 Sycan Marsh frames the maximum is exactly 187.424\,$^\circ$C, yet FLAME~3 files all 67 under Fire (65 in Sycan 2D, 2 in Sycan 2A). %
These 67 frames have a median 83 pixels at maximum against 1 on other 671 frames. %
Fifty-two have more pixels at maximum than in the band from 0.5 to 1.5\,$^\circ$C below it. %
No Sycan frame has maximum between 187.424 and 200\,$^\circ$C, and 48 share a minimum of $-40.559$\,$^\circ$C, indicating thermal clipping without establishing its exact source. %
The release labels all 67 smoldering on CL1, no fire on CMR4, None on DS7/DS8, No on PD1, No Fire on PD7, and no hotspots on DS1/DS3/LD1 (603 records total). %
Nine of our 408 items evaluate these frames (five closed-form, four no-fire-branch); visual check confirms smoke plumes and burning lines. %
The core six miss 15 of the 30 grounded answers on the five closed-form items and all 24 on the four no-fire-branch items (Appendix~\ref{app:closed-form-misses}). %
Across the release, 885 of 1,285 Shoetank fire frames share maximum 205.72\,$^\circ$C (labeled 200--300 on CMR4, None on DS7), and 608.69\,$^\circ$C recurs on 160 frames. Our benchmark samples no Shoetank frame. %
An independent script verified every count above directly from raw TIFFs and release files. %

\subsection{Fire360 as a Withdrawn Task}\label{app:defect-withdrawn}

Fire360 was selected for expert questions, reported human baselines, and an open research license \citep{tiwari2025fire360}. %
Withdrawing Fire360 leaves the benchmark with five wildfire tasks and no structure-fire arm, as other candidate sets remain unreleased or restricted (Appendix~\ref{app:dataaudit}) \citep{chen2025regcheck,tao2026magrfi,kim2025firespec,grunert2026taskstate}. %

\section{Benchmark Design Detail}\label{app:benchdetail}\label{subsec:benchmark-purpose}
This appendix records benchmark design, candidate selection, grounding sources, licensing, and scoring hazards. %

\subsection{Selection Criteria and What They Selected}\label{subsec:selection-criteria}
Five criteria applied to 34 candidates (Appendix~\ref{app:dataaudit}): domain coverage, reference labels, verifiable terms, datable items, and cost. %
Five sources survived: FIgLib, WildFireVQA, Mesogeos, ICS-209-PLUS, and FPA-FOD. %
Fire360 was withdrawn after audit found no verifiable prompt-to-frame mapping (Appendix~\ref{app:defect-fire360}). %

\subsection{The Grounding Source of Each Task}\label{app:conditions-detail}
\begin{itemize}
  \item \textbf{Bare Model:} Zero-shot across all five tasks: fixed-format system message and item user message; no worked examples or reasoning parameters. %
  \item \textbf{Grounded, Smoke Detection:} The grounded arm provides an earlier clear frame from the same camera; fire-named paths are withheld to prevent label leakage. %
  FIgLib provides frames as-is; HPWREN lists CC~BY-NC-ND~4.0. %
  \item \textbf{Grounded, Aerial Question Answering:} Grounded prompts append sensor-derived \texttt{temp\_summary} statistics from WildFireVQA \citep{habibpour2026wildfirevqa}: min, max, mean, std, hottest 3 percent pixel mean, and shares exceeding 200 and 400~$^\circ$C. %
  Fixed rules on this block reproduce the labels of four of the 34 question types (48 of 408 items), given thresholds the prompt does not state (Appendix~\ref{app:defects}). %
  \item \textbf{Grounded, Fire Danger Forecasting:} Grounded prompts append monthly climatology from 2006--2019 training years beside 24 runnable meteorological and static drivers \citep{kondylatos2023mesogeos}. %
  \item \textbf{Grounded, Allocation:} Retrieved analogues supply earlier fire-days matched on size, containment, and staffing, reporting next-day staffing outcomes (sharing no incidents with evaluation items). %
  Analogue input and outcome days must both precede target report days. %
  Filtering input day alone originally let one item of 300 see a same-day outcome. %
  That item's grounded prompt was re-queried under the final rule on the five models that had run before it, which changes no other item's draw. gpt-6-astra ran under the final rule from the start. %
  \item \textbf{Grounded, Tool Use:} Bare models answer from parameters given the schema. %
  The grounded arm queries a frozen FPA-FOD SQLite snapshot (214~MB) through read-only SQL tool \texttt{query\_fpafod(sql)}. %
  Queries return up to 50 rows and 4,000 characters (at most 8 calls per question; Appendix~\ref{app:tooluse-detail}). %
\end{itemize}

\subsection{What the Benchmark Leaves Out}\label{subsec:left-out}
Structure fire is left out, and no redistribution route for NFPA or ICC code text was open under checked permissions. Two routes to it remain open: compliance on England Approved Document B (via CODE-ACCORD under CC BY 4.0) and causal reasoning over NIOSH firefighter fatality reports. %
Multi-agent suppression in CREW-Wildfire \citep{hyun2025crewwildfire} was deferred due to unmeasured trajectory costs. %
Eleven candidate sources were infeasible due to unreleased data, restrictive licenses, ungrounded references, or raster processing (Appendix~\ref{app:dataaudit}). %

\subsection{The Model Set in Full}\label{app:modelset-detail}
The benchmark evaluates six core models from five vendors on two serving paths. %
Four proprietary models (\texttt{claude-opus-4.8}, \texttt{claude-opus-5}, \texttt{gemini-3.1-pro}, \texttt{gpt-6-astra}) ran through an API gateway; two open-weight models (\texttt{qwen.qwen3-vl-\allowbreak 235b-a22b}, \texttt{us.meta.llama4-maverick-\allowbreak 17b-instruct-v1:0}) ran through Bedrock. %
Requests specified temperature 0.0 except \texttt{gpt-6-astra}, whose endpoint fixes 1.0; output cap was 1,536 tokens. Effective gateway settings were not verified (Appendix~\ref{app:reasoning}).
Sweep models ran through Bedrock under identical settings, except three reasoning models at 8,192 tokens (Appendix~\ref{app:model-sweep}). %

\subsection{The Model Sweep of 2026-09-18}\label{app:model-sweep}
\begin{figure*}[tp]
\centering
\includegraphics[width=\textwidth]{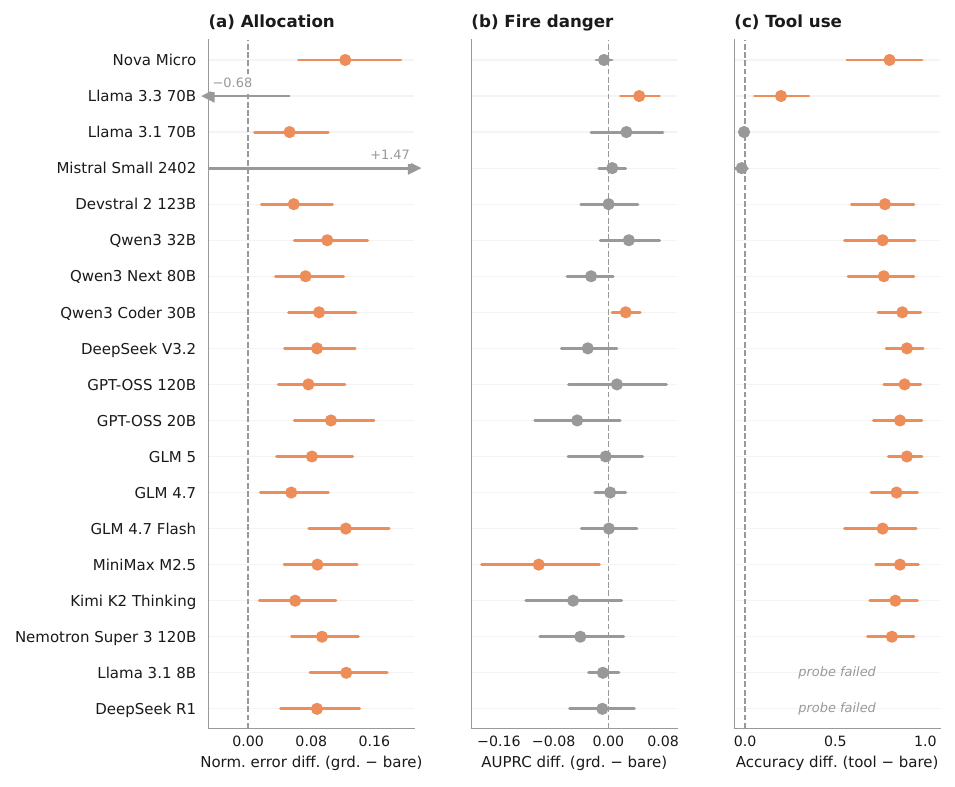}
\caption{\textbf{Across the nineteen text-only models, intervals exclude zero for the analogues' error increase on seventeen and the SQL tool's lift on fifteen of seventeen; climatology's AUPRC change excludes zero on three, in both directions.} Paired grounded-minus-bare differences (tool minus bare in c) with 95 percent cluster-bootstrap intervals, clustered as in Table~\ref{tab:cross-task-summary}, on the items each model parsed in both arms (in c, all 156 items, with an unanswered response counted wrong); coral intervals exclude zero and gray ones include it. (a)~Allocation: normalized-error difference. Llama~3.3~70B ($-0.68$) and Mistral Small 2402 ($+1.47$) lie beyond the axis, each moved by one extreme prediction (Table~\ref{tab:text-sweep}; Appendix~\ref{subsec:app-sweep-detail}), and both intervals contain zero. (b)~Fire danger: average precision difference; MiniMax M2.5's interval lies below zero. (c)~Tool use: accuracy difference; Llama~3.1~8B and DeepSeek~R1 failed the tool probe. Figure~\ref{fig:grounding-sweep} shows the sixteen full-capability models.}\label{fig:text-sweep}
\end{figure*}

A probe sent each of the 33 chat models reachable on Bedrock three requests: text, a 64-pixel image with a colour question, and one mock tool to call. Passing the tool probe required a native tool-use block. %
Two of the 33 are core models. Seven added models passed all three probes: Nova Lite, Nova Pro, Nova 2 Lite, Llama 4 Scout, Mistral Large 3, Ministral 3 8B, and Kimi K2.5. %
The three Gemma 3 sizes accept images but answer the tool probe with a fenced \texttt{tool\_code} block, Gemma's calling convention, and ran through the adapter described below. These ten joined the six core models on all five tasks. %
Twenty models rejected image input, and Mixtral 8x7B, which also rejects system messages, left the sweep. The other nineteen are text-only (Table~\ref{tab:text-sweep}): all ran allocation and fire danger, and the seventeen whose tool probe passed ran tool use. %
Llama 3.1 8B answered the probe with Python source instead of a call, and DeepSeek R1 rejects tool declarations. Palmyra Vision 7B rejects any tool declaration and was not run. %
The 29 added models ran at temperature 0.0 under a 1,536-token cap. Ten-item pilots at that cap truncated the answers of MiniMax M2.5, Kimi K2 Thinking, and DeepSeek R1, which therefore ran at 8,192 tokens. %
Even at 8,192 tokens, MiniMax M2.5 truncated 261 responses. Its allocation error rests on the 278 bare and 296 grounded predictions that parsed, and its fire-danger scores on 312 bare and 300 grounded answers. %
A score on the parsed subset can sit above or below the score on the complete item set. Tool accuracy uses all 156 items and counts an unanswered response as wrong. Every other sweep model marked at most 17 responses as truncated. %
Cluster bootstrap intervals for the text-only models use the items parsed in both arms (Figure~\ref{fig:text-sweep}). For MiniMax M2.5 that leaves 274 allocation items and 246 fire-danger items, so its interval centers differ from the table. %
The adapter renders the tool declaration into the system prompt, parses the call from the reply, and returns the result in a \texttt{tool\_output} block, with the same eight-call budget. Gemma 3 27B followed the convention on all 156 items (0.827). %
Gemma 3 12B and 4B produced no call the harness could execute and score 0.000 as interface failures. %
On Bedrock, a request after an eighth tool call omitted the tool declaration and was rejected. No core model reached that call. Nineteen rows of three added models (Nova 2 Lite 12, Ministral 3 8B 5, Kimi K2.5 2) failed there and were re-queried once the harness rendered that exchange as text. %
Bare smoke false positives span 0.000 to 0.119 (Gemma 3 27B), rising with grounding on 10 of sixteen models. %
The four highest grounded false-positive rates are Gemma 3 27B (0.250), 12B (0.202), 4B (0.179), and Mistral Large 3 (0.157). %
Bare fire-call rates span 0.044 (\texttt{gpt-6-astra}) to 1.000 (Gemma 3 12B and 4B) against the 0.339 sample positive rate. %
Five models call fire on most items: Gemma 3 27B (0.909), Mistral Large 3 (0.837), Ministral 3 8B (0.772), Qwen3-VL (0.684), and Llama 4 Scout (0.648). %
Across all 35 models, five return the previous day's count on at least 290 of the 300 bare allocation items. %
Counts reach 299 for Qwen3 Coder 30B and Gemma 3 4B, 297 for Qwen3 32B, and 294 for Nemotron Super 3 120B and GLM 4.7 Flash. %

\begin{table*}[t]
\centering
\footnotesize
\setlength{\tabcolsep}{4pt}
\renewcommand{\arraystretch}{1.08}
\caption{\textbf{The nineteen text-only models on the three text tasks.} Point estimates from single runs at temperature zero on the same items and scorers as the main tables; the cap column is the output cap in tokens. Allocation: nMAE (persistence 0.146). Fire danger: AUPRC (temperature rule 0.654), call rate (sample positive rate 0.339). Tool use: accuracy bare and with the tool (best constant per family 0.141); a dash marks the two models whose tool probe failed. MiniMax M2.5 covers 278 of 300 bare and 296 of 300 grounded allocation items, and 312 of 386 bare and 300 of 386 grounded fire-danger items. Persistence on those allocation subsets is 0.143 and 0.146; tool accuracy counts all 156 items. Mistral Small 2402's grounded error is 1.958, and 0.520 without its 866-person prediction for a two-person incident. Gemma 3 4B, in Table~\ref{tab:model-sweep}, returns the previous day's count on 299 of 300 bare items.}\label{tab:text-sweep}
\vspace{1pt}
\begin{tabular}{@{}l r cc cc cc cc@{}}
\toprule
& & \multicolumn{2}{c}{Allocation nMAE} & \multicolumn{2}{c}{Fire danger AUPRC} & \multicolumn{2}{c}{Fire-call rate} & \multicolumn{2}{c}{Tool use acc.} \\
\cmidrule(lr){3-4}\cmidrule(lr){5-6}\cmidrule(lr){7-8}\cmidrule(lr){9-10}
Model & Cap & bare & grd. & bare & grd. & bare & grd. & bare & tool \\
\midrule
Nova Micro & 1,536 & 0.166 & 0.289 & 0.339 & 0.332 & 0.000 & 0.000 & 0.019 & 0.821 \\ %
Llama 3.3 70B & 1,536 & 0.927 & 0.248 & 0.413 & 0.458 & 0.446 & 0.440 & 0.090 & 0.288 \\ %
Llama 3.1 70B & 1,536 & 0.205 & 0.257 & 0.400 & 0.427 & 0.145 & 0.202 & 0.006 & 0.000 \\ %
Mistral Small 2402 & 1,536 & 0.491 & 1.958 & 0.401 & 0.407 & 0.858 & 0.671 & 0.058 & 0.038 \\ %
Devstral 2 123B & 1,536 & 0.179 & 0.237 & 0.522 & 0.522 & 0.681 & 0.674 & 0.096 & 0.872 \\ %
Qwen3 32B & 1,536 & 0.147 & 0.247 & 0.429 & 0.458 & 0.782 & 0.821 & 0.077 & 0.840 \\ %
Qwen3 Next 80B & 1,536 & 0.153 & 0.226 & 0.416 & 0.391 & 0.811 & 0.808 & 0.083 & 0.853 \\ %
Qwen3 Coder 30B & 1,536 & 0.151 & 0.240 & 0.383 & 0.409 & 0.974 & 0.943 & 0.045 & 0.917 \\ %
DeepSeek V3.2 & 1,536 & 0.167 & 0.255 & 0.469 & 0.439 & 0.668 & 0.653 & 0.103 & 1.000 \\ %
GPT-OSS 120B & 1,536 & 0.166 & 0.242 & 0.448 & 0.460 & 0.606 & 0.435 & 0.051 & 0.936 \\ %
GPT-OSS 20B & 1,536 & 0.166 & 0.271 & 0.438 & 0.392 & 0.412 & 0.252 & 0.058 & 0.917 \\ %
GLM 5 & 1,536 & 0.189 & 0.270 & 0.570 & 0.566 & 0.280 & 0.311 & 0.103 & 1.000 \\ %
GLM 4.7 & 1,536 & 0.184 & 0.239 & 0.494 & 0.496 & 0.500 & 0.495 & 0.128 & 0.968 \\ %
GLM 4.7 Flash & 1,536 & 0.146 & 0.270 & 0.444 & 0.445 & 0.259 & 0.345 & 0.077 & 0.840 \\ %
MiniMax M2.5 & 8,192 & 0.156 & 0.244 & 0.574 & 0.467 & 0.638 & 0.577 & 0.026 & 0.885 \\ %
Kimi K2 Thinking & 8,192 & 0.193 & 0.253 & 0.560 & 0.511 & 0.650 & 0.507 & 0.058 & 0.891 \\ %
Nemotron Super 3 120B & 1,536 & 0.150 & 0.243 & 0.529 & 0.490 & 0.306 & 0.508 & 0.096 & 0.910 \\ %
Llama 3.1 8B & 1,536 & 0.175 & 0.300 & 0.334 & 0.325 & 0.000 & 0.000 & -- & -- \\ %
DeepSeek R1 & 8,192 & 0.173 & 0.260 & 0.472 & 0.463 & 0.899 & 0.868 & -- & -- \\ %
\bottomrule
\end{tabular}
\end{table*}

\subsection{Data Sources and Licensing}\label{subsec:data-sources-licensing}\label{subsec:contamination-and-dating}\label{subsec:human-expert-baselines}
Each data source was verified on its distribution portal: %
\begin{itemize}[leftmargin=*,noitemsep,topsep=2pt]
  \item \textbf{Authored Benchmark Release:} Harness, prompt templates, manifests, and scoring scripts carry the BSD 2-Clause License at \coderepo{}. %
  Stored model outputs travel under the terms of their input sources. %
  \item \textbf{FIgLib (Smoke Detection):} FIgLib is the HPWREN fire ignition image library \citep{dewangan2022figlib}. %
  Its page provides frames as-is and requires credit to HPWREN, whose portal lists CC BY-NC-ND 4.0. %
  We release sequence identifiers, labels, and fetch scripts rather than raw JPEGs; items span 2026-01-28 to 2026-09-09. %
  Public since 2020-08-18, the index runs from 2016-06-04 to 2026-09-09 with 79 sequences from 2025 and 2026. %
  We do not claim items postdate training data of evaluated models. %
  \item \textbf{WildFireVQA and FLAME 3 (Aerial QA):} WildFireVQA records Apache-2.0 in metadata and CC-BY-4.0 in card text, creating conflicting license statements. %
  FLAME 3 radiometric thermal and RGB imagery is CC BY 4.0 via IEEE DataPort \citep{habibpour2026wildfirevqa}. %
  \item \textbf{Mesogeos (Fire Danger):} CC BY 4.0 on Zenodo; items draw from the 2021--2022 temporal holdout using 24 runnable driver features \citep{kondylatos2023mesogeos}. %
  \item \textbf{ICS-209-PLUS (Allocation):} CC BY 4.0 on figshare; reports spanning 1999--2020, filtering analogues by strict temporal precedence \citep{stdenis2023ics209plus}. %
  \item \textbf{FPA-FOD and Public Services (Tool Use):} FPA-FOD is US government open data requiring citation \citep{short2022fpafod}. %
  NASA FIRMS and NWS provide free public APIs. %
  NIFC WFIGS perimeter layers are publicly accessible with a liability disclaimer and no explicit license. %
\end{itemize}

\subsection{Reference Answers and Scoring Hazards, by Task}\label{app:answers-hazards}
\begin{itemize}
  \item \textbf{Smoke Detection:} Reference: plume offset in seconds. Hazard: roughly half of each sequence is positive by construction, and frames near offset zero are visually ambiguous.
  \item \textbf{Aerial QA:} Reference: multiple choice. Hazard: thermal stats resolve 48 items in closed form; class skew creates a 0.627 majority baseline (Appendix~\ref{app:defects}).
  \item \textbf{Fire Danger:} Reference: binary presence in holdout. Hazard: negative skew; three columns carry the label and must be withheld.
  \item \textbf{Personnel Allocation:} Reference: filed staffing per fire-day. Hazard: heavy-tailed targets; filed numbers reflect staff availability alongside demand.
  \item \textbf{Tool Use:} Reference: deterministic execution over frozen SQLite snapshots. Hazard: endpoint drift requires snapshot pinning, and network errors must be distinguished from reasoning failures.
\end{itemize}

\begin{table*}[t]
\centering
\footnotesize
\setlength{\tabcolsep}{4pt}
\renewcommand{\arraystretch}{1.10}
\caption{\textbf{Every proprietary model underpredicts fire frequency on the benchmark sample, and the calendar-month prior has a lower ECE than every model run.} Calibration of the stated probabilities on fire danger forecasting, 386 items, sample positive rate 0.339. \emph{Condition} is bare or grounded; \emph{Mean $p$} is the mean stated probability; \emph{ECE} is the expected calibration error over ten equal-width bins. \emph{Reliability} and \emph{Resolution} are the two Murphy terms computed after each probability is replaced by its within-bin mean, decomposing the binned Brier score, which differs from the raw \emph{Brier} column by up to 0.005; lower reliability and higher resolution are better. \emph{Departures} counts items whose yes or no answer differs from the stated probability thresholded at one half; \emph{Omitted} counts items with a probability and no answer. The calendar-month prior is fitted on the 2006 to 2019 training years.}\label{tab:calibration}
\begin{tabular}{@{}llrrrrrrr@{}}
\toprule
Model & Condition & Mean $p$ & ECE & Brier & Reliability & Resolution & Departures & Omitted \\
\midrule
\grouprow{9}{Reference points}
\comprow Calendar-month prior & -- & 0.333 & 0.037 & 0.182 & 0.003 & 0.046 & -- & -- \\ %
\midrule
\grouprow{9}{Proprietary models}
claude-opus-4.8 & bare & 0.187 & 0.154 & 0.206 & 0.034 & 0.051 & 50 & 0 \\ %
claude-opus-4.8 & grounded & 0.220 & 0.119 & 0.200 & 0.016 & 0.040 & 50 & 2 \\ %
claude-opus-5 & bare & 0.117 & 0.222 & 0.230 & 0.069 & 0.061 & 11 & 46 \\ %
claude-opus-5 & grounded & 0.116 & 0.224 & 0.232 & 0.068 & 0.058 & 32 & 90 \\ %
gemini-3.1-pro & bare & 0.195 & 0.169 & 0.193 & 0.038 & 0.068 & 0 & 0 \\ %
gemini-3.1-pro & grounded & 0.192 & 0.185 & 0.203 & 0.036 & 0.054 & 0 & 0 \\ %
gpt-6-astra & bare & 0.060 & 0.279 & 0.281 & 0.087 & 0.025 & 0 & 0 \\ %
gpt-6-astra & grounded & 0.106 & 0.233 & 0.253 & 0.060 & 0.028 & 0 & 0 \\ %
\midrule
\grouprow{9}{Open-weight models}
Qwen3-VL & bare & 0.598 & 0.326 & 0.320 & 0.128 & 0.032 & 0 & 0 \\ %
Qwen3-VL & grounded & 0.609 & 0.326 & 0.322 & 0.136 & 0.036 & 0 & 0 \\ %
Llama 4 Maverick & bare & 0.200 & 0.218 & 0.248 & 0.050 & 0.026 & 0 & 0 \\ %
Llama 4 Maverick & grounded & 0.236 & 0.226 & 0.252 & 0.053 & 0.025 & 0 & 0 \\ %
\bottomrule
\end{tabular}
\end{table*}

On fire danger, claude-opus-5 omitted yes/no decisions on 46 bare and 90 grounded items (12 and 23 percent); other models did so twice across 3,860 items. %
Our scorer imputed missing answers by thresholding probability at 0.5. %
Grounded omissions fall on 29 percent of non-fire against 12 percent of fire cells ($p = 0.0002$, two-sided Fisher exact test). %
Bare omission rates are 13 and 11 percent, with no resolved association between omission and label. %
The grounded decision channel rests on threshold-0.5 imputation in a way the bare channel does not. %
On 267 items answered in full by every run, claude-opus-5 bare has AUPRC 0.759 and F1 0.295, while gemini-3.1-pro bare has AUPRC 0.749 and F1 0.603. %
That subset reverses their AUPRC order while preserving the ranking-versus-decision gap, and it is selected by which items the models formatted correctly. Future benchmark evaluations should re-prompt models that omit required fields rather than impute answers; calibration is measured in Appendix~\ref{app:calibration}. %
\subsection{Calibration of the Stated Probabilities on Fire Danger}\label{app:calibration}\label{subsec:sensitivity-to-evaluation-design}
Proprietary models underpredict fire frequency: gpt-6-astra places 355 items at or below 0.10 (where 29 percent burn); Qwen3-VL averages 0.60. %
The calendar-month prior is better calibrated than every model (ECE 0.037 vs 0.119--0.326; Brier 0.182 vs 0.193--0.322). %
Four models threshold stated probabilities at 0.5 on every item. claude-opus-4.8 departs on 50 items below 0.5 in both arms; claude-opus-5 departs on 11 bare and 32 grounded items and omits the answer on 46 and 90. %

\begin{table*}[tp]
\centering
\footnotesize
\setlength{\tabcolsep}{3pt}
\renewcommand{\arraystretch}{1.05}
\caption{\textbf{After a prior-shift correction, no proprietary run has a resolvably worse Brier score than the calendar-month prior, and both open-weight models stay worse.} The fire-danger probabilities of the twelve core-model runs on the 386 items of Table~\ref{tab:calibration} (rate 0.339, 352 blocks), as stated and rescored two ways, with the calendar-month prior treated alike. \emph{As stated} reproduces Table~\ref{tab:calibration}; \emph{Effective prior} is the prior that the fitted odds constant implies if a run's probabilities were calibrated under a pure label shift; it need not describe what the model believes. \emph{Prior-shifted} multiplies each run's odds by the one constant that makes its probabilities average 0.339 \citep{saerens2002adjusting}. \emph{Recalibrated} applies isotonic regression fit on four folds to the fifth, with blocks kept whole (five folds, seed 7). ECE uses ten equal-width bins; \emph{Rel.}\ and \emph{Res.}\ are the binned reliability and resolution terms of Table~\ref{tab:calibration}, and the uncertainty term is 0.224 in every row. \emph{Brier minus prior} compares each run with the prior under the same treatment by a paired 95 percent block bootstrap (20,000 resamples, seed 20260915). Recalibrated rows refit the recalibration in every resample with folds fixed at seed 7, and only gemini-3.1-pro bare resolves in all 20 fold draws; bold intervals exclude zero. The shift preserves order, so AUPRC is unchanged; isotonic maps tie neighboring values and differ by fold, which lowers AUPRC.}\label{tab:recalibration}
\begin{tabular}{@{}llrrrrrrl@{}}
\toprule
Model & Condition & Effective & ECE & Brier & Rel. & Res. & AUPRC & Brier minus prior \\
& & prior & & & & & & and 95\% CI \\
\midrule
\grouprow{9}{As stated}
\comprow Calendar-month prior & -- & 0.331 & 0.037 & 0.182 & 0.003 & 0.046 & 0.555 & -- \\ %
claude-opus-4.8 & bare & 0.149 & 0.154 & 0.206 & 0.034 & 0.051 & 0.613 & 0.024 [$-0.001$, 0.049] \\ %
claude-opus-4.8 & grounded & 0.184 & 0.119 & 0.200 & 0.016 & 0.040 & 0.581 & 0.018 [$-0.008$, 0.043] \\ %
claude-opus-5 & bare & 0.083 & 0.222 & 0.230 & 0.069 & 0.061 & 0.688 & {\bfseries\boldmath 0.048 [0.016, 0.079]} \\ %
claude-opus-5 & grounded & 0.084 & 0.224 & 0.232 & 0.068 & 0.058 & 0.682 & {\bfseries\boldmath 0.050 [0.019, 0.081]} \\ %
gemini-3.1-pro & bare & 0.036 & 0.169 & 0.193 & 0.038 & 0.068 & 0.697 & 0.010 [$-0.019$, 0.040] \\ %
gemini-3.1-pro & grounded & 0.036 & 0.185 & 0.203 & 0.036 & 0.054 & 0.682 & 0.021 [$-0.013$, 0.056] \\ %
gpt-6-astra & bare & 0.025 & 0.279 & 0.281 & 0.087 & 0.025 & 0.620 & {\bfseries\boldmath 0.099 [0.059, 0.139]} \\ %
gpt-6-astra & grounded & 0.052 & 0.233 & 0.253 & 0.060 & 0.028 & 0.590 & {\bfseries\boldmath 0.070 [0.034, 0.108]} \\ %
\addlinespace[2pt]
Qwen3-VL & bare & 0.734 & 0.326 & 0.320 & 0.128 & 0.032 & 0.485 & {\bfseries\boldmath 0.137 [0.103, 0.172]} \\ %
Qwen3-VL & grounded & 0.737 & 0.326 & 0.322 & 0.136 & 0.036 & 0.536 & {\bfseries\boldmath 0.139 [0.105, 0.173]} \\ %
Llama 4 Maverick & bare & 0.030 & 0.218 & 0.248 & 0.050 & 0.026 & 0.528 & {\bfseries\boldmath 0.065 [0.033, 0.098]} \\ %
Llama 4 Maverick & grounded & 0.056 & 0.226 & 0.252 & 0.053 & 0.025 & 0.515 & {\bfseries\boldmath 0.069 [0.037, 0.103]} \\ %
\midrule
\grouprow{9}{Prior-shifted to the 0.339 sample rate}
\comprow Calendar-month prior & -- & -- & 0.041 & 0.183 & 0.003 & 0.046 & 0.555 & -- \\ %
claude-opus-4.8 & bare & -- & 0.055 & 0.177 & 0.005 & 0.051 & 0.613 & $-0.005$ [$-0.023$, 0.012] \\ %
claude-opus-4.8 & grounded & -- & 0.064 & 0.187 & 0.006 & 0.043 & 0.581 & 0.005 [$-0.016$, 0.025] \\ %
claude-opus-5 & bare & -- & 0.067 & 0.161 & 0.009 & 0.070 & 0.688 & {\bfseries\boldmath $-0.021$ [$-0.039$, $-0.004$]} \\ %
claude-opus-5 & grounded & -- & 0.050 & 0.162 & 0.003 & 0.064 & 0.682 & {\bfseries\boldmath $-0.020$ [$-0.037$, $-0.004$]} \\ %
gemini-3.1-pro & bare & -- & 0.097 & 0.163 & 0.014 & 0.075 & 0.697 & $-0.019$ [$-0.043$, 0.005] \\ %
gemini-3.1-pro & grounded & -- & 0.116 & 0.178 & 0.019 & 0.064 & 0.682 & $-0.005$ [$-0.031$, 0.022] \\ %
gpt-6-astra & bare & -- & 0.054 & 0.176 & 0.004 & 0.052 & 0.620 & $-0.006$ [$-0.029$, 0.015] \\ %
gpt-6-astra & grounded & -- & 0.079 & 0.190 & 0.010 & 0.043 & 0.590 & 0.008 [$-0.015$, 0.031] \\ %
\addlinespace[2pt]
Qwen3-VL & bare & -- & 0.087 & 0.208 & 0.012 & 0.029 & 0.485 & {\bfseries\boldmath 0.025 [0.012, 0.038]} \\ %
Qwen3-VL & grounded & -- & 0.103 & 0.202 & 0.016 & 0.038 & 0.536 & {\bfseries\boldmath 0.019 [0.006, 0.033]} \\ %
Llama 4 Maverick & bare & -- & 0.197 & 0.249 & 0.053 & 0.028 & 0.528 & {\bfseries\boldmath 0.067 [0.036, 0.097]} \\ %
Llama 4 Maverick & grounded & -- & 0.253 & 0.274 & 0.079 & 0.028 & 0.515 & {\bfseries\boldmath 0.092 [0.057, 0.126]} \\ %
\midrule
\grouprow{9}{Recalibrated by cross-fitted isotonic regression}
\comprow Calendar-month prior & -- & -- & 0.041 & 0.184 & 0.005 & 0.046 & 0.515 & -- \\ %
claude-opus-4.8 & bare & -- & 0.071 & 0.183 & 0.009 & 0.051 & 0.566 & $-0.001$ [$-0.023$, 0.019] \\ %
claude-opus-4.8 & grounded & -- & 0.070 & 0.194 & 0.010 & 0.043 & 0.545 & 0.010 [$-0.014$, 0.031] \\ %
claude-opus-5 & bare & -- & 0.055 & 0.166 & 0.007 & 0.066 & 0.668 & $-0.018$ [$-0.040$, 0.001] \\ %
claude-opus-5 & grounded & -- & 0.040 & 0.169 & 0.005 & 0.061 & 0.665 & $-0.014$ [$-0.036$, 0.004] \\ %
gemini-3.1-pro & bare & -- & 0.032 & 0.156 & 0.005 & 0.075 & 0.685 & {\bfseries\boldmath $-0.028$ [$-0.051$, $-0.006$]} \\ %
gemini-3.1-pro & grounded & -- & 0.061 & 0.166 & 0.007 & 0.065 & 0.677 & $-0.017$ [$-0.041$, 0.004] \\ %
gpt-6-astra & bare & -- & 0.048 & 0.180 & 0.006 & 0.050 & 0.592 & $-0.004$ [$-0.028$, 0.018] \\ %
gpt-6-astra & grounded & -- & 0.068 & 0.187 & 0.007 & 0.046 & 0.555 & 0.003 [$-0.019$, 0.026] \\ %
\addlinespace[2pt]
Qwen3-VL & bare & -- & 0.032 & 0.201 & 0.009 & 0.032 & 0.454 & {\bfseries\boldmath 0.018 [0.001, 0.032]} \\ %
Qwen3-VL & grounded & -- & 0.055 & 0.187 & 0.008 & 0.046 & 0.507 & 0.003 [$-0.015$, 0.021] \\ %
Llama 4 Maverick & bare & -- & 0.087 & 0.206 & 0.015 & 0.035 & 0.499 & {\bfseries\boldmath 0.022 [0.001, 0.041]} \\ %
Llama 4 Maverick & grounded & -- & 0.049 & 0.201 & 0.005 & 0.028 & 0.500 & 0.017 [$-0.003$, 0.038] \\ %
\bottomrule
\end{tabular}
\end{table*}

\subsection{Calibration After a Prior-Shift Correction and Recalibration}\label{app:recalibration}
Prompts omit the 0.339 sample positive rate and Mesogeos's 2:1 negative subsampling ($>62$\,km from ignitions) \citep{kondylatos2023mesogeos}; models reasoning from natural fire prevalence may therefore state low probabilities.
We rescore probabilities two ways (Table~\ref{tab:recalibration}): prior-shift odds multiplication matching 0.339 \citep{saerens2002adjusting}, and 5-fold block isotonic regression (352 blocks, fold seed 7). The shift uses the sample's own positive rate, and Mesogeos violates the label-shift condition in part, because it selects negatives on their inputs. %
Stated zeros stay at zero under this shift; gemini-3.1-pro states 0 on 44 bare and 48 grounded items, of which 6 and 3 burned. %
After shifting, proprietary ECEs drop from 0.119--0.279 to 0.050--0.116 and Briers from 0.193--0.281 to 0.161--0.190 (shifted prior: 0.041 and 0.183). %
Three proprietary runs fall below the 95th percentile chance ECE \citep{brocker2007increasing}: claude-opus-4.8 bare, claude-opus-5 grounded, and gpt-6-astra bare. %
On shifted Brier, claude-opus-5 falls below the prior in both arms, no other proprietary interval excludes zero, and all open-weight runs stay above it. %
Brier decomposition \citep{dimitriadis2021stable} shows odds correction removes at least 73 percent of the gap for proprietary and Qwen3-VL runs. This share measures what one constant can repair; the cause of the gap stays unidentified. For Llama 4 Maverick that part is nil or negative, and its gap comes from residual miscalibration (0.052 and 0.077) and lower discrimination (0.015, intervals including zero). %
Recalibrated ECE depends on the fold draw: across 200 fold draws, the prior's spans 0.008 to 0.098 and the runs' medians 0.044 to 0.062. The seed-7 values of Table~\ref{tab:recalibration} (0.032--0.087, prior 0.041) are one draw. %
Over 20 fold draws, each with a refit block bootstrap of 10,000 resamples, only gemini-3.1-pro bare has a recalibrated Brier score resolvably below the prior's in all 20. claude-opus-5 bare resolves below in 13 draws, Qwen3-VL bare and Llama 4 Maverick bare resolve above in 11 and 6, and no other run resolves in more than 2. %
Shifting to the 0.331 training-year positive share moves no ECE by more than 0.017 and no Brier score by more than 0.001. %
Taking each run's mean stated probability as its source prior corrects less, because a run calibrated to a low base rate averages above that rate on this enriched sample. Under that variant no proprietary Brier score differs resolvably from the prior's, and all four open-weight runs stay above it. %
Thresholding the shifted probabilities at one half calls fire on 17 percent of items for both gpt-6-astra bare and claude-opus-5 bare, against their scored 4 and 6 percent. Their fire-class F1 would rise from 0.189 to 0.467 and from 0.269 to 0.487, a counterfactual on stored outputs with no decision loss attached. %
All 4,632 core responses carry parseable probabilities; 138 omitted answers (136 from claude-opus-5) affect only call-based columns. %

\paragraph{Rerun With the Rate Stated.}\label{app:baserate-rerun}
The rerun adds one line before the question in both arms and changes nothing else (Table~\ref{tab:baserate-rerun}). The line reads ``Base rate: in this evaluation sample, 33.9 percent of items had a wildfire of at least 30 hectares start on the target date.'' %
The proprietary four raise their mean stated probability from 0.060--0.220 to 0.267--0.355, and their ECEs fall from 0.119--0.279 to 0.025--0.148, near the prior-shifted 0.050--0.116. %
Their Brier scores improve resolvably in five of eight runs, and none is resolvably above the prior shifted to the same rate. gpt-6-astra bare falls just below it ($-0.018$, upper bound $-0.0001$). %
Their AUPRC moves by $-0.019$ to +0.063, resolvably only for gpt-6-astra grounded (0.060 [0.005, 0.113]). %
Their scored calls rise toward the sample rate, gpt-6-astra bare from 4 to 25 percent and claude-opus-5 bare from 6 to 17 percent. claude-opus-5 omits the yes-or-no field on 453 of its 772 rerun answers (136 in the first run), so its calls mostly threshold its probability at one half. %
Qwen3-VL keeps its mean stated probability near 0.61 and its ECE above 0.30. Llama 4 Maverick raises its mean to 0.37, yet calls fire on 68 and 67 percent of items. Both keep Brier scores resolvably above the shifted prior in both arms. %
With the rate stated, climatology lowers AUPRC in all six models' point estimates. The decline is resolved for claude-opus-5 ($-0.038$ [$-0.062$, $-0.015$]) and gpt-6-astra ($-0.033$, upper bound $-0.0004$). %

\begin{table*}[tp]
\centering
\scriptsize
\setlength{\tabcolsep}{1.5pt}
\renewcommand{\arraystretch}{1.08}
\caption{\textbf{Stated in the prompt, the 0.339 rate calibrates the proprietary four about as well as the prior shift, and leaves both open-weight models worse than the calendar-month prior.} Fire danger on the 386 items of Table~\ref{tab:recalibration}. Each core run is shown as first run to rerun with one added line stating the sample rate. \emph{Brier minus prior}: rerun Brier score minus that of the calendar-month prior shifted to 0.339 (0.183), with a 95 percent block-bootstrap interval (352 blocks, 20,000 resamples); * marks an interval excluding zero. \emph{Calls}: share of items called fire as scored, where an omitted call counts as $p \geq 0.5$. One run per model and arm.}\label{tab:baserate-rerun}
\begin{tabular}{@{}ll l rr l l l l@{}}
\toprule
 & & & \multicolumn{2}{c}{ECE} & & & & \\
\cmidrule(lr){4-5}
Model & Arm & Mean $p$ & First & Rerun & Brier & Brier minus prior & Calls & AUPRC \\
\midrule
\comprow Calendar prior & -- & 0.333 & 0.037 & -- & 0.182 & -- & -- & 0.555 \\ %
claude-opus-4.8 & bare & 0.187 $\to$ 0.305 & 0.154 & 0.068 & 0.206 $\to$ 0.179 & $-0.004$ [$-0.021$, 0.014] & 0.215 $\to$ 0.311 & 0.613 $\to$ 0.607 \\ %
claude-opus-4.8 & grounded & 0.220 $\to$ 0.355 & 0.119 & 0.076 & 0.200 $\to$ 0.189 & +0.006 [$-0.011$, 0.024] & 0.249 $\to$ 0.399 & 0.581 $\to$ 0.581 \\ %
claude-opus-5 & bare & 0.117 $\to$ 0.267 & 0.222 & 0.087 & 0.230 $\to$ 0.168 & $-0.015$ [$-0.035$, 0.006] & 0.065 $\to$ 0.174 & 0.688 $\to$ 0.701 \\ %
claude-opus-5 & grounded & 0.116 $\to$ 0.289 & 0.224 & 0.066 & 0.232 $\to$ 0.173 & $-0.010$ [$-0.030$, 0.010] & 0.109 $\to$ 0.241 & 0.682 $\to$ 0.663 \\ %
gemini-3.1-pro & bare & 0.195 $\to$ 0.270 & 0.169 & 0.148 & 0.193 $\to$ 0.183 & +0.001 [$-0.025$, 0.027] & 0.207 $\to$ 0.277 & 0.697 $\to$ 0.694 \\ %
gemini-3.1-pro & grounded & 0.192 $\to$ 0.284 & 0.185 & 0.148 & 0.203 $\to$ 0.188 & +0.005 [$-0.023$, 0.034] & 0.199 $\to$ 0.301 & 0.682 $\to$ 0.689 \\ %
gpt-6-astra & bare & 0.060 $\to$ 0.316 & 0.279 & 0.025 & 0.281 $\to$ 0.164 & $-0.018$ [$-0.037$, $-0.0001$] * & 0.044 $\to$ 0.251 & 0.620 $\to$ 0.683 \\ %
gpt-6-astra & grounded & 0.106 $\to$ 0.318 & 0.233 & 0.056 & 0.253 $\to$ 0.176 & $-0.007$ [$-0.028$, 0.015] & 0.078 $\to$ 0.236 & 0.590 $\to$ 0.650 \\ %
Qwen3-VL & bare & 0.598 $\to$ 0.610 & 0.326 & 0.305 & 0.320 $\to$ 0.309 & +0.127 [0.095, 0.158] * & 0.684 $\to$ 0.723 & 0.485 $\to$ 0.533 \\ %
Qwen3-VL & grounded & 0.609 $\to$ 0.600 & 0.326 & 0.313 & 0.322 $\to$ 0.306 & +0.123 [0.096, 0.151] * & 0.710 $\to$ 0.751 & 0.536 $\to$ 0.503 \\ %
Llama 4 Maverick & bare & 0.200 $\to$ 0.367 & 0.218 & 0.131 & 0.248 $\to$ 0.211 & +0.029 [0.011, 0.045] * & 0.256 $\to$ 0.681 & 0.528 $\to$ 0.529 \\ %
Llama 4 Maverick & grounded & 0.236 $\to$ 0.365 & 0.226 & 0.189 & 0.252 $\to$ 0.244 & +0.061 [0.040, 0.083] * & 0.303 $\to$ 0.668 & 0.515 $\to$ 0.503 \\ %
\bottomrule
\end{tabular}
\end{table*}

\subsection{Trained Non-LLM Baselines on the Same Items}\label{app:trained-baselines}
Both non-LLM fits read only bare-prompt features on disjoint training data and use benchmark scorers. %

\paragraph{Fire Danger.}
Trained on 19,504 samples from 2006--2019, a histogram gradient-boosted classifier on the prompt's 121 numbers scores 0.851 [0.780, 0.907] AUPRC on the 386 items. %
That exceeds gemini-3.1-pro bare by 0.154 [0.084, 0.227] on the paired block bootstrap, and reaches 0.874 on the full holdout. %
The 2,292 samples of 2020 tune the learning rate and iteration count. Logistic regression on the same 121 numbers scores 0.768. These fits achieve Brier scores of 0.108 to 0.147 against 0.193 to 0.322 for the LLMs. %
The 121 features are the numbers the bare prompt prints, at the four significant figures it prints them. The coordinates and dates the prompt also states enter the fit only through the calendar month. %

\paragraph{Allocation.}
Trained on 30,869 fire-days from 1,499 disjoint incidents, a gradient-boosted regressor on the 23 bare-prompt features reaches 0.144 normalized MAE on the 300 items (Table~\ref{tab:trained-allocation}). %
Training used pre-2015 or California incidents; cross-validation grouped by incident within training chose hyperparameters. %
It matches persistence at 0.146 normalized error and beats every prompted run. %
Adding the analogue median leaves test error unchanged (0.1441 to 0.1447). %

\begin{table*}[t]
\centering
\footnotesize
\setlength{\tabcolsep}{3.5pt}
\renewcommand{\arraystretch}{1.10}
\caption{\textbf{A boosted regressor on the bare prompt numbers reaches 0.144 normalized error on allocation, matching persistence without gaining from the displayed analogues.} Trained regressors on daily personnel allocation, in the columns of Table~\ref{tab:results-allocation}, on the same 300 items. \emph{Fit} specifies the model and target; \emph{MAE} is mean absolute error; \emph{Norm.} is MAE normalized by incident mean staffing; \emph{Beats base} is the share of items beating persistence; \emph{Stable days} and \emph{Moving days} give normalized error by true movement; \emph{False move} and \emph{Missed move} are directional error shares; \emph{Dir.\ right} is the share moving the correct direction. Each fit uses 30,869 fire-days from 1,499 incidents disjoint from the evaluation incidents and reads the 23 bare-prompt fields; the second row also reads the displayed analogue median.}\label{tab:trained-allocation}
\begin{tabular}{@{}>{\raggedright\arraybackslash}p{2.3in}rrrrrrrr@{}}
\toprule
Fit & MAE & Norm. & Beats & Stable & Moving & False & Missed & Dir. \\
& & & base & days & days & move & move & right \\
\midrule
\grouprow{9}{Reference points}
\comprow Boosted regressor, next-day ratio & 12.97 & 0.144 & 0.48 & 0.035 & 0.317 & 0.03 & 0.93 & 0.72 \\ %
\comprow Same, plus the displayed analogue median & 13.03 & 0.145 & 0.48 & 0.035 & 0.319 & 0.03 & 0.94 & 0.71 \\ %
\comprow Boosted regressor, next-day count & 13.68 & 0.201 & 0.42 & 0.126 & 0.319 & 0.14 & 0.81 & 0.66 \\ %
\comprow Ridge regression, next-day count & 19.05 & 0.416 & 0.28 & 0.412 & 0.423 & 0.63 & 0.47 & 0.57 \\ %
\bottomrule
\end{tabular}
\end{table*}

\subsection{Prompt Sensitivity on Fire Danger}\label{app:prompt-sensitivity}
Two bare prompt paraphrases (p1, p2) reorder model standings relative to original prompt p0 (Table~\ref{tab:prompt-sensitivity}): gemini-3.1-pro leads under p0 (0.697), claude-opus-5 under p1 (0.731), and gpt-6-astra under p2 (0.711). %
Llama 4 Maverick falls to 0.393 (p1) and 0.400 (p2), behind Qwen3-VL. %
Under p2, call rates for gpt-6-astra and claude-opus-4.8 drop to 0.003 and 0.010. %
Byte-identical repeats of p0 moved claude-opus-5 by 0.019 [$-0.001$, 0.039] and gemini-3.1-pro by 0.009 [$-0.021$, 0.039], comparable to the 0.01 lead gap. %

\begin{table*}[t]
\centering
\footnotesize
\setlength{\tabcolsep}{3pt}
\renewcommand{\arraystretch}{1.10}
\caption{\textbf{Single runs of two paraphrases reorder the six models: gpt-6-astra rises from third to first, the two leaders swap, and Llama 4 Maverick loses 0.13. Two models nearly stop calling fire.} Prompt sensitivity on fire danger forecasting, bare condition, 386 items, one run per paraphrase. \emph{Model} is the evaluated system. \emph{Prompt} specifies the paraphrase: p0 is the paper prompt, and p1 and p2 differ in framing, field order, and layout. \emph{Difference in AUPRC} is AUPRC under the paraphrase minus p0, with a paired block-by-month bootstrap interval. \emph{Spearman} is the rank correlation of stated probabilities across the two prompts.}\label{tab:prompt-sensitivity}
\begin{tabular}{@{}llrrrrlr@{}}
\toprule
Model & Prompt & AUPRC & F1 on fire & Call rate & Omitted & Difference in AUPRC & Spearman \\
\midrule
\grouprow{8}{Proprietary models}
claude-opus-4.8 & p0 & 0.613 & 0.561 & 0.215 & 0 &  &  \\ %
claude-opus-4.8 & p1 & 0.605 & 0.234 & 0.060 & 0 & $-0.008$ [$-0.053$, 0.035] & 0.93 \\ %
claude-opus-4.8 & p2 & 0.631 & 0.059 & 0.010 & 0 & 0.018 [$-0.025$, 0.057] & 0.91 \\ %
\midrule
claude-opus-5 & p0 & 0.688 & 0.269 & 0.065 & 46 &  &  \\ %
claude-opus-5 & p1 & 0.731 & 0.308 & 0.065 & 0 & 0.043 [0.009, 0.080] & 0.92 \\ %
claude-opus-5 & p2 & 0.703 & 0.240 & 0.049 & 4 & 0.015 [$-0.018$, 0.049] & 0.93 \\ %
\midrule
gemini-3.1-pro & p0 & 0.697 & 0.607 & 0.207 & 0 &  &  \\ %
gemini-3.1-pro & p1 & 0.648 & 0.549 & 0.161 & 0 & $-0.049$ [$-0.093$, $-0.009$] & 0.77 \\ %
gemini-3.1-pro & p2 & 0.672 & 0.452 & 0.119 & 0 & $-0.026$ [$-0.075$, 0.020] & 0.70 \\ %
\midrule
gpt-6-astra & p0 & 0.620 & 0.189 & 0.044 & 0 &  &  \\ %
gpt-6-astra & p1 & 0.649 & 0.074 & 0.013 & 0 & 0.028 [$-0.033$, 0.093] & 0.78 \\ %
gpt-6-astra & p2 & 0.711 & 0.015 & 0.003 & 0 & 0.091 [0.033, 0.147] & 0.77 \\ %
\midrule
\grouprow{8}{Open-weight models}
Qwen3-VL & p0 & 0.485 & 0.567 & 0.684 & 0 &  &  \\ %
Qwen3-VL & p1 & 0.496 & 0.521 & 0.396 & 0 & 0.010 [$-0.046$, 0.067] & 0.65 \\ %
Qwen3-VL & p2 & 0.472 & 0.571 & 0.541 & 0 & $-0.013$ [$-0.063$, 0.035] & 0.74 \\ %
\midrule
Llama 4 Maverick & p0 & 0.528 & 0.513 & 0.256 & 0 &  &  \\ %
Llama 4 Maverick & p1 & 0.393 & 0.359 & 0.238 & 0 & $-0.136$ [$-0.200$, $-0.069$] & 0.33 \\ %
Llama 4 Maverick & p2 & 0.400 & 0.208 & 0.060 & 0 & $-0.128$ [$-0.199$, $-0.059$] & 0.31 \\ %
\bottomrule
\end{tabular}
\end{table*}
\subsection{A Second Retrieval Rule for the Analogues}\label{app:retrieval-v2}
A nearest-neighbour rule (v2) tests whether analogue error increases stemmed from movement dynamics mismatches. %
It was selected on a development set of 599 items from 520 incidents that share no incident with the evaluation set. Three rule families were searched, with the selection rule fixed before any design ran. It picked the lowest development error among rules whose movement AUC exceeded the paper's rule by 0.05 and whose false-move rate fell below it. %
The winner's development error and interval are therefore conditional on that selection. Its gain is smaller on the 73 development items from 2015 or later than on the earlier 526. %
The rule selects six pool rows from other incidents within the target staffing band preceding the report day. %
Candidates are ranked by weighted squared distance on seven standardized features fitted on pre-2015 rows, and ties use the analogue identifier with no random draw. %
These features are log personnel, \texttt{log1p} acres, containment clipped to 0 to 100, log run day, the one-day log personnel change, \texttt{log1p} new acres, and \texttt{log1p} aerial resources, weighted $(1,1,1,1,4,3,1)$. %
On the 300 evaluation items, the analogue-only baseline improved from 0.249 (v1) to 0.190 (v2), and false-move rate dropped from 0.40 to 0.20. %
Under v2, grounded error fell across all six models (Table~\ref{tab:retrieval-v2}), and v2 $-$ v1 intervals excluded zero for gpt-6-astra and Qwen3-VL. False movement decreased to 0.20--0.22 on all six (down from 0.27--0.39 under v1). %
The rule changes eligibility, distance, and the random step together and isolates none of them. %

\begin{table*}[t]
\centering
\footnotesize
\setlength{\tabcolsep}{3pt}
\renewcommand{\arraystretch}{1.10}
\caption{\textbf{Under the nearest-neighbour retrieval rule (v2), grounded allocation error falls on every model and false movement decreases across all six.} Daily personnel allocation, 300 items, under the paper's analogue rule (v1) and the frozen nearest-neighbour rule (v2). \emph{Bare}, \emph{v1}, and \emph{v2} report normalized error as in Table~\ref{tab:results-allocation}; the two difference columns carry incident-clustered 95 percent intervals; \emph{False move} is the share of stable days moved by more than a tenth, bare / v1 / v2; \emph{Copies} counts items whose prediction equals persistence times the displayed median ratio, v1 / v2. The analogue-only rules themselves score 0.249 (v1) and 0.190 (v2) against persistence at 0.146.}\label{tab:retrieval-v2}
\begin{tabular}{@{}lrrrrrll@{}}
\toprule
Model & Bare & v1 & v2 & v2 $-$ bare & v2 $-$ v1 & False move & Copies \\
\midrule
\grouprow{8}{Proprietary models}
claude-opus-4.8 & 0.157 & 0.170 & 0.166 & 0.010 & $-0.004$ & 0.14 / 0.29 / 0.22 & 123 / 145 \\ %
 & & & & {\scriptsize [$-0.002$, 0.023]} & {\scriptsize [$-0.020$, 0.014]} & & \\ %
claude-opus-5 & 0.165 & 0.173 & 0.166 & 0.000 & $-0.007$ & 0.25 / 0.27 / 0.20 & 82 / 93 \\ %
 & & & & {\scriptsize [$-0.011$, 0.011]} & {\scriptsize [$-0.022$, 0.006]} & & \\ %
gemini-3.1-pro & 0.179 & 0.177 & 0.166 & $-0.012$ & $-0.011$ & 0.23 / 0.29 / 0.20 & 154 / 152 \\ %
 & & & & {\scriptsize [$-0.026$, 0.001]} & {\scriptsize [$-0.027$, 0.005]} & & \\ %
gpt-6-astra & 0.161 & 0.220 & 0.176 & 0.014 & $-0.044$ & 0.18 / 0.30 / 0.20 & 199 / 227 \\ %
 & & & & {\scriptsize [$-0.000$, 0.032]} & {\scriptsize [$-0.089$, $-0.007$]} & & \\ %
\midrule
\grouprow{8}{Open-weight models}
Qwen3-VL & 0.161 & 0.248 & 0.187 & 0.026 & $-0.062$ & 0.08 / 0.39 / 0.20 & 284 / 283 \\ %
 & & & & {\scriptsize [0.003, 0.052]} & {\scriptsize [$-0.114$, $-0.016$]} & & \\ %
Llama 4 Maverick & 0.193 & 0.246 & 0.237 & 0.045 & $-0.008$ & 0.26 / 0.37 / 0.21 & 259 / 258 \\ %
 & & & & {\scriptsize [$-0.021$, 0.150]} & {\scriptsize [$-0.094$, 0.111]} & & \\ %
\bottomrule
\end{tabular}
\end{table*}

\subsection{A Defect in This Benchmark: The Output Cap Chose Who Got Scored}\label{app:defect-ours}
Early caps (200 tokens for smoke, 250 for fire danger, 300 for allocation) truncated reasoning models before emitting answers. %
Under the 200-token cap, gemini-3.1-pro lost 40 of 196 grounded smoke items, claude-opus-5 lost 8, and claude-opus-4.8 lost none. %
On those 40 items, gemini-3.1-pro stored only an opening code fence against 196 completion tokens. %
Because reasoning length correlates with item difficulty, this cap selected which items each model completed. %

The benchmark established a single 1,536-token cap across all tasks, models, and conditions. %
Across 36 primary files (10,752 responses), 45 gemini-3.1-pro responses used at least 1,520 tokens, but each reached its answer fields. %
Bare tool use is the exception: 61 of gemini-3.1-pro's 156 bare responses reached the cap before their ANSWER line. %
Fallback scoring credited 3 of the 61 correct against 22 of its other 95 bare items; claude-opus-5 and Qwen3-VL each lost one bare answer. %
The cap may have lowered these bare tool-use scores, gemini-3.1-pro's 0.160 most of all (Appendix~\ref{app:reasoning}). %

\paragraph{A Unit Gloss in the Aerial Prompt.}
Our grounded aerial prompt introduces the thermal block as ``degrees Celsius; pct fields are shares of pixels.'' %
Both fields actually hold percentages, computed as 100 times the pixel count over 327,680 pixels across all 12,194 release records. %
On 53 of our 408 items a value exceeds 1. %
On the 48 closed-form items, Qwen3-VL and Llama 4 Maverick gave answers matching a fraction reading on 4 and 3 items answered correctly bare; no proprietary model did. %
Writing ``percent of pixels'' would remove ambiguity; these seven misses remain in reported scores. %

\subsection{Sensitivity to Image Width on FIgLib}\label{app:figlib-resolution}
On 43 frames captured at least 25 minutes post-plume, smoke recall was 0.535 at 1024 pixels and 0.628 at 1568 pixels, gaining 4 frames with none lost. %
This pilot does not separate width from run variation; reported FIgLib evaluations use 1568 pixels. %

\subsection{FIgLib Scored Without Pairing}\label{app:figlib-unpaired}
Scoring each condition on all the items it completed (224 bare versus 196 grounded) gives the bare arm 28 pre-plume clear frames that the grounded arm lacks. These frames inflate bare accuracy to 0.737--0.808 and leave recall on smoke unchanged; the paired scoring of Table~\ref{tab:results-figlib} supersedes this one. %
On 196 paired items, detection accuracy is lowest within the first ten minutes (0.25--0.57) and highest after 25 minutes (0.51--0.81), while pre-plume clear frames score 0.94--1.00 (Table~\ref{tab:figlib-buckets}). %

\begin{table*}[t]
\centering
\footnotesize
\setlength{\tabcolsep}{4pt}
\renewcommand{\arraystretch}{1.10}
\caption{\textbf{Detection accuracy is lowest in the first ten minutes after the plume across every evaluated model and condition.} Accuracy on FIgLib frames after the first visible plume, by offset bucket, on 196 paired items (buckets hold 28, 47, and 37 frames; pre-plume clear frames score 0.94--1.00).}\label{tab:figlib-buckets}
\begin{tabular}{@{}llrrr@{}}
\toprule
Model & Condition & 0 to 10 min & 10 to 25 min & 25 min or more \\
\midrule
\grouprow{5}{Proprietary models}
claude-opus-4.8 & bare & 0.43 & 0.57 & 0.62 \\ %
claude-opus-4.8 & grounded & 0.46 & 0.66 & 0.68 \\ %
claude-opus-5 & bare & 0.50 & 0.66 & 0.62 \\ %
claude-opus-5 & grounded & 0.50 & 0.72 & 0.65 \\ %
gemini-3.1-pro & bare & 0.25 & 0.64 & 0.65 \\ %
gemini-3.1-pro & grounded & 0.50 & 0.81 & 0.81 \\ %
gpt-6-astra & bare & 0.46 & 0.70 & 0.70 \\
gpt-6-astra & grounded & 0.57 & 0.83 & 0.76 \\
\midrule
\grouprow{5}{Open-weight models}
Qwen3-VL & bare & 0.36 & 0.57 & 0.65 \\ %
Qwen3-VL & grounded & 0.43 & 0.64 & 0.73 \\ %
Llama 4 Maverick & bare & 0.36 & 0.60 & 0.51 \\ %
Llama 4 Maverick & grounded & 0.36 & 0.64 & 0.59 \\ %
\bottomrule
\end{tabular}
\end{table*}

\subsection{Tool Use on FPA-FOD: Harness and Failure Detail}\label{app:tooluse-detail}
The 156 tool-use items were generated deterministically across twelve SQL templates over 2,303,566 FPA-FOD records. %
Filters excluded degenerate cases and the 38 records with date-year mismatches. Family guards cut degenerate cells: a floor on the cell size, a unique argmax with a stated margin, a non-blank name, and leap-free windows. Survivors are reduced to one cell per state and sampled by an even stride. No category takes more than 4 of the 13 slots in the two categorical families. %
Each item carries its scoring rule: exact match for counts, years, and names, a relative tolerance of 0.5 percent (floor one acre) for acres, and 0.1 point for percentages. %
The grounded arm queries a frozen SQLite snapshot via read-only \texttt{query\_fpafod(sql)}, restricted to 50 rows, 4,000 characters, and 8 calls per question. %
The tool accepts one \texttt{SELECT} or \texttt{WITH} statement; both arms receive the schema and must end with \texttt{ANSWER: <value>} or \texttt{ANSWER: unknown}, and the bare arm has no tool. %
Output cap was 1,536 tokens; requests specified temperature 0.0 for five models, while gpt-6-astra's Responses API fixes 1.0. Effective gateway settings were not verified.
Tool reasoning tokens for gpt-6-astra were unrecorded; bare reasoning averaged 22.5 tokens. %
The best constant per family bounds constant guessing from above, describes no usable predictor, and scores 22 of 156 (Table~\ref{tab:results-tooluse}). %

Two rewrites of all 156 questions test whether the tool ceiling reflects templates: a phrasing rewrite changing syntax, and a de-schematized rewrite removing schema tokens (Section~\ref{subsec:results-tooluse}). %
Both keep every non-question field byte-identical to the original, the reference query and answer included, so the scorer and the tolerances apply unchanged. %

Proprietary models completed 624 items with one failure (a gpt-6-astra refusal). %
Open-weight models made 35 errors of two kinds. %
Thirty are missed calendar windows, from filtering the unpadded \texttt{m/d/yyyy} strings in \texttt{DISCOVERY\_DATE} against zero-padded or ISO dates the model guessed instead of using \texttt{DISCOVERY\_DOY}. %
Five are Python code snippets from Llama 4 Maverick on percentage queries, parsed as 2.0 from \texttt{.2f}. %
On the 26 calendar-window items, proprietary models used day of year (three on all 26, gpt-6-astra on 17) or inspected values and recovered. %
On the numeric questions the models did answer bare, the median relative error runs from 0.28 for claude-opus-4.8 (31 answers) to 1.00 for gemini-3.1-pro (64 answers). %

\subsection{Served Models and the Run Manifest}\label{app:manifest}
Every response row records the model identifier returned by serving infrastructure. %
The scored-response total counts recorded responses on the paired item inventories, unanswered responses included. It leaves out the 448 bare smoke rows with no grounded counterpart: $78{,}092 - 448 = 77{,}644$. %
Each run carries one served identifier. %
The core record holds 36 primary files (10,752 responses), 12 tool-use files (1,872 responses), and 12 aerial files (4,896 responses) with their image map. %
The sweep adds 210 files (60,572 responses). %
The gateway served \texttt{claude-opus-4.8}, \texttt{claude-opus-5}, \texttt{gemini-3.1-pro}, and \texttt{gpt-6-astra}; Bedrock served \texttt{qwen.qwen3-vl-\allowbreak 235b-a22b} and \texttt{us.meta.llama4-maverick-\allowbreak 17b-instruct-v1:0}. %
Metadata, SHA-256 hashes, and timestamps are tracked in \texttt{manifest-v1.json}, excluding a ten-item credential probe and 1,024-pixel pilot, and recording the override for five repaired allocation files. %

\subsection{How the Smoke Task Got Its Third Family, and the Repeat Runs}\label{app:figlib-gap}
The reported gemini-3.1-pro smoke run is the third dispatch of that configuration. The first lost 40 grounded items to the initial 200-token cap (Appendix~\ref{app:defect-ours}), and the second hit gateway rate limits on all 420 calls. %
Repeat runs measure run-to-run variation (Appendix~\ref{app:prompt-sensitivity}). %
Qwen3-VL changed 1 of 224 smoke verdicts and moved fire danger average precision from 0.485 to 0.509, flipping 8 of 386 binary decisions. %
Llama 4 Maverick moved allocation error from 0.186 to 0.194 on 296 common items, replacing 4 initial timeouts. %
gpt-6-astra at temperature 1.0 moved allocation error from 0.1614 to 0.1613, changing 57 predictions by 0.99 personnel on average. %
Its endpoint accepts only that default temperature, and the 1,536-token cap also counts the reasoning tokens it bills; its longest answer, reasoning included, used 318 tokens. %
claude-opus-5 and gemini-3.1-pro moved fire danger AUPRC by 0.019 and 0.009. %

\subsection{Metrics, by Task}\label{app:metrics-detail}
Reported intervals are 95 percent percentile bootstraps with 20,000 resamples over the task's clustering unit, paired across conditions at seed 20260915. %
Clustering units are 17 fires (smoke), 390 frames (aerial QA, holding 34 types fixed), 245 incidents (allocation), 352 spatial blocks by month (fire danger), and 12 question families (tool use). %

\paragraph{Small-Cluster Sensitivity Expansion.}
With few clusters a percentile interval may undercover. %
As a sensitivity analysis we expand the three small-unit intervals about their midpoints by $F_G = t_{G-1,0.975}\sqrt{G/(G-1)}/z_{0.975}$. %
Widths grow by 11.5 percent for smoke ($G=17$), 17.3 percent for tool use ($G=12$), and 19.2 percent for the 143-item subset ($G=11$). %
The expansion is a heuristic: it establishes no nominal coverage, it is not a studentized bootstrap, and it dilates symmetrically without respecting the bounds of a bounded metric. %
Intervals on the three larger units remain plain percentile intervals. %

\subsection{Aerial Question Answering: Build, Harness, and Strata}\label{app:wildfirevqa-detail}\label{app:pending}\label{subsec:results-pending}

\paragraph{Dataset Sampling.}
The source release contains 207,195 valid question records after discarding 92 records with a missing or zero applicability score and 11 records whose listed answer does not match any option. Filtering them to the Sycan Marsh burn yields 25,027 records across 738 images and 34 question types. %
Sampling 12 items per type yields 408 evaluation items over 390 distinct frames. %
Burn units and fire conditions divide into Sycan 2A with fire (201 items), Sycan 2D with fire (144 items), and Sycan 2D without fire (63 items). %
Target labels originate from radiometric formulas (108 items), multimodal verification (276), drone telemetry (12), and object detectors (12). %
Dataset applicability is 1.0 on 284 items and below 1.0 on 124 items (minimum 0.1). %
Selecting pool majority reaches 0.627 accuracy on the 408 items (0.583 outside four near-degenerate types). %
Combining the majority with closed-form formulas on types CL1, CMR4, DS7, and DS8 yields 0.664. %

\paragraph{Image Alignment and Rendering.}
All 390 frames matched FLAME~3 images unambiguously by Celsius TIFF statistics; Sycan 2D fire 00168 required a 0.0015 mean tolerance from summation order. %
Thermal inputs render from each Celsius TIFF with matplotlib's inferno colormap scaled from minimum to maximum temperature. %

\paragraph{Prompt and Harness.}
Bare prompts supply RGB and rendered inferno thermal frames. %
Grounded prompts add a seven-number thermal block: minimum, maximum, mean, standard deviation, mean temperature of the hottest 3 percent of pixels, and percentages of pixels above 200$^\circ$C and 400$^\circ$C. %
The prompt glosses those two fields as shares of pixels, though they hold percentages (Appendix~\ref{app:defect-ours}). %
The grounded prompt directs models to treat minimum and maximum temperatures as monotonic colormap endpoints. %
Models return a JSON payload with the chosen option text and an applicability score in [0.0, 1.0]. The parser resolves choices by normalized string match, then longest unique option containment, and, when the response carries no JSON answer, a free-text match that names exactly one option. %
The parser resolved all 408 items in all 12 arms, after four retried gateway errors on bare gemini-3.1-pro. %

\begin{table*}[t]
\centering
\footnotesize
\setlength{\tabcolsep}{5pt}
\renewcommand{\arraystretch}{1.10}
\caption{\textbf{The thermal block lifts Cross-Modal Reasoning and Distribution on most models while lowering Classification and Presence.} Accuracy by category on WildFireVQA across all six models, bare and grounded, beside the majority baseline. Categories: Classification (CL, 72 items), Cross-Modal Reasoning (CMR, 48), Distribution and Segmentation (DS, 96), Flight Planning (FP, 48), Localization and Direction (LD, 48), and Presence and Detection (PD, 96).}\label{tab:wildfirevqa-category}
\vspace{1pt}
\begin{tabular}{@{}ll rrrrrr@{}}
\toprule
Model & Condition & CL & CMR & DS & FP & LD & PD \\
\midrule
\comprow \multicolumn{2}{@{}l}{Per-question majority} & 0.708 & 0.438 & 0.573 & 0.604 & 0.542 & 0.771 \\ %
\midrule
\grouprow{8}{Proprietary models}
claude-opus-4.8 & bare & 0.625 & 0.562 & 0.625 & 0.625 & 0.521 & 0.781 \\ %
claude-opus-4.8 & grounded & 0.569 & 0.708 & 0.604 & 0.604 & 0.542 & 0.771 \\ %
claude-opus-5 & bare & 0.694 & 0.625 & 0.594 & 0.646 & 0.562 & 0.729 \\ %
claude-opus-5 & grounded & 0.681 & 0.750 & 0.625 & 0.646 & 0.542 & 0.688 \\ %
gemini-3.1-pro & bare & 0.514 & 0.771 & 0.510 & 0.625 & 0.625 & 0.781 \\ %
gemini-3.1-pro & grounded & 0.528 & 0.792 & 0.625 & 0.583 & 0.583 & 0.750 \\ %
gpt-6-astra & bare & 0.569 & 0.750 & 0.521 & 0.708 & 0.542 & 0.698 \\ %
gpt-6-astra & grounded & 0.625 & 0.729 & 0.562 & 0.667 & 0.521 & 0.708 \\ %
\midrule
\grouprow{8}{Open-weight models}
Qwen3-VL & bare & 0.750 & 0.688 & 0.552 & 0.479 & 0.521 & 0.667 \\ %
Qwen3-VL & grounded & 0.708 & 0.833 & 0.562 & 0.417 & 0.542 & 0.635 \\ %
Llama 4 Maverick & bare & 0.625 & 0.479 & 0.479 & 0.417 & 0.500 & 0.625 \\ %
Llama 4 Maverick & grounded & 0.597 & 0.604 & 0.490 & 0.438 & 0.458 & 0.625 \\ %
\bottomrule
\end{tabular}
\end{table*}

\paragraph{Collinear Label Sources.}
Each of the 34 question types draws its reference answers from exactly one source. %
Appendix~\ref{subsec:app-wildfirevqa-detail} reports accuracy by label source and by applicability (Table~\ref{tab:aerial-strata}) and explains why the split cannot bound label error. %

\paragraph{Clustering Units.}
Items cross 390 frames and 34 question types, with the majority comparator defined per type. %
Frame clustering treats the question set as fixed, matching reporting scope. %
A two-way pigeonhole bootstrap resamples both dimensions and weights each item by the product of its draw counts. %
Table~\ref{tab:wildfirevqa-clustering} reports paired contrasts under all three designs. %
The pooled arm contrast contains zero for all six models under every resampling. %
Llama 4 Maverick's separation below the majority comparator is the one result that depends on the choice, excluding zero by frame but reaching zero when question types are resampled. %

\begin{table*}[t]
\centering
\footnotesize
\setlength{\tabcolsep}{5pt}
\renewcommand{\arraystretch}{1.05}
\caption{\textbf{The pooled arm contrast holds under every resampling; one comparator result does not.} Paired contrasts on the 408 aerial items under three bootstrap designs: clustering on the 390 frames, clustering on the 34 question types, and a two-way pigeonhole bootstrap over both. All use 20,000 resamples at seed~20260915. \emph{Grounded $-$ bare} is the pooled arm contrast. \emph{Bare $-$ majority} is the contrast against the held-out per-question majority comparator of 0.627.}\label{tab:wildfirevqa-clustering}
\vspace{1pt}
\begin{tabular}{@{}l r ccc@{}}
\toprule
Model & Diff. & By frame & By question type & Two-way \\
\midrule
\grouprow{5}{Grounded $-$ bare}
claude-opus-4.8 & $+0.000$ & [$-0.027$, $+0.027$] & [$-0.039$, $+0.056$] & [$-0.056$, $+0.071$] \\  %
claude-opus-5 & $+0.007$ & [$-0.020$, $+0.035$] & [$-0.037$, $+0.064$] & [$-0.050$, $+0.079$] \\  %
gemini-3.1-pro & $+0.015$ & [$-0.015$, $+0.044$] & [$-0.025$, $+0.059$] & [$-0.040$, $+0.077$] \\  %
gpt-6-astra & $+0.012$ & [$-0.012$, $+0.037$] & [$-0.012$, $+0.039$] & [$-0.030$, $+0.056$] \\  %
Qwen3-VL & $+0.000$ & [$-0.030$, $+0.030$] & [$-0.039$, $+0.042$] & [$-0.057$, $+0.060$] \\  %
Llama 4 Maverick & $+0.010$ & [$-0.022$, $+0.043$] & [$-0.032$, $+0.056$] & [$-0.053$, $+0.079$] \\  %
\midrule
\grouprow{5}{Bare $-$ majority comparator}
claude-opus-4.8 & $+0.015$ & [$-0.046$, $+0.076$] & [$-0.071$, $+0.098$] & [$-0.108$, $+0.136$] \\  %
claude-opus-5 & $+0.022$ & [$-0.037$, $+0.085$] & [$-0.074$, $+0.118$] & [$-0.110$, $+0.149$] \\  %
gemini-3.1-pro & $+0.005$ & [$-0.059$, $+0.069$] & [$-0.093$, $+0.100$] & [$-0.126$, $+0.135$] \\  %
gpt-6-astra & $-0.005$ & [$-0.072$, $+0.062$] & [$-0.105$, $+0.093$] & [$-0.142$, $+0.129$] \\  %
Qwen3-VL & $-0.010$ & [$-0.072$, $+0.054$] & [$-0.113$, $+0.091$] & [$-0.144$, $+0.123$] \\  %
Llama 4 Maverick & $-0.093$ & [$-0.156$, $-0.027$] & [$-0.189$, $+0.000$] & [$-0.225$, $+0.037$] \\  %
\bottomrule
\end{tabular}
\end{table*}

\section{By-Task Detail}\label{app:bytask}
This appendix holds the per-category record behind Section~\ref{sec:crosscut}.
Table~\ref{tab:bytask-full} reports confirmed findings, verified model counts, and grounding status for each of the 25 re-checked fire records across the task categories of Table~\ref{tab:taxonomy}. %
Table~\ref{tab:bytask} in Appendix~\ref{app:moved-floats} summarizes these records by category. %

\begin{table*}[t]
\centering
\footnotesize
\setlength{\tabcolsep}{4pt}
\renewcommand{\arraystretch}{1.10}
\caption{\textbf{Confirmed findings, confirmed model counts, and grounding status across the 25 re-checked fire records.} One row per record, grouped by the task category its confirmed finding measures. \emph{Multi-hazard} marks a disaster-wide evaluation that includes a fire task; \emph{Both} marks a study that compares a bare condition against a grounded one.}\label{tab:bytask-full}
\vspace{1pt}
\begin{tabular}{@{} >{\raggedright\arraybackslash}p{1.18in} >{\raggedright\arraybackslash}p{0.90in} >{\raggedright\arraybackslash}p{0.52in} >{\raggedright\arraybackslash}p{0.88in} >{\raggedright\arraybackslash}p{0.58in} >{\raggedright\arraybackslash}p{1.64in} @{}}
\toprule
Work & Category & Domain & Models confirmed & Bare or grounded & Headline number \\
\midrule
RegCheck-Hybrid \citep{chen2025regcheck} & Knowledge QA & Structure & Rule-LLM hybrid, RAG LLM baseline & Both & 89\% F1 on building fire code compliance \\ %
Triple-Phase control \citep{sun2026triplephase} & Knowledge QA & Structure & Commercial RAG LLMs, pure LLMs & Both & Hallucination on numerical thresholds eliminated \\ %
WFDSS barrier analysis \citep{epstein2025barriers} & Documents & Wildfire & GPT-4 (1) & Bare & 13 barrier types over 24,254 entries; roads 42\% \\ %
Task-state monitoring \citep{grunert2026taskstate} & Documents & Fire service & gpt-5.2, gpt-5.4 (2) & Bare & Strong incrementally; unit-assignment timing weakest \\ %
DisasTeller \citep{chen2026disasteller} & Documents & Multi-hazard & GPT-4o, Gemma3 (2) & Grounded & Damage reports on earthquake, flood, and bushfire \\ %
Fire protection specifications \citep{kim2025firespec} & Documents & Structure & Gemini, Claude, ChatGPT (3) & Bare & Adequate structures; expert review still required \\ %
MAGR-FI \citep{tao2026magrfi} & Documents & Structure & MAGR-FI, baseline LLMs & Grounded & Causal reasoning 6.95 to 8.51 over 1,051 cases \\ %
Zero-shot satellite VLMs \citep{kohli2025zeroshot} & Detection & Wildfire & Gemini 2.5 Flash, Llama 3.2-Vision, LLaVA (3) & Bare & 0.98, 0.966, and 0.728 accuracy on 500 images \\ %
ForestFireVLM \citep{seidel2025forestfirevlm} & Detection & Wildfire & ForestFireVLM-7B, Gemini 2.0 Pro, GPT-4o & Bare & 76.6\% against 65.5\% for Gemini 2.0 Pro \\ %
SmokeBench \citep{qi2026smokebench} & Detection & Wildfire & 7 MLLMs and 2 detectors & Bare & Large plumes classified well; early localization poor \\ %
DisasterVQA \citep{almohannadi2026disastervqa} & Detection & Multi-hazard & 7 VLMs & Bare & 0.93--1.00 binary, 0.84 multiple choice \\ %
WildFireVQA \citep{habibpour2026wildfirevqa} & Detection & Wildfire & 4 open MLLMs of 8B or fewer & Both & 207,298 questions; temperature reasoning weak \\ %
WildfireVLM \citep{ayanzadeh2026wildfirevlm} & Detection & Wildfire & GPT-4o, Claude 3.5 Sonnet, 4 detectors (6) & Grounded & 7.03 out of 10 against 6.16 under a judge rubric \\ %
\bottomrule
\end{tabular}
\end{table*}

\begin{table*}[t]
\centering
\ContinuedFloat
\footnotesize
\setlength{\tabcolsep}{4pt}
\renewcommand{\arraystretch}{1.10}
\caption{(continued)}
\vspace{1pt}
\begin{tabular}{@{} >{\raggedright\arraybackslash}p{1.18in} >{\raggedright\arraybackslash}p{0.90in} >{\raggedright\arraybackslash}p{0.52in} >{\raggedright\arraybackslash}p{0.88in} >{\raggedright\arraybackslash}p{0.58in} >{\raggedright\arraybackslash}p{1.64in} @{}}
\toprule
Work & Category & Domain & Models confirmed & Bare or grounded & Headline number \\
\midrule
DisasterBench \citep{zhang2026disasterbench} & Detection & Multi-hazard & 21 baseline MLLMs plus DisasterVL & Bare & 2B DisasterVL matched GPT-4o at 72.60\% \\ %
WildfireGPT assessment \citep{ramesh2025wildfiregpt} & Forecasting & Wildfire & WildfireGPT (GPT-4), TabNet (2) & Grounded & MAE 14.849 and $R^2$ 0 against MAE 0.055 \\ %
FireScope \citep{markov2026firescope} & Forecasting & Wildfire & Qwen2.5-VL-7B, GPT-5 (3) & Grounded & Out-of-distribution transfer from US to Europe \\ %
Retrieval over Response \citep{cheng2026retrieval} & Forecasting & Wildfire & Not listed & Both & $R=0.009$ direct against $R=0.598$ guided \\ %
IC-EO \citep{lahouel2026iceo} & Geospatial & Wildfire & IC-EO agent, GPT-4o, LLaVA (3) & Both & 50\% against 0\% for GPT-4o and LLaVA \\ %
Geospatial-aware agents \citep{chen2025geospatial} & Operations & Wildfire & About 14 LLM configurations & Both & Text-only over-allocated personnel 2--3 times \\ %
OGC ER 25-012 \citep{ogc2026genaiwildfire} & Operations & Wildfire & None listed & Not applicable & Over 200 Canadian datasets assessed; no model run \\ %
OGC D-123 \citep{tricomi2024ogcd123} & Operations & Wildfire & None listed & Not applicable & Lifecycle mapping and governance; no model run \\ %
CREW-Wildfire \citep{hyun2025crewwildfire} & Simulation & Wildfire & GPT-4o (1) & Grounded & Suppression and rescue below a do-nothing baseline \\ %
MOASEI wildfire track \citep{patino2025moasei} & Simulation & Wildfire & LLM meta-optimization loop & Grounded & Policies adapted; neural policies scored higher \\ %
BEACON system \citep{kulkarni2026beacon} & Communication & Wildfire & Claude, GPT (2) & Grounded & End-to-end evacuation agent demonstrated \\ %
BEACON dialog \citep{kulkarni2026beaconrs} & Communication & Wildfire & Claude (1) & Grounded & Multilingual evacuation dialog demonstrated \\ %
\bottomrule
\end{tabular}
\end{table*}

\section{Floats Referenced from the Body}\label{app:moved-floats}
This appendix holds the tables and figures that the main text cites.

\begin{table*}[tp]
\centering
\footnotesize
\setlength{\tabcolsep}{4pt}
\renewcommand{\arraystretch}{1.08}
\caption{\textbf{Ten working task categories organize the survey from perception to operations.} Assignments are provisional; Appendix~\ref{app:taxonomy} records boundary decisions.}\label{tab:taxonomy}
\begin{tabularx}{\textwidth}{@{}p{3.4cm} X p{5.0cm}@{}}
\toprule
Category & Definition & Example Re-Checked Fire Works \\
\midrule
knowledge-qa & Fire science, engineering, codes, or safety question answering. & \citep{chen2025regcheck, sun2026triplephase} \\
document-understanding & Incident reports, after-action reviews, records, and extraction. & \citep{epstein2025barriers, grunert2026taskstate, kim2025firespec, tao2026magrfi, chen2026disasteller} \\
detection-perception & Fire or smoke recognition from imagery or video by VLMs or MLLMs. & \citep{seidel2025forestfirevlm, qi2026smokebench, kohli2025zeroshot, habibpour2026wildfirevqa, ayanzadeh2026wildfirevlm, zhang2026disasterbench, almohannadi2026disastervqa} \\
forecasting-prediction & Spread, risk, danger, occurrence, or behavior prediction by LLMs or agents. & \citep{ramesh2025wildfiregpt, cheng2026retrieval, markov2026firescope} \\
geospatial-analysis & GIS reasoning, spatial computation, map or remote-sensing interpretation. & \citep{lahouel2026iceo} \\
decision-support-operations & Resource allocation, dispatch, evacuation, and incident command. & \citep{chen2025geospatial, ogc2026genaiwildfire} \\
simulation-coupled-agents & Agents that drive or interpret fire simulators. & \citep{hyun2025crewwildfire, patino2025moasei} \\
communication-alerts & Public warnings, crisis communication, and social media analysis. & \citep{kulkarni2026beacon, kulkarni2026beaconrs} \\
\bottomrule
\end{tabularx}
\vspace{1pt}
\par\noindent{\footnotesize Two further categories hold no re-checked fire records: \emph{data-retrieval-tools} (agents that query fire databases, APIs, or tools; tool use appears under other categories) and \emph{other} (fire tasks outside the preceding nine categories).}
\end{table*}

\begin{figure*}[tp]
\centering
\includegraphics[width=0.92\textwidth]{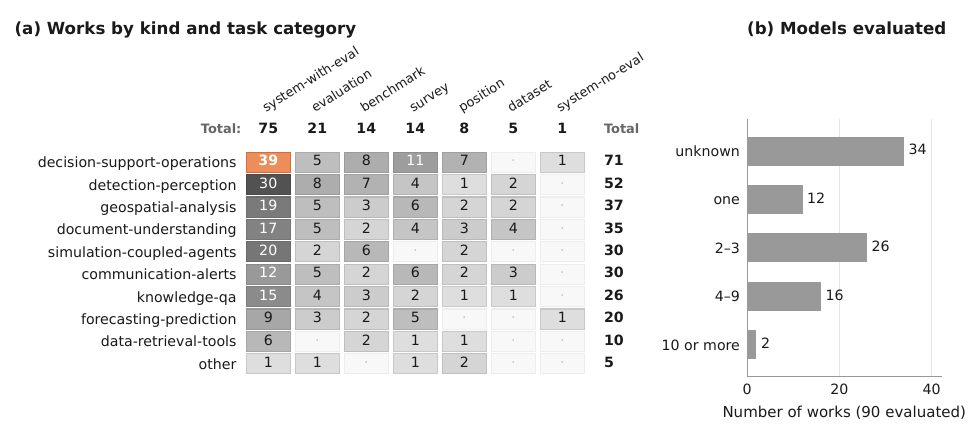}
\caption{\textbf{The 138 kept works by kind and task category, and the number of models the evaluated ones test.} (a)~Task category by kind of work over the 138 kept works (316 tags, since a work can carry several); coral marks the largest cell, the 39 systems with an evaluation tagged decision-support-operations. (b)~Number of models tested by the 90 works that a keyword screen marks as LLM work with an evaluation. Every count is a search-unit tag that was never re-checked.}\label{fig:survey-landscape}
\end{figure*}

\begin{table*}[tp]
\centering
\footnotesize
\setlength{\tabcolsep}{4pt}
\renewcommand{\arraystretch}{1.08}
\caption{\textbf{What the 25 re-checked fire records show, by task category: what they report, and what they leave unsettled.} Every row describes the re-checked subset alone, so no column is a prevalence estimate; Table~\ref{tab:bytask-full} in Appendix~\ref{app:bytask} gives one row per record.}\label{tab:bytask}
\vspace{2pt}
\begin{tabular}{@{}>{\raggedright\arraybackslash}p{1.20in} r >{\raggedright\arraybackslash}p{2.30in} >{\raggedright\arraybackslash}p{2.30in}@{}}
\toprule
Category & $n$ & What the records show & What they do not settle \\
\midrule
Detection and perception & 7 & Binary fire recognition scores high; early-stage smoke and quantitative estimates do not & The datasets fix camera set, prevalence, and smoke stage; no study shares a model set with another \\
Document understanding & 5 & Models sort incident records, command transcripts, and investigation cases at working scale & None of the five reports an error rate validated against human agreement \\
Operations and allocation & 3 & Geospatial grounding improved personnel and cost allocation & One study against baselines its authors chose; filed staffing is a proxy for decision quality \\
Forecasting and prediction & 3 & The one direct comparison put an LLM agent below tabular deep learning on fire radiative power & A single comparison on a GPT-4-era model; no ranking against other categories follows \\
Simulation-coupled agents & 2 & Coordination held on small suppression tasks and broke down on large ones & One base model; the second record is a separate result rather than a replication \\
Knowledge question answering & 2 & Rule-first and retrieval systems removed hallucination on numerical thresholds & Both are structure-fire code tasks, each against baselines its authors chose \\
Communication and alerts & 2 & Routing and alert generation work as system demonstrations & Neither reports an accuracy against a reference answer \\
Geospatial reasoning & 1 & One record, which is too few to state a pattern & Reported for completeness of the category set \\
\bottomrule
\end{tabular}
\end{table*}

\begin{table*}[tp]
\centering
\footnotesize
\setlength{\tabcolsep}{4pt}
\renewcommand{\arraystretch}{1.10}
\caption{\textbf{Four survey patterns, the design decision each led to, and what the tasks returned.}}\label{tab:crosswalk}
\begin{tabular}{@{}>{\raggedright\arraybackslash}p{1.35in}>{\raggedright\arraybackslash}p{1.45in}>{\raggedright\arraybackslash}p{3.10in}@{}}
\toprule
Survey pattern (Section~\ref{sec:crosscut}) & Design decision & Result (Section~\ref{sec:results}) \\
\midrule
Grounding helped in each re-checked study that tried it & Same models bare and grounded on every task, with an information-only comparator per source & Where an addition carried the answer and reached the model, every resolved difference was a gain; additions lacking it moved scores in both directions, and surveyed systems are not replicated (Section~\ref{sec:results}) \\
Recognition high; early smoke and quantities weak & Smoke frames scored by minutes since the plume & Smoke accuracy is 0.25 to 0.57 in the first ten minutes (Section~\ref{subsec:results-figlib}). On aerial question answering the thermal block moved pooled accuracy by at most 0.015, and no core model's paired interval against the 0.627 majority comparator lies wholly above zero (Section~\ref{subsec:results-wildfirevqa}) %
\\
One direct forecasting comparison favored a tabular model & Trivial rules and published reference points beside every model & Four of twelve prompted configurations beat a temperature rule, and trained tabular classifiers beat all twelve (Section~\ref{subsec:results-mesogeos}) \\
Evaluations hard to reconstruct; two records test models behind the frontier & Access, reference answers, and leakage audited; every response stored; frontier models of the run date & Three hazards in public releases and one in our harness (Section~\ref{subsec:source-audit}); 11 of 34 candidate sources not feasible \\
\bottomrule
\end{tabular}
\end{table*}

\begin{table*}[tp]
\centering
\footnotesize
\setlength{\tabcolsep}{4pt}
\renewcommand{\arraystretch}{1.10}
\caption{\textbf{No prompted model beats persistence in normalized error; the v1 analogues raise it on five of six models.} Daily personnel allocation on ICS-209-PLUS, 300 items, all answered. \emph{Persistence} repeats today's filed count; \emph{analogue-only} multiplies it by the median next-day ratio of the analogues the grounded arm saw; the \emph{trained regressor} fits that ratio on 1,499 other incidents (Appendix~\ref{app:trained-baselines}). The right columns split the items by whether the filed count moved by more than a tenth (116 moving days): \emph{false move} is the share of stable days moved by more than a tenth, \emph{missed move} the share of moving days held flat, \emph{dir.\ right} the share of moving days moved the same way. \emph{Norm.} is error normalized by each fire's mean staffing.}\label{tab:results-allocation}
\vspace{1pt}
\begin{tabular}{@{}ll rrrrrrrr@{}}
\toprule
Model & Condition & MAE & Norm. & Beats & Stable & Moving & False & Missed & Dir. \\
& & & & base & days & days & move & move & right \\
\midrule
\grouprow{10}{Reference points}
\comprow \multicolumn{2}{@{}l}{Persistence baseline} & 13.76 & 0.146 &  & 0.027 & 0.337 & 0.00 & 1.00 &  \\ %
\comprow \multicolumn{2}{@{}l}{Analogue-only rule} & 19.88 & 0.249 & 0.23 &  &  &  &  &  \\ %
\comprow \multicolumn{2}{@{}l}{Trained ratio regressor} & 12.97 & 0.144 & 0.48 & 0.035 & 0.317 & 0.03 & 0.93 & 0.72 \\ %
\midrule
\grouprow{10}{Proprietary models}
claude-opus-4.8 & bare & 14.31 & 0.157 & 0.29 & 0.061 & 0.310 & 0.14 & 0.66 & 0.55 \\ %
claude-opus-4.8 & grounded & 15.28 & 0.170 & 0.29 & 0.089 & 0.299 & 0.29 & 0.59 & 0.59 \\ %
claude-opus-5 & bare & 15.12 & 0.165 & 0.33 & 0.084 & 0.295 & 0.25 & 0.49 & 0.67 \\ %
claude-opus-5 & grounded & 14.52 & 0.173 & 0.32 & 0.099 & 0.290 & 0.27 & 0.53 & 0.63 \\ %
gemini-3.1-pro & bare & 16.17 & 0.179 & 0.19 & 0.090 & 0.318 & 0.23 & 0.54 & 0.43 \\ %
gemini-3.1-pro & grounded & 16.07 & 0.177 & 0.21 & 0.090 & 0.316 & 0.29 & 0.54 & 0.49 \\ %
gpt-6-astra & bare & 14.55 & 0.161 & 0.22 & 0.068 & 0.310 & 0.18 & 0.69 & 0.42 \\ %
gpt-6-astra & grounded & 18.04 & 0.220 & 0.25 & 0.134 & 0.356 & 0.30 & 0.64 & 0.51 \\ %
\midrule
\grouprow{10}{Open-weight models}
Qwen3-VL & bare & 14.73 & 0.161 & 0.07 & 0.049 & 0.339 & 0.08 & 0.86 & 0.16 \\ %
Qwen3-VL & grounded & 19.85 & 0.248 & 0.21 & 0.174 & 0.366 & 0.39 & 0.60 & 0.45 \\ %
Llama 4 Maverick & bare & 15.65 & 0.193 & 0.26 & 0.106 & 0.330 & 0.26 & 0.64 & 0.50 \\ %
Llama 4 Maverick & grounded & 19.88 & 0.246 & 0.21 & 0.167 & 0.371 & 0.37 & 0.60 & 0.45 \\ %
\bottomrule
\end{tabular}
\end{table*}

\begin{figure*}[tp]
\centering
\includegraphics[width=\textwidth]{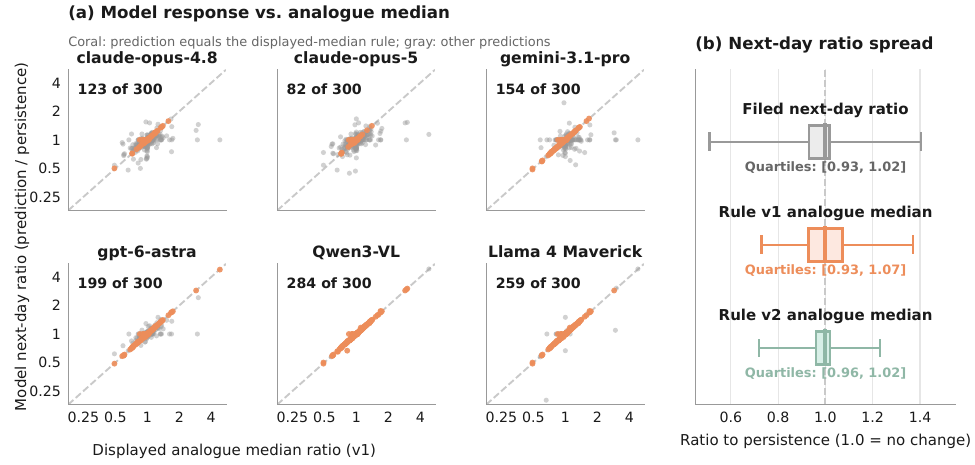}
\caption{\textbf{The two open-weight models match the displayed-median rule on most items; rule~v1 has a wider interquartile range than the filed next-day ratios, and rule~v2 narrows it.} (a)~Personnel allocation under retrieval rule~v1, 300 items per model: the model's next-day ratio (prediction over persistence) against the displayed median ratio of its six analogues, on $\log_2$ axes with the diagonal $y = x$ dashed. Coral marks a prediction equal to persistence times the displayed median, rounded to an integer, and each panel prints the count of such items. (b)~Across the 300 items, the filed next-day ratio (gray), the displayed median under rule~v1 (coral), and the displayed median under rule~v2 (mint): boxes span the quartiles, the bar is the median, whiskers run from the 5th to the 95th percentile, and the dashed line is persistence.}\label{fig:allocation-analogues}
\end{figure*}

\begin{table*}[tp]
\centering
\footnotesize
\setlength{\tabcolsep}{4pt}
\renewcommand{\arraystretch}{1.10}
\caption{\textbf{The reference frame raises smoke recall on every model and false positives on four.} Wildfire smoke detection on FIgLib, scored per model on the 196 items both of its conditions answered; no item was lost to truncation. \emph{Sequences detected} counts the smoke sequences in which at least one frame was called. The frame-difference detector thresholds the mean absolute difference against the reference frame, with the cut chosen by leave-one-fire-out cross-validation.}\label{tab:results-figlib}
\vspace{1pt}
\begin{tabular}{@{}ll rrrr r@{}}
\toprule
Model & Condition & Items & Accuracy & Recall on smoke & FPR & Sequences detected \\
\midrule
\grouprow{7}{Reference points}
\comprow \multicolumn{2}{@{}l}{Constant: always smoke} & 196 & 0.571 & 1.000 & 1.000 & 28 of 28 \\ %
\comprow \multicolumn{2}{@{}l}{Constant: always clear} & 196 & 0.429 & 0.000 & 0.000 & 0 of 28 \\ %
\comprow \multicolumn{2}{@{}l}{Frame difference, held out} & 196 & 0.770 & 0.920 & 0.429 & 28 of 28 \\ %
\midrule
\grouprow{7}{Proprietary models}
claude-opus-4.8 & bare & 196 & 0.745 & 0.554 & 0.000 & 23 of 28 \\ %
claude-opus-4.8 & grounded & 196 & 0.781 & 0.616 & 0.000 & 23 of 28 \\ %
claude-opus-5 & bare & 196 & 0.770 & 0.607 & 0.012 & 24 of 28 \\ %
claude-opus-5 & grounded & 196 & 0.781 & 0.643 & 0.036 & 24 of 28 \\ %
gemini-3.1-pro & bare & 196 & 0.740 & 0.545 & 0.000 & 23 of 28 \\ %
gemini-3.1-pro & grounded & 196 & 0.837 & 0.732 & 0.024 & 26 of 28 \\ %
gpt-6-astra & bare & 196 & 0.786 & 0.643 & 0.024 & 24 of 28 \\ %
gpt-6-astra & grounded & 196 & 0.832 & 0.741 & 0.048 & 26 of 28 \\ %
\midrule
\grouprow{7}{Open-weight models}
Qwen3-VL & bare & 196 & 0.740 & 0.545 & 0.000 & 22 of 28 \\ %
Qwen3-VL & grounded & 196 & 0.781 & 0.616 & 0.000 & 24 of 28 \\ %
Llama 4 Maverick & bare & 196 & 0.709 & 0.509 & 0.024 & 20 of 28 \\ %
Llama 4 Maverick & grounded & 196 & 0.719 & 0.554 & 0.060 & 21 of 28 \\ %
\bottomrule
\end{tabular}
\end{table*}

\begin{figure*}[tp]
\centering
\includegraphics[width=\textwidth]{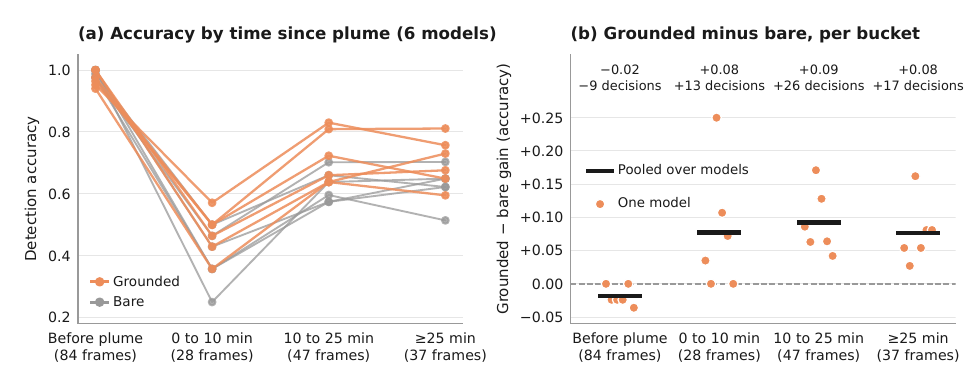}
\caption{\textbf{Detection is lowest in the first ten minutes after the plume. The reference frame adds about eight to nine accuracy points in each post-plume window, pooled over models.} (a)~Accuracy per offset bucket for each of the six models, bare (gray) and grounded (coral), on the 196 frames per model that both conditions answered; the frame counts under the axis are per model, and the three buckets before the plume are pooled into one. (b)~The grounded-minus-bare accuracy difference per bucket, one coral point per model and a black bar pooled over the six, with the pooled difference and the correct decisions it adds, summed over the six models, printed above each bucket. No intervals are drawn; Section~\ref{subsec:results-figlib} reports the fire-clustered intervals for the whole-task differences.}\label{fig:figlib-timeline}
\end{figure*}

\begin{table*}[tp]
\centering
\footnotesize
\setlength{\tabcolsep}{4pt}
\renewcommand{\arraystretch}{1.10}
\caption{\textbf{Two models rank above the temperature rule, and a classifier trained on the prompt's own numbers ranks above all twelve.} Fire danger forecasting on Mesogeos Track A, 386 items, sample positive rate 0.339. The \emph{temperature rule} is the best trivial strategy found; the \emph{trained} rows read the 121 numbers the bare prompt prints (Appendix~\ref{app:trained-baselines}); \emph{call rate} is the share of items answered yes; \emph{omitted} counts items with a probability but no yes or no answer, filled from the probability at one half. In claude-opus-5's grounded arm those omissions fall on 29 percent of non-fire items against 12 percent of fire items ($p = 0.0002$, two-sided Fisher exact test), so its grounded decision column rests on that imputation. Its bare rates are 13 and 11 percent, with no resolved association (Appendix~\ref{app:answers-hazards}).}\label{tab:results-mesogeos}
\vspace{1pt}
\begin{tabular}{@{}ll rrrr@{}}
\toprule
Model & Condition & AUPRC & F1 on fire & Call rate & Omitted \\
\midrule
\grouprow{6}{Reference points}
\comprow \multicolumn{2}{@{}l}{Calendar-month prior, training years} & 0.555 & & & \\ %
\comprow \multicolumn{2}{@{}l}{Last-day maximum 2 m temperature} & 0.654 & & & \\ %
\comprow \multicolumn{2}{@{}l}{Published baselines, full holdout} & 0.853 to 0.858 & & & \\ %
\comprow \multicolumn{2}{@{}l}{Trained boosted classifier, prompt numbers} & 0.851 & 0.783 & 0.316 & 0 \\ %
\comprow \multicolumn{2}{@{}l}{Trained logistic regression, prompt numbers} & 0.768 & 0.697 & 0.285 & 0 \\ %
\midrule
\grouprow{6}{Proprietary models}
claude-opus-4.8 & bare & 0.613 & 0.561 & 0.215 & 0 \\ %
claude-opus-4.8 & grounded & 0.581 & 0.546 & 0.249 & 2 \\ %
claude-opus-5 & bare & 0.688 & 0.269 & 0.065 & 46 \\ %
claude-opus-5 & grounded & 0.682 & 0.312 & 0.109 & 90 \\ %
gemini-3.1-pro & bare & 0.697 & 0.607 & 0.207 & 0 \\ %
gemini-3.1-pro & grounded & 0.682 & 0.587 & 0.199 & 0 \\ %
gpt-6-astra & bare & 0.620 & 0.189 & 0.044 & 0 \\ %
gpt-6-astra & grounded & 0.590 & 0.298 & 0.078 & 0 \\ %
\midrule
\grouprow{6}{Open-weight models}
Qwen3-VL & bare & 0.485 & 0.567 & 0.684 & 0 \\ %
Qwen3-VL & grounded & 0.536 & 0.553 & 0.710 & 0 \\ %
Llama 4 Maverick & bare & 0.528 & 0.513 & 0.256 & 0 \\ %
Llama 4 Maverick & grounded & 0.515 & 0.524 & 0.303 & 0 \\ %
\bottomrule
\end{tabular}
\end{table*}

\begin{figure*}[tp]
\centering
\includegraphics[width=\textwidth]{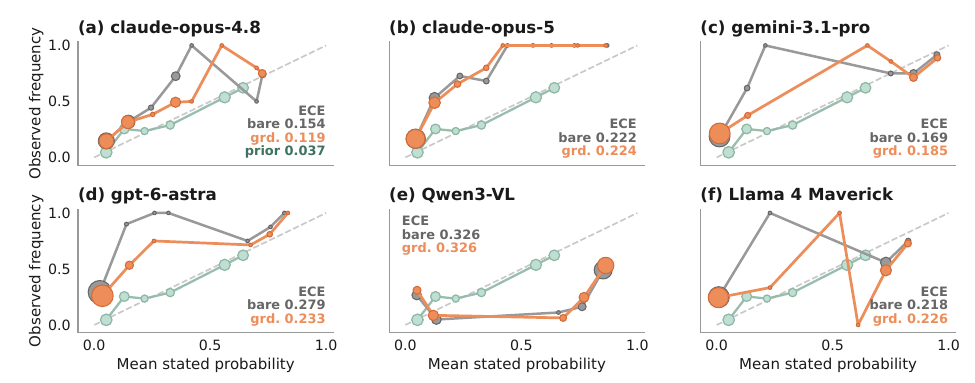}
\caption{\textbf{Every proprietary model states mean probabilities of 0.06 to 0.22 against the 0.339 benchmark sample positive rate, underpredicting fire on this selected population. The calendar-month prior has a lower ECE than all twelve model runs.} Reliability diagrams on the 386 fire-danger items: the mean stated probability in each of ten equal-width bins against the observed fire frequency in that bin, bare in gray and grounded in coral; marker area is $\max(0.35n, 3)$ square points for a bin of $n$ items, so bins of fewer than nine items share the smallest marker, and empty bins are omitted; the dashed diagonal is perfect calibration. The calendar-month prior (mint, the same curve on every panel) was fitted on identically sampled training data. Its expected calibration error (ECE) is 0.037 against 0.119 to 0.326 for the twelve model runs. After one odds shift to the sample rate, claude-opus-5 falls below the prior on Brier score in both arms and no other proprietary interval excludes zero (Table~\ref{tab:recalibration}).}\label{fig:calibration}
\end{figure*}

\begin{figure*}[tp]
\centering
\includegraphics[width=\textwidth]{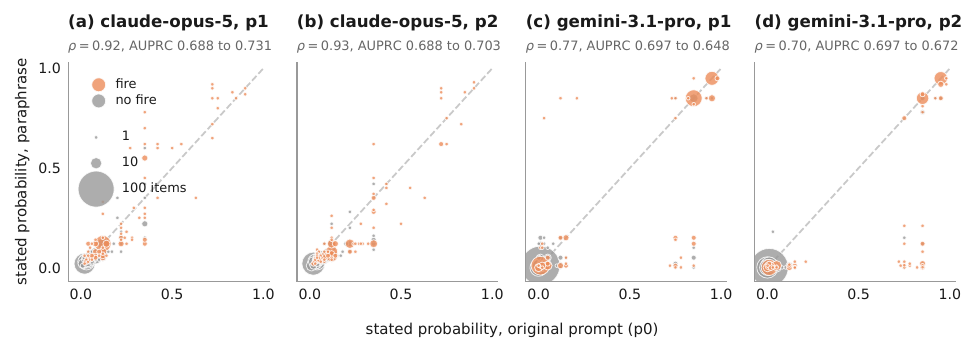}
\caption{\textbf{Single runs of two paraphrases move the two leading models in opposite directions and reverse their order; the paraphrase reorders gemini-3.1-pro's items far more than claude-opus-5's.} Stated fire probability on the 386 Mesogeos Track A items under the paper's prompt (p0, horizontal) against a paraphrase (vertical) for claude-opus-5 (a, b) and gemini-3.1-pro (c, d). For each label, one bubble represents the items that share a distinct pair of stated values, with area proportional to that count; fire items ($n = 131$) are coral and no-fire items ($n = 255$) gray, and bubbles of the two labels can share a center, with coral drawn above gray. The dashed diagonal marks unchanged probabilities. The subtitle of each panel gives the Spearman correlation between the two runs and the AUPRC under p0 and under the paraphrase. Paired AUPRC differences carry 95 percent intervals from a cluster bootstrap over 1$^\circ$ blocks by calendar month (20,000 resamples). They are 0.043 [0.009, 0.080] and 0.015 [$-$0.018, 0.049] for claude-opus-5, and $-$0.049 [$-$0.093, $-$0.009] and $-$0.026 [$-$0.075, 0.020] for gemini-3.1-pro (Table~\ref{tab:prompt-sensitivity}, which adds the other four models). Each paraphrase ran once, bare.}\label{fig:prompt-sensitivity}
\end{figure*}

\begin{table*}[tp]
\centering
\footnotesize
\setlength{\tabcolsep}{4pt}
\renewcommand{\arraystretch}{1.10}
\caption{\textbf{One read-only SQL tool lifts every core model from at most 0.160 accuracy to 0.885 or above, and the four proprietary models to 0.994 or above. Most open-weight errors fall in the calendar-window count family.} Fire data tool use on FPA-FOD, 156 items in twelve families, one run per arm. \emph{Best constant} is the one fixed answer per family that scores highest under the family's tolerance, chosen with the answers in hand. \emph{Abstained} counts \texttt{unknown} answers, scored as wrong. \emph{Calls} is the mean number of tool calls per item. \emph{Tool $-$ bare} is the paired accuracy difference, with a 95 percent bootstrap interval clustered by family (12 clusters, 20,000 resamples).}\label{tab:results-tooluse}
\begin{tabular}{@{}ll rrrr rr l@{}}
\toprule
Model & Arm & Accuracy & \begin{tabular}[b]{@{}r@{}}Single\\lookup\end{tabular} & \begin{tabular}[b]{@{}r@{}}Filtered\\aggregate\end{tabular} & Multi-step & Abstained & Calls & Tool $-$ bare \\
\midrule
\grouprow{9}{Reference point}
\comprow \multicolumn{2}{@{}l}{Best constant per family} & 0.141 & 0.077 & 0.077 & 0.231 & & & \\ %
\midrule
\grouprow{9}{Proprietary models}
claude-opus-4.8 & bare & 0.103 & 0.000 & 0.000 & 0.246 & 99 & 0.00 & \\ %
claude-opus-4.8 & tool & 1.000 & 1.000 & 1.000 & 1.000 & 0 & 1.13 & 0.897 [0.756, 1.000] \\ %
claude-opus-5 & bare & 0.160 & 0.000 & 0.038 & 0.338 & 35 & 0.00 & \\ %
claude-opus-5 & tool & 1.000 & 1.000 & 1.000 & 1.000 & 0 & 1.17 & 0.840 [0.660, 0.974] \\ %
gemini-3.1-pro & bare & 0.160 & 0.000 & 0.026 & 0.354 & 59 & 0.00 & \\ %
gemini-3.1-pro & tool & 1.000 & 1.000 & 1.000 & 1.000 & 0 & 1.62 & 0.840 [0.692, 0.974] \\ %
gpt-6-astra & bare & 0.000 & 0.000 & 0.000 & 0.000 & 156 & 0.00 & \\ %
gpt-6-astra & tool & 0.994 & 1.000 & 0.987 & 1.000 & 0 & 1.07 & 0.994 [0.981, 1.000] \\ %
\midrule
\grouprow{9}{Open-weight models}
Qwen3-VL & bare & 0.083 & 0.000 & 0.000 & 0.200 & 22 & 0.00 & \\ %
Qwen3-VL & tool & 0.891 & 1.000 & 0.872 & 0.892 & 6 & 1.08 & 0.808 [0.654, 0.936] \\ %
Llama 4 Maverick & bare & 0.071 & 0.000 & 0.000 & 0.169 & 0 & 0.00 & \\ %
Llama 4 Maverick & tool & 0.885 & 1.000 & 0.833 & 0.923 & 0 & 1.00 & 0.814 [0.635, 0.949] \\ %
\bottomrule
\end{tabular}
\end{table*}

\begin{table*}[tp]
\centering
\footnotesize
\setlength{\tabcolsep}{5pt}
\renewcommand{\arraystretch}{1.05}
\caption{\textbf{Surface rewording moves neither open-weight model's interval clear of zero; removing schema literals leaves four proprietary models unchanged and costs two open-weight models 0.175 and 0.126 on the clean 143-item subset.} Accuracy under original items, a phrasing rewrite, and a de-schematized rewrite. The table reports all 156 items (where open models drop 0.199 and 0.154); the footnote reports the clean 143 items. Intervals expand the clustered 95 percent percentile intervals by the sensitivity procedure of Appendix~\ref{app:metrics-detail}. The 39-column schema reaches models under all three.}\label{tab:tooluse-paraphrase}
\vspace{1pt}
\begin{tabular}{@{}l rrr rr@{}}
\toprule
Model & Original & Phrasing & De-schem. & Phrasing $-$ orig. & De-schem. $-$ orig. \\
\midrule
\grouprow{6}{Proprietary models}
claude-opus-4.8 & 1.000 & 1.000 & 1.000 & +0.000 & +0.000 \\ %
claude-opus-5 & 1.000 & 1.000 & 1.000 & +0.000 & +0.000 \\ %
gemini-3.1-pro & 1.000 & 1.000 & 1.000 & +0.000 & +0.000 \\ %
gpt-6-astra & 0.994 & 0.994 & 0.994 & +0.000 & +0.000 \\ %
\midrule
\grouprow{6}{Open-weight models}
Qwen3-VL & 0.891 & 0.923 & 0.692 & +0.032 & $-$0.199 \\ %
 & & & & {\scriptsize [$-$0.008, +0.098]} & {\scriptsize [$-$0.296, $-$0.108]} \\ %
Llama 4 Maverick & 0.885 & 0.917 & 0.731 & +0.032 & $-$0.154 \\ %
 & & & & {\scriptsize [$-$0.022, +0.099]} & {\scriptsize [$-$0.310, $-$0.017]} \\ %
\bottomrule
\end{tabular}
\par\noindent{\footnotesize The clean subset excludes the ambiguous \texttt{count\_county\_year} family, leaving 143 items across eleven families. Qwen3-VL falls 0.175 [$-$0.266, $-$0.091]; Llama 4 Maverick falls 0.126 [$-$0.284, $+$0.025]. Only Qwen3-VL's displayed interval excludes zero. The four proprietary models are unchanged item for item. The table reports all 156 items for comparison.}
\end{table*}

\begin{table*}[tp]
\centering
\footnotesize
\setlength{\tabcolsep}{3.5pt}
\renewcommand{\arraystretch}{1.08}
\caption{\textbf{Among the sixteen models evaluated on all five tasks, no bare model beats persistence. Analogues raise allocation error on thirteen, a reference frame raises smoke recall on fifteen, and the tool lifts the fourteen whose interface worked. No added model reaches the aerial majority baseline.} The model sweep of 2026-09-18 beside the core six: point estimates per model and arm on the same items and scorers as Tables~\ref{tab:results-allocation}, \ref{tab:results-figlib}, \ref{tab:results-mesogeos}, \ref{tab:results-tooluse}, and~\ref{tab:results-wildfirevqa}. Allocation: normalized error, lower is better. Smoke: recall on the smoke frames of the items both arms answered, 112 of 196 for fifteen models and 111 of 194 for Mistral Large 3. Fire danger: AUPRC. Tool use: accuracy bare and with the tool. Aerial: accuracy over 408 items. The comparator row holds persistence, the temperature rule, the best constant per family, and the majority baseline. Nova 2 Lite's grounded allocation error is 0.248 without one item on which it copied an analogue's 866 personnel onto a two-person incident. Gemma 3 12B and 4B score 0.000 with the tool because neither executed a valid database query under the declared interface (Appendix~\ref{app:model-sweep}). Appendix~\ref{app:model-sweep} adds the false-positive and fire-call rates.}\label{tab:model-sweep}
\vspace{1pt}
\begin{tabular}{@{}l cc cc cc cc cc@{}}
\toprule
& \multicolumn{2}{c}{Allocation nMAE} & \multicolumn{2}{c}{Smoke recall} & \multicolumn{2}{c}{Fire danger AUPRC} & \multicolumn{2}{c}{Tool use acc.} & \multicolumn{2}{c}{Aerial acc.} \\
\cmidrule(lr){2-3}\cmidrule(lr){4-5}\cmidrule(lr){6-7}\cmidrule(lr){8-9}\cmidrule(lr){10-11}
Model & bare & grd. & bare & grd. & bare & grd. & bare & tool & bare & grd. \\
\midrule
\grouprow{11}{Core set (Section~\ref{subsec:model-set})}
claude-opus-4.8 & 0.157 & 0.170 & 0.554 & 0.616 & 0.613 & 0.581 & 0.103 & 1.000 & 0.642 & 0.642 \\ %
claude-opus-5 & 0.165 & 0.173 & 0.607 & 0.643 & 0.688 & 0.682 & 0.160 & 1.000 & 0.650 & 0.657 \\ %
gemini-3.1-pro & 0.179 & 0.177 & 0.545 & 0.732 & 0.697 & 0.682 & 0.160 & 1.000 & 0.632 & 0.647 \\ %
gpt-6-astra & 0.161 & 0.220 & 0.643 & 0.741 & 0.620 & 0.590 & 0.000 & 0.994 & 0.623 & 0.635 \\ %
Qwen3-VL & 0.161 & 0.248 & 0.545 & 0.616 & 0.485 & 0.536 & 0.083 & 0.891 & 0.618 & 0.618 \\ %
Llama 4 Maverick & 0.193 & 0.246 & 0.509 & 0.554 & 0.528 & 0.515 & 0.071 & 0.885 & 0.534 & 0.544 \\ %
\midrule
\grouprow{11}{Added on 2026-09-18 through Bedrock}
Nova Lite & 0.272 & 0.250 & 0.446 & 0.527 & 0.417 & 0.377 & 0.019 & 0.885 & 0.527 & 0.556 \\ %
Nova Pro & 0.171 & 0.242 & 0.589 & 0.652 & 0.477 & 0.479 & 0.090 & 0.827 & 0.554 & 0.554 \\ %
Nova 2 Lite & 0.163 & 1.687 & 0.321 & 0.223 & 0.442 & 0.478 & 0.058 & 0.865 & 0.488 & 0.507 \\ %
Llama 4 Scout & 0.230 & 0.246 & 0.384 & 0.482 & 0.473 & 0.460 & 0.032 & 0.910 & 0.588 & 0.566 \\ %
Mistral Large 3 & 0.260 & 0.225 & 0.414 & 0.577 & 0.473 & 0.442 & 0.083 & 0.923 & 0.515 & 0.559 \\ %
Ministral 3 8B & 0.208 & 0.250 & 0.384 & 0.491 & 0.467 & 0.420 & 0.071 & 0.833 & 0.515 & 0.515 \\ %
Kimi K2.5 & 0.173 & 0.254 & 0.420 & 0.545 & 0.518 & 0.529 & 0.115 & 0.942 & 0.583 & 0.613 \\ %
Gemma 3 27B & 0.169 & 0.252 & 0.554 & 0.688 & 0.532 & 0.466 & 0.058 & 0.827 & 0.480 & 0.532 \\ %
Gemma 3 12B & 0.177 & 0.270 & 0.589 & 0.598 & 0.446 & 0.389 & 0.058 & 0.000 & 0.429 & 0.453 \\ %
Gemma 3 4B & 0.147 & 0.175 & 0.536 & 0.607 & 0.347 & 0.339 & 0.058 & 0.000 & 0.392 & 0.404 \\ %
\midrule
\comprow Comparator & 0.146 & & & & 0.654 & & 0.141 & & 0.627 & \\ %
\bottomrule
\end{tabular}
\end{table*}

\begin{figure*}[tp]
\centering
\includegraphics[width=\textwidth]{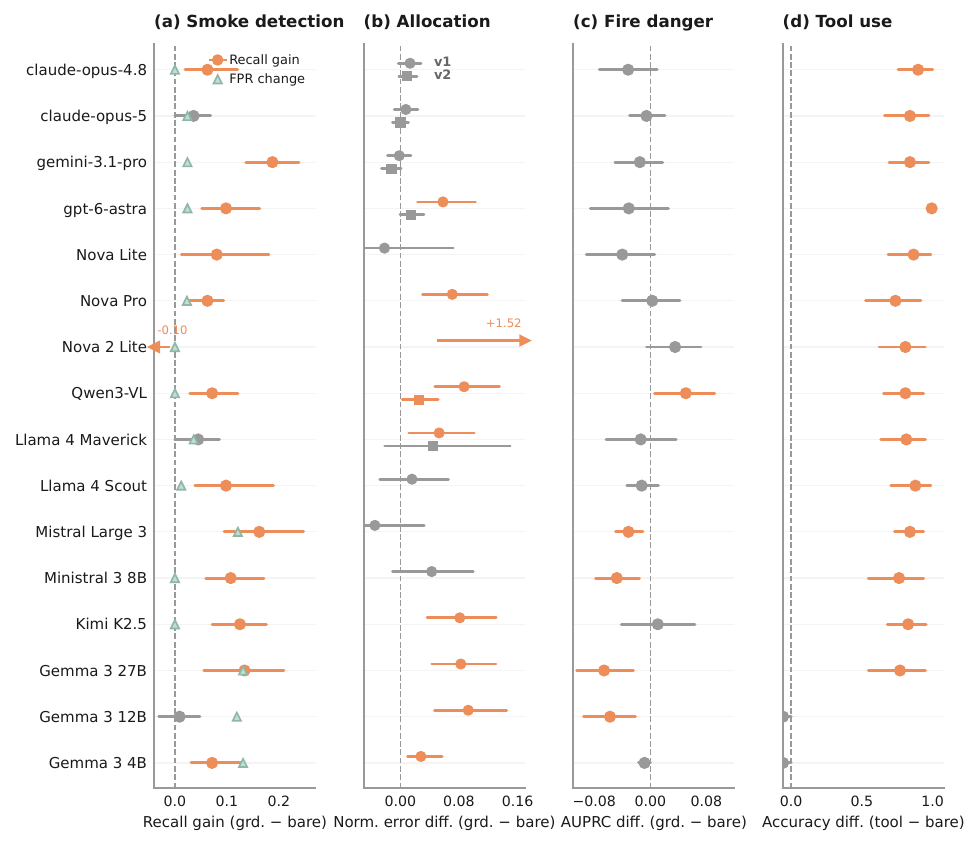}
\caption{\textbf{Across sixteen models, intervals exclude zero for the reference frame's recall gain on twelve, the analogues' error increase on nine, and the tool's lift on fourteen. Climatology's AUPRC change excludes zero on five models, in both directions.} Grounded minus bare per model on four tasks, with 95 percent cluster bootstrap intervals as in Figure~\ref{fig:grounding-effects}; the ten added models carry retrieval rule v1 only. Individual 95 percent intervals carry no multiplicity adjustment across the sixty-four contrasts, so these counts are descriptive summaries rather than formal joint inferences. Coral marks an interval excluding zero. Nova 2 Lite's allocation point (+1.52) lies beyond the axis (one analogue copy; Table~\ref{tab:model-sweep}), and its recall change is the one negative interval.}\label{fig:grounding-sweep}
\end{figure*}

\section{Extended Survey of Existing Fire Tasks and Related Work}\label{sec:extended-background}

\subsection{Scope and Search Corpus}

Searches for work first released between 2025-01-01 and 2026-09-13 kept 138 works, with one dating exception (Appendix~\ref{sec:method}), and tagged each by task category (Figure~\ref{fig:survey-landscape}). %
Broader disaster or Earth-science work enters scope when it includes at least one fire task, while disaster work without a fire task is cited only as adjacent work \citep{thomas2026imdg,ticona2026translation}. %
Table~\ref{tab:taxonomy} defines the ten working task categories and lists example re-checked fire works \citep[e.g.,][]{ayanzadeh2026wildfirevlm}. %
The 31 records cited by the scoping summary were re-checked claim by claim, 25 of them with a fire task. %
Table~\ref{tab:bytask} gives what those records show by category and what they leave unsettled, and Table~\ref{tab:bytask-full} in Appendix~\ref{app:bytask} carries one row per record. %

\subsection{Survey Patterns Driving Benchmark Design}

The paragraphs below state the patterns found in the 25 re-checked fire records; the scoping summary cited these records, so every pattern describes that subset alone. %

\paragraph{Grounding Across Tasks and Models.}\label{subsec:patterns-intro}\label{subsec:grounding-helps}
Grounding helped in each re-checked study that tried it, six studies in all, each against baselines its authors chose. %
It improved wildfire risk evaluation, post-wildfire damage assessment, personnel and cost allocation, fire code compliance, and regulation retrieval \citep{cheng2026retrieval,lahouel2026iceo,chen2025geospatial,chen2025regcheck,sun2026triplephase}. %
Retrieval also assisted the stronger tested MLLMs on aerial wildfire questions \citep{habibpour2026wildfirevqa}. %

\paragraph{Perception, Localization, and Tuned Models.}\label{subsec:recognition-localization}\label{subsec:small-tuned-models}
Four re-checked perception studies evaluate vision models on fire imagery. %
Binary fire recognition scores high, and most evaluated models classified large smoke plumes well \citep{kohli2025zeroshot,almohannadi2026disastervqa,qi2026smokebench}. %
Models in these studies localized early smoke poorly, erred on quantities, and showed deficits in temperature-grounded reasoning \citep{qi2026smokebench,almohannadi2026disastervqa,habibpour2026wildfirevqa}. %
Small domain-tuned models can match larger general models: fine-tuned ForestFireVLM-7B surpassed Gemini 2.0 Pro on forest fire scenes \citep{seidel2025forestfirevlm}, and DisasterVL matched GPT-4o on disaster reasoning \citep{zhang2026disasterbench}. %
Smaller reasoning models also matched larger ones on resource allocation \citep{chen2025geospatial}. %

\paragraph{Quantitative Forecasting, Multi-Agent Teams, and Fire Records.}\label{subsec:forecasting-weakest}\label{subsec:multiagent-breakdown}\label{subsec:analysts-fire-records}
The one direct forecasting comparison placed an LLM agent below tabular deep learning on fire radiative power \citep{ramesh2025wildfiregpt}. %
Chain-of-thought reasoning guided a wildfire risk predictor in FireScope and improved out-of-distribution generalization \citep{markov2026firescope}. %
In simulation, one study found that multi-agent frameworks coordinated on simple tasks while coordination broke down on large suppression and rescue problems \citep{hyun2025crewwildfire}. %
That study tested one base model, and the second record \citep{patino2025moasei} is a separate result rather than a replication. %
Three studies apply LLMs to incident records, command transcripts, and investigation cases without reporting validated error rates against human agreement \citep{epstein2025barriers,tao2026magrfi,grunert2026taskstate}. %

\paragraph{Evaluation Hygiene and Multi-Hazard Context.}\label{subsec:model-choice-trails-frontier}\label{subsec:separate-literatures}\label{subsec:isolated-evaluations}
Across the 25 re-checked records, two evaluate models behind the frontier \citep{ayanzadeh2026wildfirevlm,habibpour2026wildfirevqa}, and four test GPT-5-class models \citep{chen2025geospatial,markov2026firescope,zhang2026disasterbench,grunert2026taskstate}. %
Five evaluations test one system against self-chosen baselines \citep{cheng2026retrieval,lahouel2026iceo,chen2025geospatial,chen2025regcheck,sun2026triplephase}; the 25 were not coded for this, so it is a count of examples rather than a majority. %

\section{Closest Works, the Positioning Table, and the Late Find}\label{sec:related-work-detail}

Prior fire-related AI includes disaster surveys \citep{lei2025disasterllm}, architecture reports without model runs \citep{tricomi2024ogcd123,ogc2026genaiwildfire}, vision benchmarks without language models \citep{xu2026wildfirefm,tran2025land8fire,zhao2025tssatfire}, and narrow model evaluations \citep{hyun2025crewwildfire,zhang2026disasterbench,chen2025geospatial}. %
No study found varies grounding strategies across task families on a single shared model set (Appendix~\ref{app:bytask}). %
Checks after the window closed found three works released inside it. %
ORCH \citep{ji2026orch}, released on 2026-09-10, evaluates eight LLMs on 25 simulated wildfire-response missions and falls under simulation-coupled agents. %
DORA \citep{wang2026dora}, released on 2026-05-12, evaluates 13 LLM agents on 515 expert-authored disaster-response tasks over 45 disaster events with a 108-tool geospatial library. Wildfire is one of its ten disaster types. %
FlameVQA \citep{habibpour2026flamevqa} was released on 2026-06-25 by four of the five WildFireVQA authors. It builds 34 multiple-choice questions per frame on about 6,100 FLAME~3 RGB and thermal pairs and evaluates two open-weight MLLMs. Gemini 2.5 Pro proposes its labels, which fixed radiometric thresholds and expert review then check; its text never mentions clipping or saturation (Appendix~\ref{app:defect-ceiling}). %
None is in the survey counts. %
DORA names disaster-specific failure modes with their shares, and its gold tool-order hints raise final accuracy by only 1.08 to 4.40 points. Table~\ref{tab:failure-modes} and insight I1 adapt both ideas to fire tasks. %

\begin{table*}[t]
\centering
\footnotesize
\setlength{\tabcolsep}{3.8pt}
\renewcommand{\arraystretch}{1.08}
\caption{\textbf{None of the listed prior works combines a re-checked literature survey with a bare-against-grounded evaluation of several model families across fire tasks.} Prior fire surveys, benchmarks, and LLM evaluations, one row each.}\label{tab:positioning-full}
\vspace{1pt}
\begin{tabular}{@{} >{\raggedright\arraybackslash}p{4.6cm} cccccc r@{}}
\toprule
Work & Surv. & Tasks & LLMs & Fam. & B/G & Non-LLM & Scale \\
\midrule
\grouprow{8}{Fire AI reports with no model evaluation}
OGC D-123 \citep{tricomi2024ogcd123} & -- & -- & -- & -- & -- & -- & NIMS/ICS phases \\
OGC 25-012 \citep{ogc2026genaiwildfire} & -- & -- & -- & -- & -- & -- & $>$200 datasets \\
\midrule
\grouprow{8}{Fire benchmarks and datasets that evaluate no LLM or agent}
WILDFIRE-FM \citep{xu2026wildfirefm} & -- & 1 & -- & -- & -- & \checkmark & 6 task forms, 11 FMs \\ %
Land8Fire \citep{tran2025land8fire} & \checkmark\textsuperscript{$\circ$} & 1 & -- & -- & -- & \checkmark & 4 nets, 55 GB \\
TS-SatFire \citep{zhao2025tssatfire} & \checkmark\textsuperscript{$\circ$} & 2 & -- & -- & -- & \checkmark & 3 tasks, 11 models \\
\midrule
\grouprow{8}{LLM and agent evaluations on fire tasks}
CREW-Wildfire \citep{hyun2025crewwildfire} & -- & 1 & 4 & 1 & -- & \checkmark & 4 frameworks \\
DisasterBench \citep{zhang2026disasterbench} & -- & 3 & 22 & 8 & -- & -- & 9 tasks, 402 fire inst. \\ %
Geospatial Agents \citep{chen2025geospatial} & -- & 1 & 14 & 3 & \checkmark & \checkmark & 5 fires, 2 tasks \\
SmokeBench \citep{qi2026smokebench} & -- & 1 & 7 & 6 & -- & \checkmark & 7 MLLMs, 2 detectors \\ %
WildfireGPT \citep{ramesh2025wildfiregpt} & -- & 1 & 1 & 1 & -- & \checkmark & 1 task, 2 models \\
WildFireVQA \citep{habibpour2026wildfirevqa} & -- & 1 & 4 & 4 & \checkmark & -- & 207{,}298 items \\
DisasterVQA \citep{almohannadi2026disastervqa} & -- & 1 & 7 & 5 & -- & -- & 669 fire questions \\ %
\midrule
\rowcolor{abCoral!25}AI4Fire (this paper) & \checkmark & 4 & 35 & 12 & \checkmark & \checkmark & 1{,}446 items, 5 tasks \\ %
\bottomrule
\end{tabular}
\vspace{2pt}
\par\noindent{\footnotesize \checkmark\ present; \checkmark\textsuperscript{$\circ$}\ partial survey without claim re-checks; -- absent or unstated. \emph{Tasks}: evaluated fire task categories from Table~\ref{tab:taxonomy}. \emph{LLMs}: evaluated LLM or agent configurations, a paper's own model included and each size counted once. \emph{Fam.}: model vendor or architecture families. \emph{B/G}: compares bare and grounded arms on the same items. \emph{Non-LLM}: scores a trained, rule, or do-nothing comparator on the same items. Scale preserves each paper's unit. Of AI4Fire's 35 models, sixteen ran all five tasks, seventeen ran allocation, fire danger, and tool use, and two ran allocation and fire danger. The column counts each paper's model roster rather than a complete model-by-task evaluation (Section~\ref{subsec:results-sweep}).}
\end{table*}

\section{Extended Results, Per-Task Analyses, and Model Sweep}\label{sec:extended-results}\label{app:extended-results}

\subsection{Failure Modes in Detail}\label{app:failure-modes}

Table~\ref{tab:failure-modes} lists seven diagnostic patterns; here we define each, add two omitted patterns, and quote stored responses for model patterns and the unit gloss. %
A pattern counts as a model failure when a fair reading of the prompt leads away from the model's answer. It counts as a property of the benchmark or its source when the item, its label, or the harness invites the error. %

\paragraph{Median Copying (Allocation).}
The grounded prompt displays retrieved analogues and their median staffing ratio; a forecast copies when equaling today's count times that median, rounded. %
Qwen3-VL and Llama 4 Maverick copy on 284 and 259 of 300 items, and gpt-6-astra on 199 (Table~\ref{tab:retrieval-v2}). %
The rule scores 0.249 normalized error against 0.146 for persistence; under retrieval rule~v2 the two open-weight models still copy on 283 and 258 items. %
On the Parker Ridge fire (2015-08-19), Qwen3-VL answered 108, reasoning that ``the median historical ratio of 1.38 \ldots{} multiplying by today's 78 personnel yields approximately 108.'' The filed count was 78. %

\paragraph{Guessed Date Format (Tool Use).}
\texttt{DISCOVERY\_DATE} holds unpadded \texttt{m/d/yyyy} text, unstated in schema or question. %
Thirty of 35 open-weight tool failures filter with zero-padded or ISO dates and accept zero results (Appendix~\ref{app:tooluse-detail}). %
The proprietary four answer all 26 date-window items correctly by inspecting stored dates, using day-of-year, or correcting empty queries. %
Asked for Arizona fires discovered between 1 June and 31 August 2001, Qwen3-VL filtered \texttt{DISCOVERY\_DATE BETWEEN \textquotesingle 06/01/2001\textquotesingle{} AND \textquotesingle 08/31/2001\textquotesingle} and received 0. It answered 0 against reference 2,081. %

\paragraph{Unit Gloss (Aerial, Our Harness).}
The thermal block's pct fields hold percentages, but the prompt glosses them as shares of pixels, introducing 9 closed-form misses across open-weight models (Appendix~\ref{app:defect-ours}). %
Seven predictions match fraction binning, one bins the other field, and one answers None for a displayed zero rounding one hot pixel. %
With 0.703 percent of pixels above 200\,$^\circ$C, Qwen3-VL answered more than 15 percent on DS8, having answered below 5 percent bare. The block's hottest-3-percent mean of 172.6\,$^\circ$C rules out that reading, but the gloss invites it, making the pattern a harness property. %

\paragraph{Clipped Maximum and Ambiguous Wording (Aerial).}
Appendix~\ref{app:defect-ceiling} documents the clipping, and Appendix~\ref{app:closed-form-misses} breaks down the 34 closed-form misses, including 9 fitting an area reading of ``dominant fire behavior''. %

\paragraph{Time-Gap Cue (Smoke) and Unstated Base Rate (Fire Danger).}
Appendix~\ref{app:figlib-timing} gives reference-to-target time gaps and a rerun that matches them across labels. Appendix~\ref{app:recalibration} shows what a prior shift to the unstated 0.339 rate repairs. %

\paragraph{Two Further Patterns.}
Literal dependence (tool use, model failure): plain-word rewrites cost Qwen3-VL 0.175 and Llama 4 Maverick 0.126 on the 143 clean items. The proprietary four answer identically item for item (Table~\ref{tab:tooluse-paraphrase}). %
Output cap (tool use, our harness): the cap cut 61 of gemini-3.1-pro's 156 bare answers before the answer line (Appendix~\ref{app:defect-ours}). %

\subsection{Extended Analysis: Daily Personnel Allocation}\label{subsec:app-allocation-detail}

The task predicts next-day filed personnel from one situation report across 300 items drawn from 245 incidents in 19 states (Table~\ref{tab:results-allocation}). %
The grounded arm adds up to six analogues from the same size, containment, and staffing band with subsequent outcomes. %
Evaluation clusters by incident (245 clusters), disjoint from the analogue pool. %
An analogue is eligible only if its input and outcome days precede the report day; one item was re-queried under this rule (Appendix~\ref{app:conditions-detail}). %

\paragraph{Persistence Baseline and Regressor Training.}
No prompted model beats persistence on either error measure, scoring 0.157 to 0.248 normalized error bare and grounded against 0.146 for persistence (Table~\ref{tab:results-allocation}). %
A gradient-boosted regressor trained on 30,869 fire-days from 1,499 other incidents, reading the same report fields, matches persistence at 0.144 normalized error (difference $-0.002$ [$-0.008$, 0.003]; Appendix~\ref{app:trained-baselines}). %

\paragraph{Analogue Copying Mechanism and Agreement.}
Retrieved analogues raise normalized error on five of six models (Table~\ref{tab:results-allocation}). %
Grounded-minus-bare error differences exclude zero for gpt-6-astra (0.058 [0.023, 0.103]), Qwen3-VL (0.087 [0.047, 0.135]), and Llama 4 Maverick (0.053 [0.012, 0.101]; Figure~\ref{fig:grounding-effects}b). %
Both Claude models and gemini-3.1-pro beat the 0.249 analogue-only baseline, while open-weight models score 0.248 and 0.246 by copying the median (Figure~\ref{fig:allocation-analogues}a; Appendix~\ref{app:failure-modes}). %
Whether the two Claude models and gemini-3.1-pro use analogues selectively or mostly ignore them remains unresolved. %
Bare Qwen3-VL repeats persistence on 239 of 300 items. %

\paragraph{Movement-Matched Retrieval Rule~v2 and Repeat Runs.}
Under retrieval rule~v2, grounded error falls across all six models relative to rule~v1 (Table~\ref{tab:retrieval-v2}). %
The four proprietary models land within 0.015 of bare error, with no v2-minus-bare interval excluding zero (Figure~\ref{fig:grounding-effects}b); for gemini-3.1-pro the interval reaches $-0.026$ (Table~\ref{tab:retrieval-v2}). %
The analogue-only prediction scores 0.190 normalized error against 0.249 for rule~v1, but rule~v2 still exceeds persistence by 0.044 [0.019, 0.072]. %
A byte-identical bare repeat moved gpt-6-astra's error by 0.0001. %
Llama 4 Maverick moved by 0.008, changing 69 of 296 predictions, about a sixth of its grounding penalty. %
No allocation repeat is released for the other four core models, and single repeats supply weak variance estimates (Appendix~\ref{app:figlib-gap}). %
\subsection{Extended Analysis: Wildfire Smoke Detection}\label{subsec:app-figlib-detail}

The task evaluates wildfire smoke detection on 224 items, with 112 smoke frames and 112 clear frames (Table~\ref{tab:results-figlib}). Items come from 28 camera sequences across 17 fires, dated 2026-01-28 to 2026-09-09, and we do not claim they postdate any model's training cutoff (Appendix~\ref{subsec:contamination-and-dating}). %
Offsets span 300 to 2400 seconds from plume visibility, dropping frames within 180 seconds. %
Scoring clusters by fire on 196 paired items, excluding 28 reference frames (Appendix~\ref{app:figlib-unpaired}). %

\paragraph{Reference Frame Contrasts and Clustered Sensitivity.}
The reference frame turns 4 to 21 of the 112 smoke frames from missed to found across the six models, losing none (Table~\ref{tab:results-figlib}). Recall on smoke rises from 0.509--0.643 bare to 0.554--0.741 grounded, clearing zero for four models (Figure~\ref{fig:grounding-effects}a). %
Intervals clear zero for claude-opus-4.8 [0.014, 0.126], gemini-3.1-pro [0.130, 0.245], gpt-6-astra [0.045, 0.169], and Qwen3-VL [0.024, 0.126]. %
They span zero for claude-opus-5 [$-0.004$, 0.073] and Llama 4 Maverick [$-0.005$, 0.091]. %
Intervals expand percentile widths by 11.5 percent for 17 fire clusters (Appendix~\ref{app:metrics-detail}), establishing no nominal coverage. %
Clear false-positive rates rise on four models from 0.000--0.024 bare to 0.024--0.060 grounded, whereas claude-opus-4.8 and Qwen3-VL maintain zero in both conditions. %

\paragraph{Detection Timeline and Early Plume Visibility.}
Detection accuracy is lowest within ten minutes of plume appearance (0.25--0.50 bare, 0.36--0.57 grounded across 28 frames; Figure~\ref{fig:figlib-timeline}). %
The reference frame adds 13 correct decisions of 168 pooled pairs in that window (Figure~\ref{fig:figlib-timeline}b). %
In later windows, it adds 26 of 282 decisions from 10 to 25 minutes (lift of 0.092) and 17 of 222 decisions after 25 minutes (Table~\ref{tab:figlib-buckets}). %

\paragraph{Frame-Difference Detector Comparator.}\label{app:figlib-baseline}
A held-out frame-difference detector evaluated by leave-one-fire-out cross-validation achieves 0.770 accuracy and 0.920 recall with a 0.429 false-positive rate on clear frames. %
No core model accuracy interval against it excludes zero (margins $-0.061$ to $+0.066$). %
All twelve model arms beat the always-smoke constant on accuracy by 0.138 to 0.265, every interval excluding zero. %
Every model finds fewer smoke frames than the detector (recall margins $-0.179$ to $-0.411$) and has lower false-positive rates than the detector on clear frames. %
With gaps matched and capture times masked, the detector falls to 0.536 accuracy, below every model arm (Appendix~\ref{app:figlib-rerun}). %

\paragraph{Reference-to-Target Gap.}\label{app:figlib-timing}
Each of the 28 sequences contributes the frames nearest eight target offsets from the first visible plume: $-2400$, $-1800$, $-1200$, $-600$, 300, 900, 1500, and 2400 seconds. %
The earliest frame (39 to 40 minutes before plume) serves as reference. %
This reference precedes 84 clear targets by 9 to 30 minutes and 112 smoke targets by 44 to 80 minutes. %
Any threshold between 31 and 43 minutes separates all 196 items. %
Every frame also prints its capture date and time in the top strip, so a model could read the gap as well as infer it from scene change. %
No stored core-model response mentions elapsed time, a clock reading, or a date, though each holds only one sentence of reasoning. Eight no-smoke answers name the time of day (dawn, morning, or afternoon), each matching the printed clock. %

Table~\ref{tab:figlib-timing} reports paired change and its slope on the gap. %
Pooled recall gain across six models is 0.077, 0.065, 0.101, and 0.089 at 45, 55, 65, and 79 minutes, each widened interval above zero. %
Slope on gap is +0.005 per 10 minutes ($-0.029$ to $0.034$), with no model interval excluding zero; claude-opus-5 and Llama 4 Maverick gain only from 55 minutes. %
Bare misses drop from 102 of 168 pairs at 45 minutes to 57--72 later; rescue rates rise from 0.13 to 0.19--0.28 (slope +0.027 per 10 minutes [$-0.048$, 0.073]). %
At such low rates, the clear-frame slope cannot separate a scene-change cue from none. %
A cue that switches on between 30 and 44 minutes would leave the smoke gain flat across positions, so these slopes cannot exclude it. %

On clear frames, longer gaps increase scene change without smoke. %
The frame-difference detector's score, the mean absolute gray-level difference from the reference, rises from a median of 3.7 at 10 minutes to 5.9 at 30 minutes. %
Its clear false-positive rate climbs from 0.286 to 0.571 (+0.150 per 10 minutes [0.070, 0.247]); smoke recall climbs from 0.786 at 45 minutes to 1.000 at 79 minutes. %
Given its false-positive rise, part of its 0.920 recall may reflect elapsed time as well as plume growth. %
Model pooled grounded clear false-positive rates are 0.036, 0.012, and 0.036 at 10, 20, and 30 minutes (changes +0.030, 0.000, +0.024; slope $-0.003$ per 10 minutes [$-0.015$, 0.013]). %
Matched detector rates are 0.030, 0.012, and 0.042 (slope +0.008 [$-0.034$, 0.047]). %
The 11 model pairs becoming grounded false positives show median score 4.3 versus 5.0 for 488 staying negative. %
This clear-frame check covers gaps of 9 to 30 minutes, so it cannot exclude a cue that switches on between 30 and 44 minutes. %

Weighting frames equally, the reference raises balanced accuracy by 0.033 pooled [0.017, 0.046] and 0.004 to 0.082 per model (Table~\ref{tab:figlib-timing}b). %
Widened intervals clear zero for claude-opus-4.8, gemini-3.1-pro, and Qwen3-VL, and span zero for the others. %
Because the design ties label to gap, these checks neither confirm nor exclude elapsed time use. %
The rerun below removes that tie. %

\begin{table*}[tp]
\centering
\footnotesize
\setlength{\tabcolsep}{4pt}
\renewcommand{\arraystretch}{1.10}
\caption{\textbf{The reference frame gains smoke frames at every gap and its clear-frame false positives stay flat, but the gap alone separates the labels by design.} Smoke detection on FIgLib, six core models, one run per model and arm, on the 196 paired items. The reference is each sequence's frame about 40 minutes before the first visible plume. \emph{Gap} is the median minutes from the reference to the target at each ladder position, which holds 28 frames per model. (a)~Per model, clear columns count new/removed false positives and smoke columns count gained/lost smoke frames, grounded against bare. \emph{Pooled} rows give the six-model rate or change per position with its 95 percent interval. The mint rows give the held-out frame-difference detector's false-positive rate (clear) and recall (smoke). (b)~Least-squares slope, per 10 minutes of gap, of the paired change (grounded minus bare) over smoke or clear items. Balanced accuracy is the mean of recall on smoke frames and specificity on clear frames. The detector reads the same reference frame, so its balanced accuracy sits under \emph{Grounded}, and its slopes are those of its recall and false-positive rate. Intervals resample the 17 fires (20,000 resamples) and widen percentile widths by 11.5 percent as in Table~\ref{tab:cross-task-summary}; $\blacktriangle$ marks a per-model interval above zero. The 20-minute pooled interval is degenerate because its one new and one removed false positive fall on the same frame. No interval is adjusted for multiplicity. The mint rows' slopes are level slopes; \emph{Detector, matched} cuts the detector to each model's own grounded false-positive count and pools as for the models.}\label{tab:figlib-timing}
\begin{tabular}{@{}l rrr rrrr@{}}
\toprule
 & \multicolumn{3}{c}{Clear frames} & \multicolumn{4}{c}{Smoke frames} \\
\cmidrule(lr){2-4}\cmidrule(lr){5-8}
Gap (min) & 10 & 20 & 30 & 45 & 55 & 65 & 79 \\
\midrule
\grouprow{8}{(a) Changed verdicts by ladder position}
claude-opus-4.8 & 0/0 & 0/0 & 0/0 & 1/0 & 2/0 & 3/0 & 1/0 \\ %
claude-opus-5 & 1/0 & 1/0 & 1/1 & 0/0 & 1/0 & 2/0 & 1/0 \\ %
gemini-3.1-pro & 1/0 & 0/0 & 1/0 & 7/0 & 4/0 & 4/0 & 6/0 \\ %
gpt-6-astra & 1/0 & 0/1 & 2/0 & 3/0 & 3/0 & 3/0 & 2/0 \\ %
Qwen3-VL & 0/0 & 0/0 & 0/0 & 2/0 & 1/0 & 2/0 & 3/0 \\ %
Llama 4 Maverick & 2/0 & 0/0 & 1/0 & 0/0 & 0/0 & 3/0 & 2/0 \\ %
Pooled, bare & 0.006 & 0.012 & 0.012 & 0.393 & 0.661 & 0.643 & 0.571 \\ %
Pooled, grounded & 0.036 & 0.012 & 0.036 & 0.470 & 0.726 & 0.744 & 0.661 \\ %
Pooled change & +0.030 & 0.000 & +0.024 & +0.077 & +0.065 & +0.101 & +0.089 \\ %
\quad 95\% interval & {\scriptsize [0.009, 0.058]} & {\scriptsize [0.000, 0.000]} & {\scriptsize [0.002, 0.058]} & {\scriptsize [0.035, 0.130]} & {\scriptsize [0.026, 0.114]} & {\scriptsize [0.033, 0.193]} & {\scriptsize [0.008, 0.165]} \\ %
\comprow Frame difference & 0.286 & 0.429 & 0.571 & 0.786 & 0.929 & 0.964 & 1.000 \\ %
\bottomrule
\end{tabular}

\vspace{6pt}
\begin{tabular}{@{}l rr rrr@{}}
\toprule
 & \multicolumn{2}{c}{Slope per 10 min of gap} & \multicolumn{3}{c}{Balanced accuracy} \\
\cmidrule(lr){2-3}\cmidrule(lr){4-6}
 & Smoke gain & Clear FPR change & Bare & Grounded & Change \\
\midrule
\grouprow{6}{(b) Trend on the gap and balanced accuracy}
claude-opus-4.8 & +0.001 & 0.000 & 0.777 & 0.808 & +0.031$^{\blacktriangle}$ [0.007, 0.063] \\ %
claude-opus-5 & +0.010 & $-0.020$ & 0.798 & 0.804 & +0.006 [$-0.041$, 0.038] \\ %
gemini-3.1-pro & $-0.008$ & +0.001 & 0.772 & 0.854 & +0.082$^{\blacktriangle}$ [0.042, 0.118] \\ %
gpt-6-astra & $-0.012$ & +0.017 & 0.810 & 0.847 & +0.037 [$-0.008$, 0.081] \\ %
Qwen3-VL & +0.013 & 0.000 & 0.772 & 0.808 & +0.036$^{\blacktriangle}$ [0.012, 0.064] \\ %
Llama 4 Maverick & +0.025 & $-0.016$ & 0.743 & 0.747 & +0.004 [$-0.037$, 0.032] \\ %
Pooled, core six & +0.005 [$-0.029$, 0.034] & $-0.003$ [$-0.015$, 0.013] & 0.779 & 0.811 & +0.033 [0.017, 0.046] \\ %
\comprow Frame difference & +0.060 [0.016, 0.102] & +0.150 [0.070, 0.247] & -- & 0.746 & -- \\ %
\comprow Detector, matched & -- & +0.008 [$-0.034$, 0.047] & -- & -- & -- \\ %
\bottomrule
\end{tabular}
\end{table*}

\paragraph{Rerun With Matched Gaps and Masked Timestamps.}\label{app:figlib-rerun}
The rerun matches reference-to-target gaps across labels, masks printed times, and uses neutral reference wording (Table~\ref{tab:figlib-rerun}). %
Its 168 targets are the first design's frames at $-1800$, $-1200$, $-600$, 300, 900, and 1500 seconds. Each is paired with the frame nearest 600, 1200, or 1800 seconds earlier in the full FIgLib listing, at least 180 seconds before the plume. %
Median gaps agree across labels to within one second at every position, the widest within-sequence mismatch is 62 seconds, and the gap's AUROC for the smoke label is 0.574. %
Both arms black out the top 32 rows of every frame; across the 224 distinct frames shown, the printed caption ends by row 24. Clear-target references precede the first visible plume by about 40 minutes and smoke-target references by about five, so lighting or scene change could still reveal timing. %
The prompt calls the reference ``an earlier frame from the same camera'' in place of ``the same camera earlier in the day, before any plume''. %
Pooled over the six models, the reference raises recall by 0.063 [0.035, 0.102]. On the same frames the first design raised it by 0.081, and the difference, $-0.018$ [$-0.043$, 0.006], is unresolved. %
The gain is 0.060, 0.060, and 0.071 at gaps of 10, 20, and 30 minutes. Clear-frame false positives rise by 0.016 [$-0.005$, 0.052], five of them from claude-opus-5. %
In the bare arm, the masked and re-encoded rerun changes pooled recall by 0.018 [$-0.004$, 0.041] relative to the first run. No answer in either design quotes a clock reading or a date. %
On the rerun's pairs the frame-difference detector reaches 0.536 accuracy (recall 0.560, false-positive rate 0.488), against 0.726 on the first design's pairs for the same targets. %
Its recall and false-positive rate both rise with the gap, from 0.43 and 0.36 at 10 minutes to 0.75 and 0.61 at 30. It tracks scene change more than smoke. %
Every model arm beats it by 0.190 to 0.327 in accuracy, each widened interval above zero. %
The rerun covers the core six once per arm; the 29 sweep models and the 79-minute frames keep the first design. %

\begin{table*}[t]
\centering
\scriptsize
\setlength{\tabcolsep}{4pt}
\renewcommand{\arraystretch}{1.10}
\caption{\textbf{With matched gaps and masked timestamps, the reference still raises smoke recall for the four proprietary models.} Smoke detection rerun on 168 FIgLib targets from 28 sequences of 17 fires. The 84 clear frames lie 30, 20, and 10 minutes before the first visible plume, and the 84 smoke frames 5, 15, and 25 minutes after it. Each target's reference is the frame nearest 10, 20, or 30 minutes earlier, always before the plume, so each label meets each gap once per sequence. Both arms see every frame with its printed capture time masked, and the prompt calls the reference ``an earlier frame from the same camera''. One run per model and arm. \emph{Gained/lost}: smoke frames the reference turns from missed to found and back. \emph{FP $+$/$-$}: clear frames it turns into false positives and back. \emph{First design}: the recall change of Table~\ref{tab:figlib-timing}'s run on the same 84 smoke frames. Intervals resample the 17 fires (20,000 resamples) and widen percentile widths by 11.5 percent; $\blacktriangle$ marks a per-model recall interval above zero.}\label{tab:figlib-rerun}
\begin{tabular}{@{}l rr l r r l r@{}}
\toprule
 & \multicolumn{4}{c}{Recall on 84 smoke frames} & 84 clear & \multicolumn{2}{c}{Both labels} \\
\cmidrule(lr){2-5}\cmidrule(lr){6-6}\cmidrule(lr){7-8}
Model & Bare & Grounded & Change & Gained/lost & FP $+$/$-$ & Bal.\ acc.\ change & First design \\
\midrule
claude-opus-4.8 & 0.536 & 0.583 & +0.048$^{\blacktriangle}$ [0.004, 0.117] & 4/0 & 0/0 & +0.024 [0.002, 0.058] & +0.071 \\ %
claude-opus-5 & 0.607 & 0.667 & +0.060$^{\blacktriangle}$ [0.017, 0.112] & 5/0 & 5/0 & 0.000 [$-0.069$, 0.048] & +0.036 \\ %
gemini-3.1-pro & 0.595 & 0.726 & +0.131$^{\blacktriangle}$ [0.063, 0.220] & 11/0 & 1/0 & +0.060 [0.020, 0.104] & +0.179 \\ %
gpt-6-astra & 0.655 & 0.774 & +0.119$^{\blacktriangle}$ [0.046, 0.227] & 10/0 & 2/2 & +0.060 [0.000, 0.134] & +0.107 \\ %
Qwen3-VL & 0.571 & 0.607 & +0.036 [$-0.004$, 0.072] & 3/0 & 1/0 & +0.012 [$-0.001$, 0.024] & +0.060 \\ %
Llama 4 Maverick & 0.536 & 0.524 & $-0.012$ [$-0.047$, 0.003] & 1/2 & 1/0 & $-0.012$ [$-0.037$, 0.002] & +0.036 \\ %
Pooled, core six & 0.583 & 0.647 & +0.063 [0.035, 0.102] & 34/2 & 10/2 & +0.024 [0.000, 0.048] & +0.081 \\ %
\bottomrule
\end{tabular}
\end{table*}
\subsection{Extended Analysis: Aerial Question Answering}\label{subsec:app-wildfirevqa-detail}

Inputs show an RGB aerial frame and a thermal image; grounded prompts add the seven-number thermal block. %
Intervals are 95 percent frame-clustered bootstraps. %

\begin{table*}[t]
\centering
\footnotesize
\setlength{\tabcolsep}{4pt}
\renewcommand{\arraystretch}{1.05}
\caption{\textbf{A trivial hybrid rule outscores every model by point estimate, five of six sit within paired error of the held-out majority comparator, and Llama 4 Maverick sits below it; the thermal block's point estimates rise on the closed-form questions and fall on the remaining items.} Aerial question answering on WildFireVQA, 408 items over 390 FLAME~3 frame clusters, scored by exact match. Point estimates are followed by 95 percent frame-clustered bootstrap intervals. \emph{Gained / lost} counts item-level flips. \emph{Closed-form 48} and \emph{Other 360} report accuracy bare $\to$ grounded on the respective question subsets.}\label{tab:results-wildfirevqa}
\vspace{1pt}
\begin{tabular}{@{}l rrr c rr@{}}
\toprule
Model & Bare acc. & \begin{tabular}[b]{@{}r@{}}Grounded\\acc.\end{tabular} & Grd. $-$ Bare & \begin{tabular}[b]{@{}c@{}}Gained /\\lost\end{tabular} & Closed-form 48 & Other 360 \\
\midrule
\grouprow{7}{Reference points}
\comprow Per-question majority & 0.627 & 0.627 & & & 0.688 & 0.619 \\ %
\comprow Majority + closed-form & & 0.664 & & & 1.000 & 0.619 \\ %
\comprow Closed-form rule & & & & & 1.000 & \\ %
\midrule
\grouprow{7}{Proprietary models}
claude-opus-4.8 & 0.642 & 0.642 & +0.000 & 17 / 17 & 0.667 $\to$ 0.896 & 0.639 $\to$ 0.608 \\ %
 & {\scriptsize [0.597, 0.687]} & {\scriptsize [0.596, 0.688]} & {\scriptsize [$-$0.029, +0.028]} & & & \\ %
claude-opus-5 & 0.650 & 0.657 & +0.007 & 18 / 15 & 0.646 $\to$ 0.917 & 0.650 $\to$ 0.622 \\ %
 & {\scriptsize [0.603, 0.695]} & {\scriptsize [0.612, 0.701]} & {\scriptsize [$-$0.020, +0.035]} & & & \\ %
gemini-3.1-pro & 0.632 & 0.647 & +0.015 & 22 / 16 & 0.667 $\to$ 0.938 & 0.628 $\to$ 0.608 \\ %
 & {\scriptsize [0.587, 0.678]} & {\scriptsize [0.601, 0.692]} & {\scriptsize [$-$0.015, +0.045]} & & & \\ %
gpt-6-astra & 0.623 & 0.635 & +0.012 & 15 / 10 & 0.792 $\to$ 0.938 & 0.600 $\to$ 0.594 \\ %
 & {\scriptsize [0.575, 0.670]} & {\scriptsize [0.589, 0.681]} & {\scriptsize [$-$0.012, +0.036]} & & & \\ %
\midrule
\grouprow{7}{Open-weight models}
Qwen3-VL & 0.618 & 0.618 & +0.000 & 20 / 20 & 0.729 $\to$ 0.771 & 0.603 $\to$ 0.597 \\ %
 & {\scriptsize [0.570, 0.665]} & {\scriptsize [0.570, 0.665]} & {\scriptsize [$-$0.031, +0.030]} & & & \\ %
Llama 4 Maverick & 0.534 & 0.544 & +0.010 & 26 / 22 & 0.688 $\to$ 0.833 & 0.514 $\to$ 0.506 \\ %
 & {\scriptsize [0.487, 0.582]} & {\scriptsize [0.495, 0.593]} & {\scriptsize [$-$0.024, +0.043]} & & & \\ %
\bottomrule
\end{tabular}
\end{table*}

\paragraph{Majority Reference and Hybrid Baseline Evaluation.}
The 0.627 majority rule is fitted on the 25,027-record pool that contains the evaluation items. Refitting it on the 24,619 records left once the 408 evaluation items are removed leaves the majority class unchanged for all 34 question types. The comparator therefore scores the same 0.627 held out as in place. %
No core model's paired 95 percent interval against the majority lies wholly above zero. Five include zero in both arms, with point differences from $-0.010$ to $+0.022$ bare and $-0.010$ to $+0.029$ grounded. %
Only Llama 4 Maverick excludes zero, trailing the majority bare ($-0.093$, [$-0.158$, $-0.027$]) and grounded ($-0.083$, [$-0.148$, $-0.017$]). %
For Llama 4 Maverick, resampling question types gives [$-0.189$, $+0.000$] bare, and two-way resampling gives [$-0.225$, $+0.037$] (Appendix~\ref{app:wildfirevqa-detail}). %
The 0.664 hybrid rule exceeds all model point estimates (highest: claude-opus-5 grounded at 0.657; Table~\ref{tab:results-wildfirevqa}). %
Against that hybrid too, only Llama 4 Maverick resolves below, bare ($-0.130$, [$-0.195$, $-0.063$]) and grounded ($-0.120$, [$-0.183$, $-0.056$]). This one holds under either resampling unit taken alone, at [$-0.228$, $-0.034$] bare over question types, and closes only when both are resampled together. %

\paragraph{Accuracy by Label Source and by Applicability.}
On the 132 items labeled by formula, telemetry, or detector, every core-model arm scores below the 0.614 majority, from 0.470 to 0.523 bare and from 0.515 to 0.606 grounded. %
Without the closed-form items, 84 items over 7 types remain, and every arm trails the 0.571 majority there by 0.143 to 0.238. %
On DS1, DS3, PD1, and PD7, 14 references read no hotspot or no fire although the frame maximum lies between 272 and 567\,$^\circ$C. %
The formula requires connected regions above 200\,$^\circ$C with equivalent radius at least 0.75\,m; the prompt defines hotspots by local contrast. %
In the pool, 241 of 538 frames above 200\,$^\circ$C carry this reference across these four types. %
All 16 models miss all 14 items, which the majority comparator scores. %
Removing the 14 after the fact raises every contrast against the majority, and claude-opus-5 grounded then resolves above it on the 394 items left, $+0.066$ [$+0.005$, $+0.128$]. Because we chose this exclusion after seeing the failures, we read it only as a pointer for a label audit of these types. %
On the 276 model-verified items, 10 of 12 arms exceed the 0.634 majority. %
Bare claude-opus-4.8 and claude-opus-5 unadjusted intervals lie above it ($+0.076$ [$+0.007$, $+0.145$] and $+0.101$ [$+0.032$, $+0.173$]). %
Because source and question type are collinear, this split cannot bound label error. %
Resampling question types leaves one separation (claude-opus-5 bare on model-verified items, [$+0.014$, $+0.188$]); two-way resampling leaves none. %
Split by applicability, no core model resolves above majority in either stratum (Table~\ref{tab:aerial-applicability}); Llama 4 Maverick resolves below bare at 1.0 and grounded below 1.0. %
Grounded minus bare is positive for all six at 1.0 (36 closed-form items) and negative for five below 1.0. %
Holm adjustment over twelve arms leaves no comparator $p<0.05$ (smallest 0.073). %

\paragraph{Exact Paired Tests by Item Group.}\label{app:aerial-verification}
These exact McNemar tests condition on the stored items and do not assess variation across question types or repeated model runs. %
Table~\ref{tab:aerial-exact} shows gains on the 48 closed-form items for all six core models (eight of sixteen at unadjusted $p<0.05$, with no decline at that threshold); the proprietary four lost none. %
On the 347 remaining items, accuracy falls for all six core models, reaching $p<0.05$ only for the two Claude models. %
Neither Claude decline (unadjusted $p$ of 0.019 and 0.035) survives a Holm correction. %
Pooled arm contrasts contain zero for all six core models. %

\begin{table*}[t]
\centering
\footnotesize
\setlength{\tabcolsep}{6pt}
\renewcommand{\arraystretch}{1.05}
\caption{\textbf{On the 48 closed-form aerial items the thermal summary block raises accuracy for 14 of the sixteen full-capability models, eight at unadjusted item-level $p<0.05$, with no decline at that threshold; on the 347 remaining items, accuracy falls for all six core models and only the two Claude models meet that threshold.} Exact two-sided McNemar paired tests on the prespecified aerial item partition. \emph{Gained / lost} counts item-level flips from bare to grounded. Values of $p < 0.05$ indicate unadjusted two-sided significance. The 347-item columns report the core six only, so the sweep rows leave them blank. On the 13 no-fire-branch items, whose labels a maximum below 200\,$^\circ$C fixes, no core model shows a resolved change. A PD1 rule of the same form would move 3 more items out of the 347 without changing any gained, lost, or $p$ entry. Llama 4 Scout moves from 33 to 31 of the 48 and Ministral 3 8B from 34 to 30, with no recorded error in either arm.}\label{tab:aerial-exact}
\vspace{1pt}
\begin{tabular}{@{}l cc ccc@{}}
\toprule
& \multicolumn{2}{c}{48 closed-form items} & \multicolumn{3}{c}{347 remaining items} \\
\cmidrule(lr){2-3} \cmidrule(lr){4-6}
Model & Gained / lost & $p$ & Bare $\to$ grounded & Gained / lost & $p$ \\
\midrule
\grouprow{6}{Core set}
claude-opus-4.8 & 11 / 0 & 0.001 & 0.640 $\to$ 0.608 & 6 / 17 & 0.035 \\ %
claude-opus-5 & 13 / 0 & $<$ 0.001 & 0.654 $\to$ 0.622 & 4 / 15 & 0.019 \\ %
gemini-3.1-pro & 13 / 0 & $<$ 0.001 & 0.628 $\to$ 0.608 & 9 / 16 & 0.230 \\ %
gpt-6-astra & 7 / 0 & 0.016 & 0.617 $\to$ 0.608 & 7 / 10 & 0.629 \\ %
Qwen3-VL & 8 / 6 & 0.791 & 0.617 $\to$ 0.611 & 12 / 14 & 0.845 \\ %
Llama 4 Maverick & 10 / 3 & 0.092 & 0.530 $\to$ 0.519 & 14 / 18 & 0.597 \\ %
\midrule
\grouprow{6}{Sweep}
Nova Lite & 15 / 3 & 0.008 & & & \\ %
Nova Pro & 15 / 7 & 0.134 & & & \\ %
Nova 2 Lite & 7 / 3 & 0.344 & & & \\ %
Llama 4 Scout & 8 / 10 & 0.815 & & & \\ %
Mistral Large 3 & 17 / 8 & 0.108 & & & \\ %
Ministral 3 8B & 9 / 13 & 0.523 & & & \\ %
Kimi K2.5 & 17 / 3 & 0.003 & & & \\ %
Gemma 3 27B & 18 / 0 & $<$ 0.001 & & & \\ %
Gemma 3 12B & 15 / 0 & $<$ 0.001 & & & \\ %
Gemma 3 4B & 6 / 1 & 0.125 & & & \\ %
\bottomrule
\end{tabular}
\end{table*}

\begin{table*}[t]
\centering
\footnotesize
\setlength{\tabcolsep}{4pt}
\renewcommand{\arraystretch}{1.05}
\caption{\textbf{On the 132 items labeled by formula, telemetry, or detector, every arm scores below the held-out majority; on the 276 model-verified items, two bare arms have unadjusted intervals above it.} %
Accuracy on WildFireVQA per core model and arm, pooled and within five strata, beside the held-out per-question majority scored on the same items. %
Column heads give items and, in parentheses, question types. %
Each question type has one label source, so the label-source columns compare disjoint sets of question types and cannot separate label quality from question type. %
The 84-item column removes the 48 closed-form items, all of which carry formula labels. %
Markers flag an unadjusted paired 95 percent frame-clustered interval wholly above ($^{\blacktriangle}$) or below ($^{\blacktriangledown}$) zero: model minus majority, and grounded minus bare in the last block. %
Because differences move in whole items, five unmarked cells have a frame-clustered interval ending at exactly zero and could gain a marker under another seed. %
With question types resampled, only claude-opus-5 bare on model-verified items keeps a comparator marker, and five arm-contrast markers remain. %
Resampling both dimensions leaves none. %
Unadjusted exact McNemar tests reach $p<0.05$ on every comparator marker and on six of the ten arm-contrast markers. %
Holm adjustment over the twelve arms of each column leaves no comparator $p$-value below 0.05. %
Intervals use 20,000 resamples and base seed 20260915.}\label{tab:aerial-applicability}\label{tab:aerial-strata} %
\vspace{1pt}
\begin{tabular}{@{}ll cccccc@{}}
\toprule
& & & \multicolumn{3}{c}{Label source} & \multicolumn{2}{c}{Applicability} \\
\cmidrule(lr){4-6} \cmidrule(lr){7-8}
& & All & \begin{tabular}[b]{@{}c@{}}Formula, telemetry,\\or detector\end{tabular} & \begin{tabular}[b]{@{}c@{}}Same, without\\closed form\end{tabular} & \begin{tabular}[b]{@{}c@{}}Model-\\verified\end{tabular} & $= 1.0$ & $< 1.0$ \\
Model & Arm & 408 (34) & 132 (11) & 84 (7) & 276 (23) & 284 (33) & 124 (25) \\ %
\midrule
\comprow \multicolumn{2}{@{}l}{Held-out majority} & 0.627 & 0.614 & 0.571 & 0.634 & 0.637 & 0.605 \\ %
\midrule
\grouprow{8}{Proprietary models}
claude-opus-4.8 & bare & 0.642 & 0.500 & 0.405 & 0.710$^{\blacktriangle}$ & 0.651 & 0.621 \\ %
claude-opus-4.8 & grounded & 0.642 & 0.583 & 0.405 & 0.670 & 0.676 & 0.565 \\ %
claude-opus-5 & bare & 0.650 & 0.470$^{\blacktriangledown}$ & 0.369$^{\blacktriangledown}$ & 0.736$^{\blacktriangle}$ & 0.665 & 0.613 \\ %
claude-opus-5 & grounded & 0.657 & 0.576 & 0.381$^{\blacktriangledown}$ & 0.696 & 0.701 & 0.556 \\ %
gemini-3.1-pro & bare & 0.632 & 0.515 & 0.429 & 0.688 & 0.683 & 0.516 \\ %
gemini-3.1-pro & grounded & 0.647 & 0.606 & 0.417 & 0.667 & 0.701 & 0.524 \\ %
gpt-6-astra & bare & 0.623 & 0.523 & 0.369$^{\blacktriangledown}$ & 0.670 & 0.655 & 0.548 \\ %
gpt-6-astra & grounded & 0.635 & 0.576 & 0.369$^{\blacktriangledown}$ & 0.663 & 0.676 & 0.540 \\ %
\grouprow{8}{Open-weight models}
Qwen3-VL & bare & 0.618 & 0.492 & 0.357$^{\blacktriangledown}$ & 0.678 & 0.634 & 0.581 \\ %
Qwen3-VL & grounded & 0.618 & 0.523 & 0.381$^{\blacktriangledown}$ & 0.663 & 0.651 & 0.540 \\ %
Llama 4 Maverick & bare & 0.534$^{\blacktriangledown}$ & 0.470$^{\blacktriangledown}$ & 0.345$^{\blacktriangledown}$ & 0.565 & 0.549$^{\blacktriangledown}$ & 0.500 \\ %
Llama 4 Maverick & grounded & 0.544$^{\blacktriangledown}$ & 0.515 & 0.333$^{\blacktriangledown}$ & 0.558 & 0.592 & 0.435$^{\blacktriangledown}$ \\ %
\midrule
\grouprow{8}{Grounded $-$ bare}
claude-opus-4.8 & & $+0.000$ & $+0.083^{\blacktriangle}$ & $+0.000$ & $-0.040^{\blacktriangledown}$ & $+0.025$ & $-0.056^{\blacktriangledown}$ \\ %
claude-opus-5 & & $+0.007$ & $+0.106^{\blacktriangle}$ & $+0.012$ & $-0.040^{\blacktriangledown}$ & $+0.035^{\blacktriangle}$ & $-0.056^{\blacktriangledown}$ \\ %
gemini-3.1-pro & & $+0.015$ & $+0.091^{\blacktriangle}$ & $-0.012$ & $-0.022$ & $+0.018$ & $+0.008$ \\ %
gpt-6-astra & & $+0.012$ & $+0.053^{\blacktriangle}$ & $+0.000$ & $-0.007$ & $+0.021$ & $-0.008$ \\ %
Qwen3-VL & & $+0.000$ & $+0.030$ & $+0.024$ & $-0.014$ & $+0.018$ & $-0.040$ \\ %
Llama 4 Maverick & & $+0.010$ & $+0.045$ & $-0.012$ & $-0.007$ & $+0.042$ & $-0.065^{\blacktriangledown}$ \\ %
\bottomrule
\end{tabular}
\end{table*}

\paragraph{Opposite Subgroup Changes Under the Thermal Block.}
Gains of 15 to 26 items offset losses of 10 to 22, with all intervals containing zero. %
On the 48 closed-form items, accuracy rose from 0.646--0.792 bare to 0.771--0.938 grounded, but fell by 0.006 to 0.031 on the remaining 360 items (Figure~\ref{fig:wildfirevqa}b). %
On 67 fire frames those closed-form labels rest on a clipped maximum (Appendix~\ref{app:defect-ceiling}). %
Gains concentrate in Cross-Modal Reasoning and Distribution and Segmentation, which contain three closed-form types (CMR4, DS7, DS8; Table~\ref{tab:wildfirevqa-category}). %

\paragraph{What the 34 Closed-Form Misses Are.}\label{app:closed-form-misses}
Grounded, the core six miss 34 of 288 answers on the 48 closed-form items (proprietary: 5, 4, 3, 3; Qwen3-VL: 11; Llama 4 Maverick: 8). %
Fifteen fall on frames clipped at 187.424\,$^\circ$C (Appendix~\ref{app:defect-ceiling}). %
There, eight answer Active fire where CL1 reads Smoldering, six choose 100--200 where CMR4 reads no fire detected, and one gives below 5 percent where DS8 reads None. %
Nine answer Smoldering on CL1 (``dominant fire behavior'') where maximum exceeds 200\,$^\circ$C but only 0.075 to 1.153 percent of pixels do, fitting an area reading of dominance. %
Seven match a fraction reading of a percentage field, one matches the other field bin, and one answers None for 0.0 rounding a single pixel above 400\,$^\circ$C. %
The last is Llama 4 Maverick answering Smoldering on a no-fire frame (17.1\,$^\circ$C maximum). %
Of 34 misses, 21 repeat bare answers, 4 replace other errors, and 9 replace correct answers (all open weight). %
The proprietary four lost no item to the block; each of their 15 misses sits on a clipped frame or fits the area reading of CL1. %
Across ten sweep models, these classes cover 96 of 190 misses. %
On 13 no-fire-branch items, the core six miss 46 of 78 grounded answers, including all 24 on four clipped-frame items. %

\subsection{Extended Analysis: Fire Danger Forecasting}\label{subsec:app-mesogeos-detail}

Next-day fire danger forecasting evaluates 386 items from 4,120 reconstructed 2021--2022 holdout cells (Table~\ref{tab:results-mesogeos}; Mesogeos reports 4,107 cells; Appendix~\ref{app:defect-mesogeos}). %
Subsampling sets a 0.339 sample positive rate that reflects dataset selection (Appendix~\ref{app:defects}). %
Models predict next-day fire ($\geq 30$\,ha) in a 1~km cell from 12 non-leaking daily drivers, 12 static fields, and the month. %
Grounded runs add monthly climatology from 2006--2019 training years. %
Evaluation clusters by 1-degree cell and month (352 clusters; Figure~\ref{fig:grounding-effects}c). %

\paragraph{Climatology Contrast and Trained Classifiers.}
Monthly climatology lowers ranking on all four proprietary models by 0.005 to 0.032 AUPRC, with no block-clustered interval excluding zero (Table~\ref{tab:results-mesogeos}, Figure~\ref{fig:grounding-effects}c). %
Only gemini-3.1-pro and claude-opus-5 beat the 0.654 temperature rule (0.682--0.697). %
Qwen3-VL gains 0.051 [0.006, 0.092], the one interval excluding zero on this task, while Llama 4 Maverick moves by $-0.014$ [$-0.063$, 0.037]. %
A boosted classifier on 121 prompt numbers scores 0.851 (0.154 [0.084, 0.227] above the best prompted run; Appendix~\ref{app:trained-baselines}). %
Logistic regression scores 0.768, against published holdout baselines of 0.853--0.858 \citep{kondylatos2023mesogeos}. %

\paragraph{Decision Divergence, Missing Value Imputation, and Calibration.}
claude-opus-5 ranks second at 0.688 AUPRC but reaches fire-class F1 of 0.269. %
It calls fire on 0.065 of items with missing answers filled by thresholding stated probabilities at one half (sample positive rate 0.339; 46 bare, 90 grounded; Table~\ref{tab:results-mesogeos}). %
Grounded omission-label association is resolved ($p = 0.0002$, two-sided Fisher exact test). %
Bare omissions show no resolved association ($p = 0.7403$; Appendix~\ref{app:answers-hazards}). %
gpt-6-astra calls fire on 0.044 of items (F1 0.189 at 0.620 AUPRC). %
Proprietary models state probabilities compressed toward zero on this benchmark sample (mean 0.06 to 0.22). %
Expected calibration errors across twelve core runs span 0.119 to 0.326 against 0.037 for the calendar prior (Figure~\ref{fig:calibration}). %

\paragraph{Prompt Paraphrase Sensitivity and Repeat Consistency.}
In single bare runs, two prompt paraphrases move the two leading models in opposite directions (Figure~\ref{fig:prompt-sensitivity}). Under the first paraphrase, claude-opus-5 gains 0.043 [0.009, 0.080] AUPRC (0.688 to 0.731), while gemini-3.1-pro moves by $-0.049$ [$-0.093$, $-0.009$] (0.697 to 0.648), reversing their order. %
Temperature-zero repeats moved claude-opus-5 by 0.019 [$-0.001$, 0.039] and gemini-3.1-pro by 0.009 [$-0.021$, 0.039] (Appendix~\ref{app:prompt-sensitivity}). %
The leaders' gap of about 0.01 under the original prompt is comparable to those repeat changes. %
Under the second paraphrase, gpt-6-astra rises from 0.620 to 0.711 (gain 0.091 [0.033, 0.147]; Table~\ref{tab:prompt-sensitivity}), while Llama 4 Maverick drops from 0.528 to 0.393 ($-0.136$ [$-0.200$, $-0.069$]). %
The 0.851 boosted classifier stays above the best prompted run (0.731). %

\subsection{Extended Analysis: Fire Data Tool Use}\label{subsec:app-tooluse-detail}

The task evaluates 156 questions over 2,303,566 FPA-FOD records across twelve families: 13 single lookup, 78 filtered aggregate, and 65 multi-step (Table~\ref{tab:results-tooluse}). %
Tolerances are exact for counts, years, and names, 0.5 percent for acres, and 0.1 point for percentages. %
The bare arm answers from parameters; the tool arm may execute eight queries, with \texttt{unknown} scored wrong. %
gpt-6-astra used the Responses API because its chat endpoint rejects tools with reasoning; omitted reasoning tokens leave arm comparability unverified (Appendix~\ref{app:tooluse-detail}). %
Intervals are paired and clustered across the 12 families, a coarse unit that widens them (Table~\ref{tab:results-tooluse}, Figure~\ref{fig:grounding-effects}d). %

\paragraph{SQL Tool Performance and Database Access.}
Bare models answer 0 to 25 questions correctly (0.000 to 0.160; Table~\ref{tab:results-tooluse}), four trailing the 0.141 family-constant baseline. %
The SQL tool lifts proprietary models to 0.994--1.000 and open-weight models to 0.885 and 0.891 (Figure~\ref{fig:grounding-effects}d). %
Paired gains reach 0.808 to 0.994, with all intervals excluding zero. %
Mean tool calls range from 1.00 (Llama 4 Maverick) to 1.62 (gemini-3.1-pro) against the eight-call cap. %

\paragraph{Failure Taxonomy of Open-Weight Models.}
All 35 open-weight tool errors take two forms, with none from proprietary models (Table~\ref{tab:results-tooluse}). %
Twenty-three errors (10 Qwen3-VL, 13 Llama 4 Maverick) filtered \texttt{DISCOVERY\_DATE} against guessed formats (\texttt{06/01/2001}, \texttt{1994-06-01}) rather than stored \texttt{6/1/2001} or \texttt{DISCOVERY\_DOY}. %
Qwen3-VL made seven further date errors in windowed sizes (six abstentions); Llama 4 Maverick returned Python snippets on five percentage items (Appendix~\ref{app:tooluse-detail}). %
Stating date format in the schema is an untested candidate repair (Appendix~\ref{app:tooluse-detail}). %

\paragraph{Surface Paraphrases and Schema Literal Removal.}
Twelve templates generate the 156 questions, so a perfect score might reflect template familiarity rather than language understanding. %
The first rewrites phrasing while keeping \texttt{COLUMN = \textquotesingle value\textquotesingle} predicates. %
The second replaces schema literals with natural language, such as \emph{Alaska} for \texttt{STATE = \textquotesingle AK\textquotesingle}. %
The schema still reaches the model in the system prompt. %
Answers, tolerances, and reference queries stay byte-identical across all three sets, so the scorer is unchanged and every contrast is paired. %
Neither rewrite moves proprietary models item for item: three hold 1.000 and gpt-6-astra holds 0.994. %
In \texttt{count\_county\_year}, the second rewrite is ambiguous because the table carries both \texttt{COUNTY} and \texttt{FIPS\_NAME}. %
On 143 clean items across eleven families, removing schema literals costs Qwen3-VL 0.175 ([$-0.266$, $-0.091$]) and Llama 4 Maverick 0.126 ([$-0.284$, $+0.025$]). %
Only Qwen3-VL excludes zero under sensitivity expansion. %
Over all 156 items, drops are 0.199 ([$-0.296$, $-0.108$]) and 0.154 ([$-0.310$, $-0.017$]), both excluding zero. %
Twenty of 91 errors write full state names into two-letter code columns. %
Both open-weight models gained 0.032 on the first rewrite in date-window families, inseparable from run variation without repeats. %

\subsection{Extended Analysis: Model Sweep}\label{subsec:app-sweep-detail}

The sweep adds 29 models served through Amazon Bedrock and selected by a capability probe (Appendix~\ref{app:model-sweep}). Ten accept images and tools (three via the Gemma adapter) and ran all five tasks, joining the core six as sixteen full-capability models (Table~\ref{tab:model-sweep}). %
The other nineteen reject images; all ran allocation and fire danger, and seventeen ran tool use (Table~\ref{tab:text-sweep}). %
Cluster intervals cover full-capability models on four tasks (Figure~\ref{fig:grounding-sweep}); paraphrases, repeats, and calibration remain on the core six. %

\paragraph{Aerial Question Answering Across the Sixteen Full-Capability Models.}
On aerial questions, no added model reaches the 0.627 majority comparator (bare accuracy 0.392 to 0.588, grounded 0.404 to 0.613). %
The thermal block raises accuracy on 11 of sixteen models (changes $-0.022$ to $+0.051$). %
Frame-clustered intervals exclude zero on Mistral Large 3 (0.044 [0.007, 0.083]) and Gemma 3 27B (0.051 [0.020, 0.084]), both below majority. %

\paragraph{Results for Text-Only Models.}
Clustered intervals over the nineteen sharpen comparator findings: on parsed items, no bare-minus-persistence interval lies below zero (four contain zero, fifteen above), and no fire-danger interval beats the temperature rule. %
GLM 5 scores 0.570 [0.481, 0.659] on 386 items against 0.654. %
MiniMax M2.5 gives 0.583 [0.477, 0.686] on the 246 items parsed in both arms, against 0.665 on those same items. %
Bare allocation error spans 0.146 to 0.927. Llama 3.3 70B holds the high end after predicting 412 personnel for an incident staffed by two, and scores 0.244 without that item. %
Tool gains exclude zero upward on 15 of seventeen models (Figure~\ref{fig:text-sweep}). %
Neither exception loses ground beyond noise: Llama 3.1 70B $-0.006$ [$-0.019$, 0.000] and Mistral Small 2402 $-0.019$ [$-0.058$, 0.013]. %
At 0.000, Llama 3.1 70B never called the tool and answered from memory. Llama 3.3 70B wrote its call as JSON text in the message on 108 of 156 items (0.288). Mistral Small 2402 wrote the SQL in prose beside a placeholder answer (0.038). %

\section{Scoring Rules and Comparators, by Task}\label{app:scoring-detail}
This appendix details each task's scoring rule, non-LLM comparator, and interval estimation; normalized allocation error, Equation~(\ref{eq:alloc}), and paired contrast, Equation~(\ref{eq:contrast}), follow Section~\ref{subsec:metrics-and-scoring}. %

\begin{itemize}
  \item \textbf{Wildfire Smoke Detection.} \emph{Metric}: recall on smoke, false-positive rate on clear frames, and accuracy by offset bucket. \emph{Rule}: buckets partition frame offset in minutes from first visible plume at 0--10, 10--25, and 25+ min. \emph{Comparator}: two constants and a reference-frame difference detector. %
  \item \textbf{Temperature-Grounded Aerial Question Answering.} \emph{Metric}: accuracy against released answers. \emph{Rule}: unparsed answers score wrong; 48 thermal items reported separately. \emph{Comparator}: held-out majority (0.627), closed-form thermal rule, and 0.664 hybrid. %
  \item \textbf{Fire Danger Forecasting.} \emph{Metric}: AUPRC and fire-class F1. \emph{Rule}: missing calls filled as $c_i = \mathbf{1}[p_i \ge 0.5]$. \emph{Comparator}: calendar-month prior and temperature rule. %
  \item \textbf{Daily Personnel Allocation.} \emph{Metric}: normalized error, Equation~(\ref{eq:alloc}). \emph{Rule}: normalizer $s_{f(i)}$ is mean filed staffing of the fire. \emph{Comparator}: persistence and trained regressor. %
  \item \textbf{Fire Data Tool Use.} \emph{Metric}: accuracy against reference programs. \emph{Rule}: exact for counts, years, and names; a relative tolerance of 0.5 percent with a one-acre floor for acres; 0.1 point for percentages. \emph{Comparator}: best constant per family. %
\end{itemize}

\paragraph{Protocol, Every Task.}
Paired contrasts evaluate items completed by both arms. %
Intervals are 95 percent percentile bootstraps paired across arms with $B = 20{,}000$ resamples, at seed 20260915: $\mathrm{CI}_{95}(\Delta_m) = [\Delta_m^{(0.025)},\; \Delta_m^{(0.975)}]$. %
Clustering units are 17 fires, 390 frames (holding 34 types fixed), 245 incidents, 352 spatial blocks by month, and 12 question families. %
Appendix~\ref{app:metrics-detail} gives a small-cluster sensitivity expansion of these intervals and states what it does and does not establish. %

\section{Reproducibility Details}\label{app:repro}

\subsection{Example Prompts}\label{app:example-prompts}
Each block abridges one prompt pair to its structure: [...] marks a cut, bracketed labels are ours, and some lines are shortened. The released item records and run scripts reconstruct each full prompt. %

\paragraph{Personnel Allocation (ICS-209-PLUS).}
{\footnotesize
\begin{verbatim}
Item: alloc-2015_2776766_PARKER_RIDGE
[System] You forecast wildfire filings.
Answer: {"personnel": <int>,
 "reasoning": "<one sentence>"}.
[User (bare)] ICS-209 report, 2015-08-19:
- acres: 6147; new: 0; cont: 0; today: 78
- past personnel: [96, 96, 98, 78]
[...] (aerial, costs, fuel, structures)
- weather: ...red flag warning [...]
[Grounded arm adds before weather:]
Historical analogues in same band (v1):
- 130 -> 130 (ratio 1.0; ac 10000, cont 0)
[...] (6 historical analogue incidents)
- median ratio in this band: 1.38
[Question] How many personnel will this
incident report assigned on 2015-08-20?
\end{verbatim}
}

\paragraph{Smoke Detection (FIgLib).}
{\footnotesize
\begin{verbatim}
Item: figlib-20260128_Fire_wilson-s-...
[System] You judge camera images. Answer:
{"smoke": bool, "reasoning": "<str>"}.
[User (bare)]
[image: camera view, offset -1801s]
Does this frame show wildfire smoke?
Haze, fog, cloud, and dust are not smoke.
Answer for this frame alone.
[Grounded arm prepends reference frame:]
Reference frame: the same camera earlier
in the day, before any plume.
[Rerun with matched gaps:
Reference frame: an earlier frame
from the same camera.]
[image: reference frame, offset -2341s]
Frame to judge:
[image: camera view, offset -1801s]
Does this frame show wildfire smoke?
Haze, fog, cloud, and dust are not smoke.
Answer for this frame alone.
\end{verbatim}
}

\paragraph{Aerial QA (WildFireVQA).}
{\footnotesize
\begin{verbatim}
Item: wfvqa-Sycan2A-Fire-00001-CL1
[System (bare)] Expert wildfire visual QA.
Given: RGB UAV and thermal image.
Output JSON: {"answer": "<verbatim>",
 "applicability_score": <float 0-1>}
[Grounded system adds temperature info:]
- temperature summary of radiometric TIFF
- infer temps via min/max scale [...]
[User (bare)]
RGB image: [image: UAV RGB view]
Thermal: [image: inferno TIFF render]
Question: What is dominant fire behavior?
Choices: Active fire | Smoldering | Ext
Reply: {"answer": <verbatim>,
 "applicability_score": <float 0-1>}
[Grounded arm adds before question:]
Temperature summary of radiometric TIFF:
{"min": -22.8, "max": 608.7, "mean": 16,
 "std": 31.3, "top3_mean": 119.5, pct...}
\end{verbatim}
}

\paragraph{Fire Danger (Mesogeos).}
{\footnotesize
\begin{verbatim}
Item: mesogeos-neg-14596-2021-01-05
[System] You judge wildfire danger.
Output: {"probability": <0-1>,
 "fire": bool, "reasoning": "<str>"}.
[User (bare)] Cell at lon -7.50, lat 37.32.
Daily drivers, 30 days to 2021-01-04:
- t2m: last 6 days 287.2, ... | mean 288.3
[...] (12 drivers: d2m, tp, wind, rh, ...)
Static: dem 91.1, slope 1.55, [...]
Units: t2m is K; tp is m; wind is m/s
[Grounded arm adds before question:]
Climatology for month (mean, sd):
- t2m: mean 282.4, sd 5.1; d2m: 276.6 [...]
[Question] Does wildfire >= 30 ha start
in this cell on 2021-01-05?
\end{verbatim}
}

\paragraph{Fire Data Tool Use (FPA-FOD).}
{\footnotesize
\begin{verbatim}
Item: tooluse-count_state_year-01
[System] Assistant for US FPA-FOD DB.
Schema: table 'Fires' (39 columns:
 FOD_ID, STATE, FIRE_YEAR, SIZE, ...).
Contract: Last line: ANSWER: <value>.
[User (bare)]
In FPA-FOD database, how many records
have STATE = 'AK' and FIRE_YEAR = 1992?
Format answer as: a single integer.
[Grounded arm adds to user message:]
SQL tool 'query_fpafod' is available.
You must query database for answer...
[Declared tool: query_fpafod]
SELECT against FPA-FOD SQLite (50 rows)
Params: {"type": "object", "properties":
 {"sql": {"type": "string"}},
 "required": ["sql"]}
\end{verbatim}
}

\begin{table*}[tp]
\centering
\footnotesize
\setlength{\tabcolsep}{6pt}
\caption{\textbf{Only Gemini and GPT models report nonzero reasoning tokens.} Median reasoning tokens per response (bare/grounded; bare arm for tool use). n/r: not reported on every response (22 of 224 gemini-3.1-pro bare smoke records carry the field).}\label{tab:reasoning-tokens}
\begin{tabular}{@{}l ccccc@{}}
\toprule
Model & Alloc. & Smoke & Aerial & Danger & Tool \\
\midrule
claude-opus-4.8 & 0/0 & 0/0 & 0/0 & 0/0 & 0 \\
claude-opus-5 & 0/0 & 0/0 & 0/0 & 0/0 & 0 \\
gemini-3.1-pro & 431/355 & n/r/296 & 266/197 & 484/572 & 1,158 \\
gpt-6-astra & 33/31 & 0/0 & 30/21 & 55/45 & 21 \\
\bottomrule
\end{tabular}
\end{table*}

\subsection{Item Sampling, Climatology, and Serving}\label{app:sampling-serving}
Fire danger forecasting uses Mesogeos Track A 1~km daily records \citep{kondylatos2023mesogeos}. %
Positives are EFFIS fires $\ge 30$~ha; negatives lack burned area within 62~km on the window final day. %
The 4,120-item pool contains all 1,369 positive and 2,751 negative 2021--2022 samples (2.009:1 ratio). %
SHA-256 item hashes modulo ten assign ten folds without random seeds; fold zero yields 386 evaluation items (131 fires, 255 non-fires; 0.339 positive rate). %
Target dates span 2021-01-05 to 2022-09-29 (21 months), forming 352 monthly one-degree dependence blocks. %

Personnel allocation items come from ICS-209-PLUS situation reports with start years 2015--2020 \citep{stdenis2023ics209plus}. %
Selection keeps non-California reports with positive personnel and contiguous runs of at least ten daily reports. %
Linear spacing draws up to six fire-days per run, yielding 4,320 candidates. %
Stratified sampling draws 50 items per start year (seed 20260915), shuffled and capped at two items per incident. %
This yields 300 fire-days across 245 incidents (190 with one day, 55 with two). %
Report dates span 2015-06-24 to 2020-11-05; target dates span 2015-06-25 to 2020-11-06. %

The grounded fire-danger prompt appends monthly climatology for twelve dynamic surface and weather drivers. %
Variables are 2~m temperature, dewpoint, precipitation, surface pressure, wind speed, relative humidity, leaf area index, NDVI, soil moisture, daytime and night-time surface temperatures, and solar radiation. %
Statistics cover \mbox{2006--2019} Mesogeos Track A training years, omitting 2020 validation and 2021--2022 test years. %
The spatial unit is basin-wide, pooling valid Mediterranean daily observations. %
Each month provides driver means and population standard deviations over training days, appended before the question by end month. %

A calendar-month prior serves as the calibration baseline \citep{kondylatos2023mesogeos}. It is fit only on the 19,504 training samples from 2006 to 2019, indexing each fire label by the calendar month of its 30-day window end. %
Monthly positive rates range from 0.045 (April) to 0.642 (August), averaging 0.331. %
Assigned to the 386 test items by window end month, this prior achieves 0.555 average precision without accessing test labels. %

Proprietary models were queried through an OpenAI-compatible research gateway with endpoint withheld, requesting \texttt{claude-opus-4.8}, \texttt{claude-opus-5}, \texttt{gemini-3.1-pro}, and \texttt{gpt-6-astra}. %
Requests omitted provider flags and dated snapshots, leaving versions unpinned; stored served-model fields matched requested names. %
Open-weight models ran through Amazon Bedrock in \texttt{us-east-1}, pinning \texttt{qwen.qwen3-vl-\allowbreak 235b-a22b} for Qwen3-VL and \texttt{us.meta.llama4-maverick-\allowbreak 17b-instruct-v1:0} for Llama 4 Maverick. %

\subsection{Reasoning Settings and Token Use}\label{app:reasoning}
No request set reasoning, thinking, or effort parameters; models ran at endpoint defaults that released code cannot inspect. %
Requests specified temperature zero and a 1,536-token cap; the gpt-6-astra endpoint fixes temperature at 1 and treats the cap as a completion limit.
These are request settings, not verified backend settings: the gateway may omit or translate parameters, and stored records do not establish effective temperature, thinking mode, or effort.
Table~\ref{tab:reasoning-tokens} lists median reasoning tokens: Claude models report zero, gpt-6-astra reports 0--55, and gemini-3.1-pro reports 197--1,158. %
Zero reported reasoning tokens do not establish that internal reasoning was disabled. Released outputs support exact rescoring; backend defaults and versions limit rerun reproducibility.
gemini-3.1-pro neared the 1,536-token cap ($\ge 1{,}520$ tokens) on 28 bare and 16 grounded allocation items, one bare fire-danger item, and 61 bare tool-use items. %
Allocation and fire-danger responses had already given answer fields, but the 61 tool-use responses missed the ANSWER line (fallback parsing scored 3 correct; Appendix~\ref{app:defect-ours}). %
Bedrock reports no reasoning counts for open-weight models, and tool-arm records retain none because the runner logged only prompt and completion tokens. %

\section{AI Use Statement}
\label{app:ai-use}

The authors used AI coding and writing assistants to support implementation, analysis tooling, figure preparation, and manuscript drafting and revision. The reported empirical results are computed by evaluation code over benchmark records and saved experimental artifacts. The authors take full responsibility for the methods, interpretation, and final content of the paper.

\end{document}